\documentclass{article}

\usepackage[main, final]{neurips_2026}

\usepackage[utf8]{inputenc} 
\usepackage[T1]{fontenc}    
\usepackage{hyperref}       
\usepackage{url}            
\usepackage{booktabs}       
\usepackage{amsfonts}       
\usepackage{nicefrac}       
\usepackage{microtype}      
\usepackage{xcolor}         

\usepackage[utf8]{inputenc} 
\usepackage[T1]{fontenc}    
\usepackage{hyperref}       
\usepackage{url}            
\usepackage{booktabs}       
\usepackage{amsfonts}       
\usepackage{nicefrac}       
\usepackage{microtype}      
\usepackage{xcolor}         

\usepackage{amsmath}
\usepackage{amsthm}
\usepackage{enumitem}
\usepackage{wrapfig}
\usepackage{relsize}
\usepackage{graphicx}   
\usepackage{subcaption}
\usepackage{amssymb}
\newtheorem{theorem}{Theorem}[section]
\newtheorem{proposition}[theorem]{Proposition}
\newtheorem{lemma}[theorem]{Lemma}

\newtheorem{definition}[theorem]{Definition}
\newtheorem{assumption}[theorem]{Assumption}
\newtheorem{remark}[theorem]{Remark}
\newtheorem{example}[theorem]{Example}
\newtheorem{fact}[theorem]{Fact}

\title{An Axiomatic Framework for Reinforcement Learning in Open Multi-Agent Systems}

\author{%
  Jianhong Wang\thanks{Correspondence to \url{jianhong.wang@bristol.ac.uk} and \url{yang.li.cs@sjtu.edu.cn}. $^{\dagger}$Samuel Kaski is also with Aalto University and University of Manchester.} \\
  INFORMED-AI Hub\\
  University of Bristol\\
  \And
  Yang Li$^*$ \\
  School of Computer Science\\
  Shanghai Jiao Tong University\\
  \AND
  Samuel Kaski$^{\dagger}$ \\
   \\
  ELLIS Institute Finland\\
  \And
  Jonathan Lawry \\
  School of Engineering Mathematics and Technology \\
  University of Bristol
}

\begin{document}

\maketitle

\begin{abstract}
  Open multi-agent systems, where only a subset of agents are controllable, are central to real-world domains such as smart grids, yet lack a principled foundation for reinforcement learning algorithm design. We study this setting through n-agent ad hoc teamwork (NAHT), a recent framework for learning in open multi-agent environments. We propose an axiomatic framework that derives value function learning from the Shapley axioms (Efficiency, Symmetry and Linearity), treating them as structural constraints rather than heuristic design choices. This formulation builds on a novel NAHT decomposition, which characterises the fundamental difference between open and fully controllable multi-agent systems, and establishes a cooperative game-theoretic foundation for learning. We further show that enforcing these axioms on learning individual value functions recovers the Shapley value for Dec-POMDPs, providing a principled bridge between reinforcement learning and classical cooperative game theory. Building on this framework, we introduce Shapley Machines and Banzhaf Machines, two generic algorithmic templates that can be instantiated with standard reinforcement learning methods. When applied to strong base algorithms such as IPPO and POAM, our methods consistently improve learning efficiency and generalisation to unseen agent types and conventions across NAHT benchmark tasks. The code of experiments is published on \url{https://github.com/hsvgbkhgbv/axiomatic-rl-naht}.
\end{abstract}

\section{Introduction}
\label{sec:introduction}
    Multi-agent systems (MAS) have become a central paradigm for modelling real-world domains such as smart grids~\citep{wang2021multi}, railway networks~\citep{zhang2024improving}, and swarm robotics~\citep{nayak2023scalable,pmlr-v305-li25a}, where multiple agents must coordinate under uncertainty. Reinforcement learning (RL)~\citep{sutton2018reinforcement}, and in particular cooperative multi-agent reinforcement learning (CMARL), has shown promise in solving such problems. However, existing CMARL methods typically assume that all agents are controllable and that team composition is fixed, which we call the closed multi-agent systems; these assumptions rarely hold in practice.

    In many real-world scenarios, agents must operate in open multi-agent systems, where only a subset of agents is controllable and the remaining agents may be unknown or dynamically changing. This setting introduces two fundamental challenges: (i) generalisation to previously unseen teammate behaviours, and (ii) openness to varying team compositions. To address these challenges, recent work has introduced the n-agent ad hoc teamwork (NAHT)~\citep{wang2025n}, which extends CMARL to settings with partially controllable and dynamically formed teams.



    Although the NAHT objective is to maximise the team value over varying team compositions, this objective does not specify how agents can form cooperation in an open system with both controlled and uncontrolled agents. From the perspective of CMARL, an agent is represented by its individual value. This raises a fundamental question:

    \textit{How should individual value functions be structured and learned to support cooperation in open multi-agent systems?}

    In this paper, we address this question from the perspective of cooperative game theory. Specifically, we introduce an NAHT decomposition that represents a team value by two parts: internal cooperation among controlled agents and external teamwork involving uncontrolled agents. This decomposition reveals a structural distinction between open and closed multi-agent systems, which is not innately captured by the NAHT objective.

    Building on this structure, we propose an axiomatic framework to derive individual value functions, grounded in the axioms of the Shapley value~\citep{shapley1953value,dubey1975uniqueness}. Rather than constructing individual value functions directly by the explicit form of the Shapley value, we derive the individual value function learning by enforcing its axioms (Efficiency, Symmetry and Linearity) as structural constraints. This yields a principled approach to designing learning algorithms, where individual value functions are shaped to satisfy some properties, such as cooperative game-theoretic properties applied in this paper. Notably, we prove in theory that enforcing these axioms on learning individual value functions recovers the Shapley value for Dec-POMDPs~\citep{oliehoek2016concise}, while relaxing specific axioms leads to alternative payoff allocations, such as the Banzhaf index.

    More importantly, our axiomatic framework also reveals a novel connection between cooperative game theory and reinforcement learning. In particular, we show that the Linearity axiom induces a structure equivalent to the truncated TD($\lambda$)~\citep{cichosz1994truncating}, where the number of $n$-step return components can be estimated by the number of non-empty coalition basis games in cooperative game theory. This provides a theoretical explanation for the role of $n$-step returns and suggests a new perspective on the design of temporal-difference learning methods in the NAHT (see Section~\ref{subsec:relationship-between-the-number-and-the-values-of-basis-games}). 
    
    \textit{NAHT includes two extreme cases: single-agent RL with an environment of other uncontrolled agents, and CMARL with fully controllable agents. The above result therefore extends to both settings.}
    
    Based on this axiomatic framework, we introduce two generic algorithmic templates: Shapley Machines, which enforce all axioms, and Banzhaf Machines, which relax the Efficiency axiom. These templates can be instantiated with existing RL algorithms, such as IPPO~\citep{de2020independent} and POAM, to yield practical methods for solving the NAHT.

    We evaluate the proposed methods on modified MPE and SMAC benchmarks adapted to the NAHT setting~\citep{wang2025n}. Our results show that the proposed framework consistently improves learning efficiency and generalisation to unseen agent types and conventions. Furthermore, we demonstrate that relaxing specific axioms can lead to improved performance in certain regimes, highlighting the flexibility of the proposed axiomatic perspective. Due to the limit of space, we leave the related work section to Appendix~\ref{sec:related-work}.

\section{Background}
\label{sec:background}
    \subsection{N-Agent Ad Hoc Teamwork}
    \label{subsec:n-agent-aht}
        In this section, we describe a problem setting for open multi-agent systems, referred to as n-agent ad hoc teamwork (NAHT)~\citep{wang2025n}. The main challenges of NAHT are as follows: (1) cooperating with potentially unknown types of teammates (generalization), and (2) coping with a varying number of uncontrolled teammates (openness). A decentralized partially observable Markov decision process (Dec-POMDP)~\citep{oliehoek2016concise} is considered to formalize the problem. There is a team of agents $\mathcal{M}$ whose size is denoted as $M$, a state space $\mathcal{S}$, a joint action space $\mathcal{A}=\times_{i \in \mathcal{M}} \mathcal{A}_{i}$, a per-agent observation space $\mathcal{O}_{i}$, a transition function $P_{T}: \mathcal{S} \times \mathcal{A} \rightarrow \Delta(\mathcal{S})$, a team reward function $R_{t}: \mathcal{S} \times \mathcal{A} \rightarrow \mathbb{R}$, a discount factor $\gamma \in [0, 1]$ and an episode length $T \in \mathbb{Z}^{+}$. Each agent receives observations via an observation function $O_{i}: \mathcal{S} \times \mathcal{A} \rightarrow \Delta(\mathcal{O}_{i})$. $\mathcal{H}_{i}$ is defined as an agent $i$'s space of observation and action histories. Each agent is equipped with a policy $\pi_{i}: \mathcal{H}_{i} \rightarrow \Delta(\mathcal{A}_{i})$.

        In this problem, there are two groups of agents: a set of controlled agents denoted as $\mathcal{C}$ whose size is denoted as $N$, and a set of uncontrolled agents denoted as $\mathcal{U}$ whose size is denoted as $M - N$. As NAHT is an open multi-agent system, at the beginning of a multi-agent interaction process, a team of agents $\mathcal{M}$ are randomly drawn from the controlled agent pool and the uncontrolled agent pool. For conciseness, each controlled agent is characterised by its policy, and the set of controlled agents can be expressed as $\mathcal{C}(\theta) = \{\pi_{i}^{\theta}\}_{i=1}^{N}$. The above sampling procedure is denoted as $X(\mathcal{U}, \mathcal{C}(\theta))$. The aim of NAHT is learning parameters $\theta$ to solve the following optimization problem:
        \begin{equation}
            \max_{\theta} \mathbb{E}_{\pi^{(\mathcal{M})} \sim X(\mathcal{U}, \mathcal{C}(\theta))} \left[ \sum_{t=0}^{T} \gamma^{t} R_{t} \right],
        \label{eq:naht-objective}
        \end{equation}
        where $\pi^{(\mathcal{M})}$ denotes the joint policy of a team of agents $\mathcal{M}$. Note that different number of controlled agents $N$ would lead to various teams of agents $\mathcal{M}$; the expectation in Eq.~\ref{eq:naht-objective} is usually approximated by a team value which we will specify in the following section.

    \subsection{Cooperative Games and Shapley Value}
    \label{subsec:shapley-value}
        \textbf{Representation of Cooperative Games.} A cooperative game is specified by a pair $(\mathcal{M}, v)$, where $\mathcal{M}$ is the set of agents and $v: 2^{\mathcal{M}} \rightarrow \mathbb{R}_{\geq 0}$ is its characteristic function. When the agent set is fixed, we identify the game with its characteristic function $v$ for brevity. An arbitrary cooperative game $v$ can be uniquely represented by a set of basis games $\{v_{C}^{1} \ | \ \emptyset \neq C \subseteq \mathcal{M} \}$~\citep{dubey1975uniqueness}, as delineated in Definition~\ref{def:game-set}. In this paper, we mainly focus on investigating how cooperative games instantiated over a team of agents $\mathcal{M}$ are used to represent a team value of the NAHT. As per previous work in the MARL~\citep{wang2020shapley}, we consider the superadditive game as the game class of $\mathcal{G}$, which is suitable for a cooperative multi-agent task, such as the NAHT (see Appendix~\ref{subsec:superadditive-game-cmarl} for more details). 
        \begin{definition}[\textbf{Cooperative Games}~\citep{dubey1975uniqueness}]
        \label{def:game-set}
             We define a function $D \mapsto v_{C}^{z}(D)$ for an arbitrary $z \in \mathbb{R}$ such that if a coalition $D \supseteq C$, $v_{C}^{z}(D) = z$, otherwise, $v_{C}^{z}(D) = 0$. For a cooperative game described by a characteristic function $v: 2^{\mathcal{M}} \rightarrow \mathbb{R}_{\geq 0}$, we can describe a set of such games over a team of agents $\mathcal{M}$ by a set of basis games $\{v_{C}^{1} \ | \ \emptyset \neq C \subseteq \mathcal{M} \}$, such that $\mathcal{G} = \{ w \ | \ w = \sum k_{C} v_{C}^{1}, k_{C} \in \mathbb{R}, \emptyset \neq C \subseteq \mathcal{M} \}$.\footnote{The analytic form of $k_{C}$ is represented as $k_{C} = \sum_{T \subseteq C} (-1)^{|C| - |T|} v(T)$.}
        \end{definition}
        
        \textbf{Shapley Value.} Shapley value is a multidimensional linear transformation $\phi: \mathcal{G} \rightarrow \mathbb{R}^{M}$ defined on the set of cooperative game $\mathcal{G}$, given that there are $M$ agents in total. Each dimension of $\phi$ indicates the payoff allocation to an agent. It has been proved that Shapley value is a unique payoff allocation function on the set of cooperative games $\mathcal{G}$ which satisfies the following axioms: Efficiency, Symmetry and Additivity~\citep{dubey1975uniqueness},\footnote{For an arbitrary characteristic function game, Shapley value is uniquely determined by the Efficiency, Additivity, Symmetry, and Dummy Player axioms~\citep{chalkiadakis2011computational}[Chap. 2]. Since the Dummy Player axiom is embedded in the basis game construction of~\cite{dubey1975uniqueness}, it has been naturally satisfied.} highlighted in the following Theorem~\ref{thm:shapley-axioms}:
        \begin{theorem}[\textbf{Axioms of Shapley Value}~\citep{dubey1975uniqueness}]
        \label{thm:shapley-axioms}
            Shapley value is a unique payoff allocation function on the cooperative game space $\mathcal{G}$, satisfying Efficiency, Symmetry and Additivity.
        \end{theorem}
        
\section{N-Agent Ad Hoc Teamwork Decomposition}
\label{sec:a-model-for-n-agent-ad-hoc-teamwork}
    To understand how learning differs in open multi-agent systems, we first characterise the cooperative-game structure of team values of the NAHT. Recall that in the NAHT a team consists of two groups of agents: a set of controlled agents $\mathcal{C}$, whose policies are optimized, and a set of uncontrolled agents $\mathcal{U}$, whose behaviours are unknown and may vary across episodes. Unlike the closed multi-agent systems, the performance of a team in the NAHT depends not only on coordination among controlled agents, but also on their ability to adapt to diverse and potentially unseen teammates. To make this distinction explicit, we represent a team value at a given state in the NAHT as a state-specific cooperative game value that is the weighted sum of non-empty coalition basis game values, from the perspective of cooperative game theory, as follows: 
    \begin{definition}[\textbf{State-Specific Cooperative Games}]
    \label{def:generalised-basis-games}
        We define a function $D \mapsto v_{C, s}^{z}(D)$, for any real value $z \in \mathbb{R}$ and state $s \in \mathcal{S}$, such that if a coalition $D \supseteq C$, $v_{C, s}^{z}(D) = z$, otherwise, $v_{C, s}^{z}(D) = 0$. It is not difficult to observe that we have $v_{C, s}^{z} = z v_{C, s}^{1}$, where $v_{C, s}^{1}$ is a state-specific basis game for state $s$. Similarly, a set of generalised cooperative games to state $s$, referred to as state-specific cooperative games, is represented as $\mathcal{G}(s) = \{ w_{s} \ | \ w_{s} = \sum k_{C} v_{C,s}^{1}, k_{C} \in \mathbb{R}, \emptyset \neq C \subseteq \mathcal{M} \}$. In other words, an arbitrary state-specific game $v_{s} \in \mathcal{G}(s)$ can be represented as $v_{s} = \sum k_{C} v_{C,s}^{1}$.
    \end{definition}
    
    \begin{definition}[\textbf{Game Structure for NAHT}]
    \label{def:ad-hoc-teamwork-in-cooperative-game}
        $\mathcal{G}_{\texttt{NAHT}} = \mathlarger{\mathlarger{\times}}_{s \in \mathcal{S}} \mathcal{G}(s)$ denotes the set of all NAHT processes, where a team value of an NAHT process at state $s$ is represented as a state-specific cooperative game value $v_{s}(\mathcal{M})$, where $v_{s} \in \mathcal{G}(s)$. Each element belonging to $\mathcal{G}_{\texttt{NAHT}}$ is a tuple of state-specific cooperative games, defined by a team of agents $\mathcal{M}$, a state space $\mathcal{S}$, and weightings $k_{C}$ associated with the state-specific basis games of all non-empty coalitions $\emptyset \neq C \subseteq \mathcal{M}$.
    \end{definition}

    Based on generalisation of the cooperative game representation from the classic cooperative game as shown in Definition~\ref{def:generalised-basis-games} and the game structure for the NAHT in Definition~\ref{def:ad-hoc-teamwork-in-cooperative-game}, we define a state-specific cooperative game value $v_{s}(\mathcal{M})$, as the team value of the NAHT at state $s$, as follows:
    \begin{equation}
    \label{eq:naht-decomposition}
        v_{s}(\mathcal{M}) = \underbrace{\sum_{C_{I} \in \mathcal{P}_{+}(\mathcal{C})} k_{C_{I}} v_{C_{I},s}^{1}(\mathcal{M})}_{\textbf{Internal Cooperation}} \ + \ \underbrace{\sum_{C_{E} \in \mathcal{P}_{+}(\mathcal{M}) \setminus \mathcal{P}_{+}(\mathcal{C})} k_{C_{E}} v_{C_{E},s}^{1}(\mathcal{M})}_{\textbf{External Teamwork}},
    \end{equation}
    where $\mathcal{M} = \mathcal{C} \cup \mathcal{U}$ denotes the full team,  $\mathcal{P}_{+}(\cdot)$ indicates the powerset of a set excluding the empty set, and each term corresponds to a contribution from a subset (coalition) of agents.
    The first term captures internal cooperation, i.e., interactions solely among controlled agents $\mathcal{C}$. This corresponds to the standard task in the CMARL, where all agents are optimised jointly. The second term captures external teamwork, i.e., interactions between controlled agents $\mathcal{C}$ and uncontrolled agents $\mathcal{U}$, along with interactions among uncontrolled agents $\mathcal{U}$. The second term is unique to open multi-agent systems, and reflects the need to coordinate with agents whose behaviours are not controlled and may vary across episodes. When all agents are controlled, i.e., $\mathcal{C}=\mathcal{M}$, the external teamwork part vanishes, and the decomposition reduces to internal cooperation, recovering the fully controllable setting of the CMARL.
    \begin{wrapfigure}{r}{0.31\textwidth}
        \centering
        \includegraphics[width=\linewidth]{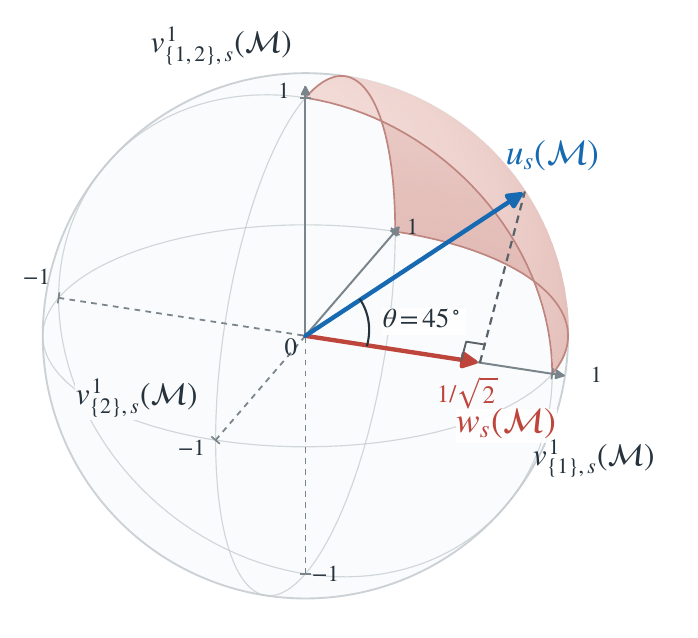}
        \caption{Geometric illustration of the NAHT decomposition. The red and blue vectors are two games represented by different basis game sets, denoted by $w_{s}(\mathcal{M})$ and $u_{s}(\mathcal{M})$. Their difference is measured by the $\theta$. The shading area in red indicates the range of generated cooperative games with that $k_{C} \geq 0$ and $\sum (k_{C})^{2} = 1$.}
    \vspace{-18mm}
    \label{fig:state-space-cooperative-game-case}
    \end{wrapfigure}
    
    \begin{example}
    \label{example:game-structured-naht}
        As shown in Figure~\ref{fig:state-space-cooperative-game-case}, suppose we have a team of agents $\mathcal{M} = \{ 1, 2 \}$ for a state $s$, where $\mathcal{C} = \{ 1 \}$ and $\mathcal{U} = \{ 2 \}$. $\theta$ is the angle between two vectors representing two situations, such that $w_{s} \in W = \text{span}\{v_{\{1\}, s}^{1}\}$ and $u_{s} \in \mathcal{G}(s) = \text{span}\{v_{\{1\}, s}^{1}, v_{\{2\}, s}^{1}, v_{\{1,2\}, s}^{1}\}$. It is obvious that $w_s(\mathcal{M})$, as a projection of $u_s(\mathcal{M})$, loses information about the external teamwork.
    \end{example}

    \textbf{Implication.} This decomposition highlights a fundamental difference between NAHT and CMARL: In the CMARL, all agents are controllable, and learning focuses exclusively on internal cooperation. In the NAHT, learning must additionally account for external teamwork with uncontrolled agents. Ignoring the external component effectively reduces the problem to a closed multi-agent system, which introduces bias and limits generalisation to unseen team compositions. The NAHT decomposition therefore provides a structural characterisation of openness, and it will serve as the foundation for the axiomatic framework introduced as follows.

\section{From Shapley Axioms to Learning Algorithms}
\label{sec:shapley-value-for-n-agent-ad-hoc-teamwork}
    Building on the NAHT decomposition, we now derive an axiomatic framework for learning in open multi-agent systems. Our approach is based on the axioms of the Shapley value, which we interpret as structural constraints on individual value functions. In detail, we aim to show that how the individual values, as the output of a multidimensional linear transformation $\phi$ applied to a state-specific cooperative game $v_s$, satisfy the Shapley value's axioms in learning. When we enforce all these axioms together, the Shapley-form individual values are uniquely shaped (see Theorem~\ref{thm:shapley-machine}). Moreover, the axioms can be directly translated into algorithmic design principles: (1) The Efficiency axiom corresponds to reward shaping; (2) The Linearity axiom corresponds to truncated TD($\lambda$)-style updates; (3) The Symmetry axiom corresponds to individual value function structures. These principles will later lead to two generic algorithmic templates, referred to as Shapley Machine and Banzhaf Machine.
    
    \subsection{From Game Structure to Dec-POMDPs}
    \label{subsec:from-game-structure-to-decpomdp}
        The game structure for the NAHT in Definition~\ref{def:ad-hoc-teamwork-in-cooperative-game} can be used to represent any sequential decision-making process evaluated in states. However, it does not necessarily imply that the process embedded in the game structure strictly follows the problem setting of the NAHT --- Dec-POMDPs~\citep{oliehoek2016concise}. To bridge this gap, we augment the game structure with Markov dynamics and per-step team rewards to link the values of different states. When the state-specific cooperative game value for each state $s_{t}$ is represented as $v_{s_{t}}(\mathcal{M}) = \mathbb{E}_{\pi}[\sum_{\tau=0}^{T-t-1} \gamma^{\tau} R_{t+\tau} \ | \ s_{t}]$, we can derive the expected $n$-step return in RL~\citep{sutton2018reinforcement}[Chap. 7], with a given horizon $n$ to link $v_{s_{t}}(\mathcal{M})$ and $v_{s_{t+n}}(\mathcal{M})$, such that
        \begin{equation}
            \begin{split}
                v_{s_{t}}(\mathcal{M}) = \mathbb{E}_{\pi} \left[ \sum_{\tau=0}^{n-1} \gamma^{\tau} R_{t+\tau} + \gamma^{n} v_{s_{t+n}}(\mathcal{M}) \ \middle\vert \ s_{t} \right] := \mathbb{E}_{\pi} \left[ G_{t:t+n} \ \middle\vert \ s_{t} \right].
            \end{split}
        \label{eq:n-step-return-game-structure}
        \end{equation}
        
        For ease of presentation, we denote the state-specific basis game value $v_{C, s}^{v_{s}(\mathcal{M})}(\mathcal{M})$ by $V(C, s)$.
        \begin{proposition}
        \label{prop:powerset-identity-main}
            By Definition~\ref{def:powerset-enum} and a horizon assignment to match coalition and $n$-step return orderings, for every state $s\in\mathcal S$,
            \begin{equation*}
                \begin{split}
                    \big(V(C_1,s), \dots, V(C_m,s)\big) \;=\; \big(\mathbb E_\pi[G_{t:t+1}\mid s],\dots,\mathbb E_\pi[G_{t:t+m}\mid s]\big),
                \end{split}
            \end{equation*}
            and, each $V(C_i,s)$ is equal to $v_{s}(\mathcal{M})$ in value.
        \end{proposition}
        By Proposition~\ref{prop:powerset-identity-main} (see the proof in Appendix~\ref{subsec:links-to-rl}), we obtain the following formula that aligns the values of state-specific basis games of various coalitions to the expected $n$-step returns with the corresponding horizons $n \in \{1, ..., m\}$ as their representation, for all $\emptyset \neq C_{n} \subseteq C_{n+1}, C_{m} = \mathcal{M}$, such that
        \begin{equation}
            V(C_{n}, s_{t}) = \mathbb{E}_{\pi} \left[ \sum_{\tau=0}^{n-1} \gamma^{\tau} R_{t+\tau} + \gamma^{n} v_{s_{t+n}}(\mathcal{M}) \ \middle\vert \ s_{t} \right].
        \label{eq:game-value-expansion-coalition}
        \end{equation}
        Specifically, the sampled $n$-step return $G_{t:t+n}$ is an unbiased estimator of $V(C_n,s)$ conditional on $s_t=s$. Under the assumptions in Appendix~\ref{sec:justification-of-size-lex}, the variance of the state-specific basis game value estimators is non-decreasing in coalition size, while $n$-step return variance is non-decreasing in horizon. This correspondence motivates the horizon assignment in Eq.~\eqref{eq:game-value-expansion-coalition}. 

    \subsection{Linearity Axiom and Temporal-Difference Learning}
    \label{subsec:linearity-axiom}
        The Linearity axiom plays a central role in connecting cooperative game theory to temporal-difference learning. We show this in the following Proposition~\ref{prop:linearity-td} (see the proof in Appendix~\ref{sec:proof-linearity-axiom}):
        \begin{proposition}
        \label{prop:linearity-td}
            Given the expression of state-specific basis game values in Eq.~\eqref{eq:game-value-expansion-coalition}, applying the Linearity axiom to state-specific cooperative games of Eq.~\eqref{eq:naht-decomposition} can lead to the following equation:
            \begin{equation}
                \phi(v_{s_{t}}) = \mathbb{E}_{\pi} \left[ \sum_{n=1}^{m} k_{C_{n}}' \cdot \sum_{\tau=0}^{n-1} \gamma^{\tau} \phi(R_{t+\tau}) + \gamma^{n} \phi(v_{s_{t+n}}) \ \middle\vert \ s_{t} \right].
            \label{eq:shapley-value-expected-game}
            \end{equation}
        \end{proposition}
        In Proposition~\ref{prop:linearity-td}, the term within the $\mathbb{E}_{\pi}[\cdot]$ takes the form of truncated $\lambda$-return~\citep{sutton2018reinforcement}[Chap. 12], when all $k_{C_{n}}'$ are seen as weightings for the $n$-step return components.

        In practice, we only need to learn $\phi_{i}(v_{s_{t}})$ directly for each controlled agent $i \in \mathcal{C}$. For brevity, we express $\phi_{i}(v_{s_{t}})$ as $V_{i}(s_{t})$ in the following paragraphs. Based on Eq.~\eqref{eq:shapley-value-expected-game}, we derive the TD error $\delta_{t,i}$ of truncated TD($\lambda$) (TTD($\lambda$))~\citep{sutton2018reinforcement}[Chap. 12] for each agent $i$ purely from the game structure for the NAHT, such that
        \begin{equation}
            \delta_{t,i} := \sum_{n=1}^{m} k_{C_{n}}' \cdot \sum_{\tau=0}^{n-1} \left( \gamma^{\tau} \phi_{i}(R_{t+\tau}) + \gamma^{n} V_{i}(s_{t+n}) \right) - V_{i}(s_{t}).
        \label{eq:td-error-general}
        \end{equation}
        \begin{remark}
            The above result provides a theoretical explanation for the role of the $n$-step return in the NAHT; it gives an insight that the truncated TD($\lambda$) learning arises naturally from the Linearity axiom and the NAHT decomposition to capture both internal cooperation and external teamwork, given an appropriate horizon.
        \end{remark}
            
    \subsection{Efficiency Axiom and Reward Shaping}
        Note that $\phi(R_{t+\tau})$ in Eq.~\eqref{eq:td-error-general} has not yet been defined so far, because $R_{t+\tau}$ is not belonging to the state-specific cooperative game space $\mathcal{G}(s)$. As a result, it is necessary to represent $\phi(R_{t+\tau})$ in the form of $\phi(v_{s})$, where $v_{s} \in \mathcal{G}(s)$ is the underlying state-specific cooperative game. The Efficiency axiom can address this problem. To satisfy the Efficiency axiom such that $\sum_{i=1}^{M} \phi_{i}(v_{s_{t}}) = v_{s_{t}}(\mathcal{M})$, we reasonably presume that its sufficient condition holds: $\sum_{i=1}^{M} \phi_{i}(R_{t}) = R_{t}$. Then, we demonstrate that how $\phi_{i}(R_{t})$ is represented by the Efficiency axiom in the following Proposition~\ref{prop:shapley-machine-reward} (see the proof in Appendix~\ref{sec:proof-linearity-symmetry-axiom}):
        \begin{proposition}
        \label{prop:shapley-machine-reward}
            Given the condition $\sum_{i=1}^{M} \phi_{i}(R_{t}) = R_{t}$, each agent's payoff allocation defined on rewards $R_{t}$, i.e., $\phi_{i}(R_{t})$, can be expressed as
            \begin{equation}
                \phi_{i}(R_{t}) := R_{t} - \sum_{j \neq i} \left( \phi_{j}(v_{s_{t}}) - \gamma \phi_{j}(v_{s_{t+1}}) \right).
            \label{eq:individual-reward-approximation}
            \end{equation}
        \end{proposition}

        By replacing $\phi_{j}(v_{s_{t}})$ with $V_{j}(s_{t})$ as there is in Eq.~\eqref{eq:td-error-general}, we get that
        \begin{equation}
            \phi_{i}(R_{t}) = R_{t} - \sum_{j \neq i} \left( V_{j}(s_{t}) - \gamma V_{j}(s_{t+1}) \right).
        \label{eq:shape-rewards}
        \end{equation}

        The condition $\sum_{i=1}^{M} \phi_{i}(R_{t}) = R_{t}$ for deriving the Efficiency axiom is also referred to as additively factored immediate reward function. The following Theorem~\ref{thm:factored-dec-pomdp} shows that the Efficiency axiom always exists in the Dec-POMDP, given a factorization scheme of the transition function.
        \begin{theorem}[\textbf{Principle of Decomposing Value Functions}~\citep{oliehoek2016concise}]
        \label{thm:factored-dec-pomdp}
            Given an additively factored reward function, for any timestep $t$ there is a factorization of the transition function, such that the value of a finite-horizon factored Dec-POMDP is decomposable across agents.
        \end{theorem}
        
        Since NAHT is considered in Dec-POMDPs (see Section~\ref{subsec:n-agent-aht}), to guarantee the feasibility of the Efficiency axiom as shown in Theorem~\ref{thm:factored-dec-pomdp}, another constraint is added to implicitly facilitate searching for a factorization of the transition function, such that
        \begin{equation}
            \sum_{i=1}^{M} V_{i}(s_{t}) = v_{s_{t}}(\mathcal{M}).
        \label{eq:efficiency}
        \end{equation}
            
    \subsection{Symmetry Axiom and Value Function Structure}
        The Symmetry axiom here means that two agents who contribute equally to every possible coalition (excluding the ones involving themselves) should receive equal payoff allocations~\citep{chalkiadakis2011computational}[Chap. 2]. In the literature from \citet{dubey1975uniqueness}, it was described as the permutation-equivariance of an multidimensional operator $\phi$ defined on a cooperative game, as shown in the following Theorem~\ref{thm:symmetry-axiom} (see the proof in Appendix~\ref{sec:proof-linearity-symmetry-axiom}):
        \begin{theorem}
        \label{thm:symmetry-axiom}
            The payoff allocation operator satisfying permutation-equivariance is a sufficient condition for the Symmetry axiom.
        \end{theorem}
        Next, we aim to show how to construct such an operator to fulfil the permutation-equivariance, such that $\phi_{\sigma(i)}(\sigma(v_{s})) = \phi_{i}(v_{s})$ can be constructed in practice, where $\sigma$ means relabelling agent IDs to another ordering, e.g., $(1,2) \mapsto (2,1)$. Recall that $\phi_{i}(v_{s})$ is expressed as $V_{i}(s)$. In plain words, this means that relabelling the agents should simply relabel their individual values, which can be implemented by the following two conditions: (1) The operator $\phi$'s output is changeable consistently with reordering agent IDs, e.g., $(1, 2) \mapsto (2, 1) \Rightarrow (V_{1}(s), V_{2}(s)) \mapsto (V_{2}(s), V_{1}(s))$; (2) The operator's output with respect to each agent should be invariant to reordering agent IDs on its input $\mathtt{inp}$ in practice (e.g., the world information $s$ and/or other agents' information), e.g., $V_{i}(\sigma(\mathtt{inp})) = V_{i}(\mathtt{inp})$. Note that the second condition usually holds when other agents' information is not directly included in the input to an agent's individual value function, such as POAM~\citep{wang2025n} and IPPO~\citep{de2020independent}.
        
    \subsection{From Shapley Axioms to Learning Algorithms}
    \label{subsec:implementation-in-marl}
        The algorithm that implements $V_{i}$ as agents' individual values to fulfil the Efficiency, Linearity and Symmetry axioms through learning is named as \textbf{Shapley Machine}: (1) Shape individual rewards $\phi_{i}(R)$ following Eq.~\eqref{eq:shape-rewards} and set Eq.~\eqref{eq:efficiency} as a regularization term, to fulfil the Efficiency axiom. (2) Implement the TTD($\lambda$) following Eq.~\eqref{eq:td-error-general}, to fulfil the Linearity axiom. The $k_{C}' > 0$\footnote{See Remark~\ref{rmk:reason-for-k-greater-than-zero} for why $k_{C}' > 0$ is chosen.} is implemented as geometric distribution following the convention of TTD($\lambda$)~\citep{sutton2018reinforcement}[Chap. 12]. (3) Structure the policy and individual value networks to fulfil the Symmetry axiom. 
        
        Since the Linearity axiom is a sufficient condition for the Additivity axiom (see Appendix~\ref{subsec:axioms-of=shapley-value}), the individual values satisfying all those axioms in implementation realizes Shapley values for the Dec-POMDPs, as highlighted in the following Theorem~\ref{thm:shapley-machine} (see the proof in Appendix~\ref{sec:proof-linearity-symmetry-axiom}):
        \begin{theorem}
        \label{thm:shapley-machine}
            Shapley Machine is an algorithm enforcing $V_{i}$ to fulfil Efficiency, Symmetry and Linearity, so the $V_{i}$ is the Shapley value for Dec-POMDPs.
        \end{theorem}
        
        To realize the Shapley Machine, we use two base algorithms structurally satisfying the Symmetry axiom: POAM~\citep{wang2025n} and IPPO~\citep{de2020independent}. We then shape the objective functions following conditions imposed by other two axioms, to obtain the algorithms \textbf{SM-POAM} and \textbf{SM-IPPO}, respectively. The details of implementation are left to Appendix~\ref{subsec:implementation-details}.
        
        Following the same principle of designing algorithms via axiomatic characterization, we also propose a variant that enforces the Linearity and Symmetry axioms while omitting Efficiency.\footnote{Without the Efficiency axiom as a constraint on $\phi(R_{t})$, Banzhaf Machine uses $\phi_i(R_t)=R_t$ as one implementation.} Since the resulting $V_{i}$ aligns with the Banzhaf index, we call this algorithm \textbf{Banzhaf Machine}. Analogous to the implementation of Shapley Machine, its instantiation on POAM and IPPO are named as \textbf{BM-POAM} and \textbf{BM-IPPO}, respectively.
        \begin{remark}
            Banzhaf Machine is an algorithm generating $V_{i}$, fulfilling Symmetry and Linearity.
        \end{remark}
        
\section{Experiments}
\label{sec:experiments}
    In experiments, we evaluate the proposed SM-POAM, SM-IPPO, BM-POAM and BM-IPPO on the modified MPE-PP and SMAC benchmark tasks tailored to the NAHT settings~\citep{wang2025n}, against the counterpart baseline algorithms POAM and IPPO. The uncontrolled agents are taken from five pre-trained policies, which were trained in fully controllable CMARL settings using the algorithms IPPO, MAPPO, VDN, QMIX, and IQL. To avoid confusion between the standard IPPO algorithm (trained under the CMARL settings) and the version trained within the NAHT process (where only a subset of agents is controllable), we refer to the latter as IPPO-NAHT. 
    
    We validate the game structure for the NAHT, a core concept in our theory, by setting the number of $n$-step return components considered in TTD($\lambda$) using the number of basis games $m$. Since the number of basis games is far more than the preset episode length (511 vs. 150) in 10v11, we approximate the number of basis games to the episode length as a trade-off. About the 8v9 scenario, the number of basis games needed in theory is close to the episode length (127 vs. 120), so the number of basis games can be naturally approximated by the episode length. As a result, for 8v9 and 10v11 Banzhaf Machine is equivalent to the baseline counterparts in implementation. The details of baselines, experimental settings and evaluation metrics are provided in Appendix~\ref{sec:experimental-details}. All results are obtained by first computing each metric with 128 episodes and then averaging these per-seed metrics across 5 random seeds, with 95\% confidence intervals reported.

    \textbf{Research Questions.} We focus on the following research questions in experimentation: (1) Is the number of coalitions $m$ used to decide the horizon of TTD($\lambda$) (see Section~\ref{subsec:linearity-axiom}) a reasonable choice? (2) Are all Shapley value's axioms equally appropriate across scenarios? (3) Which axiom is key to generalization to unseen conventions or unseen agent types? 
    
    \subsection{Main Results and Analysis}
    \label{subsec:main-result}
        \begin{figure}[ht!]
          \centering
          \includegraphics[width=0.65\linewidth]{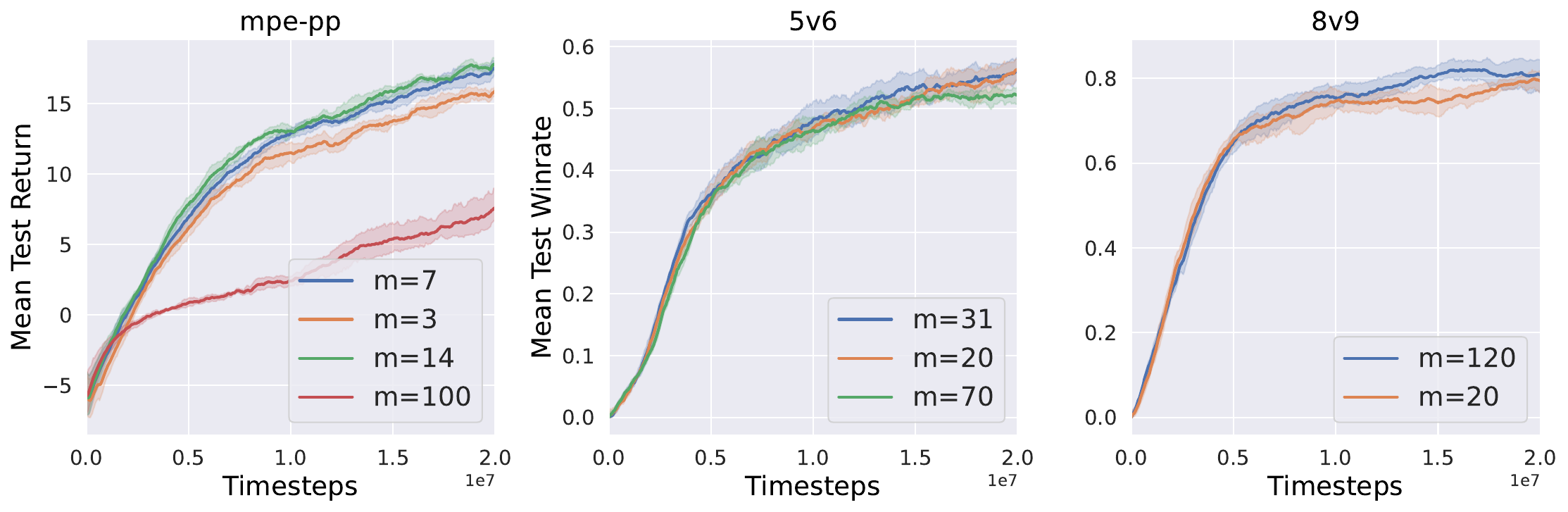}
          \caption{Experiments verifying that the number of $n$-step return components in TTD($\lambda$) can be determined by the number of non-empty coalition basis games. In MPE-PP the number of basis games is $m=7$, in 5v6 it is $m=31$, and in 8v9 it is $m=120$.}
        \label{fig:basis-games}
        \end{figure}

        \textbf{Answer to Question 1.} We conduct experiments with varying numbers of basis games $m$ using SM-POAM as the candidate algorithm. As shown in Figure \ref{fig:basis-games}, learning performance in MPE-PP and 8v9 is sensitive to the choice of $m$, while in 5v6 it is comparatively robust. Overall, these results indicate that setting the number of $n$-step return components as the number of non-empty coalition basis games can nearly maximize learning performance, thereby providing an empirical evidence to support the mathematical rationale behind the Linearity axiom that derives this setting for the NAHT. Although we do not deny that there may exist other preferred choices of $m$ (as observed in the results of MPE-PP and 5v6), this does not undermine the effectiveness of using the number of non-empty coalition basis games, as a criterion, to decide the horizon for TTD($\lambda$) in practice.
        
        \begin{figure*}[ht!]
            \centering
            \includegraphics[width=\linewidth]{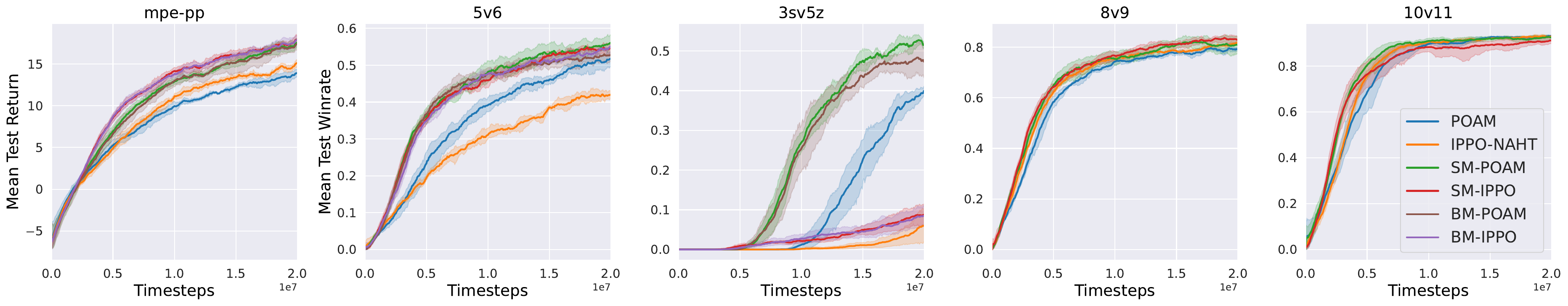}
            \caption{Evaluation curves during training in NAHT.}
        \label{fig:test-return}
        \end{figure*}
        \textbf{Answer to Question 2.} We evaluate all baselines alongside our proposed algorithms. For the large-scale tasks 8v9 and 10v11, the base algorithms IPPO-NAHT and POAM are effectively equivalent to BM-IPPO and BM-POAM under the approximation of $m$ due to limited episode lengths, and are therefore omitted in implementation. Figure~\ref{fig:test-return} shows that, for either IPPO-NAHT or POAM, at least one of their Shapley Machine or Banzhaf Machine variants achieves the best performance. Recall that Banzhaf Machine corresponds to dropping the Efficiency axiom. In MPE-PP and 3sv5z with IPPO-NAHT as the base algorithm, Banzhaf Machine performs on par with Shapley Machine. In most other scenarios (except 10v11 with IPPO-NAHT), Shapley Machine outperforms Banzhaf Machine, although sometimes by a narrow margin. Moreover, comparisons between the base algorithms and their Banzhaf counterparts in MPE-PP, 5v6, and 3sv5z underscore the importance of the Linearity axiom. Overall, the results suggest that all three Shapley axioms generally enhance learning performance (in possibly unseen team compositions), but the Efficiency axiom can, in some cases, reduce it. This justifies the value of our proposed axiomatic framework, which provides a novel perspective on analysing performance through the falsifiable axioms. We believe that this axiomatic framework, as a paradigm, can help design new algorithms in the future.
        
        \begin{figure*}[ht!]
            \begin{subfigure}[b]{\textwidth}
                \centering
                \includegraphics[width=\linewidth]{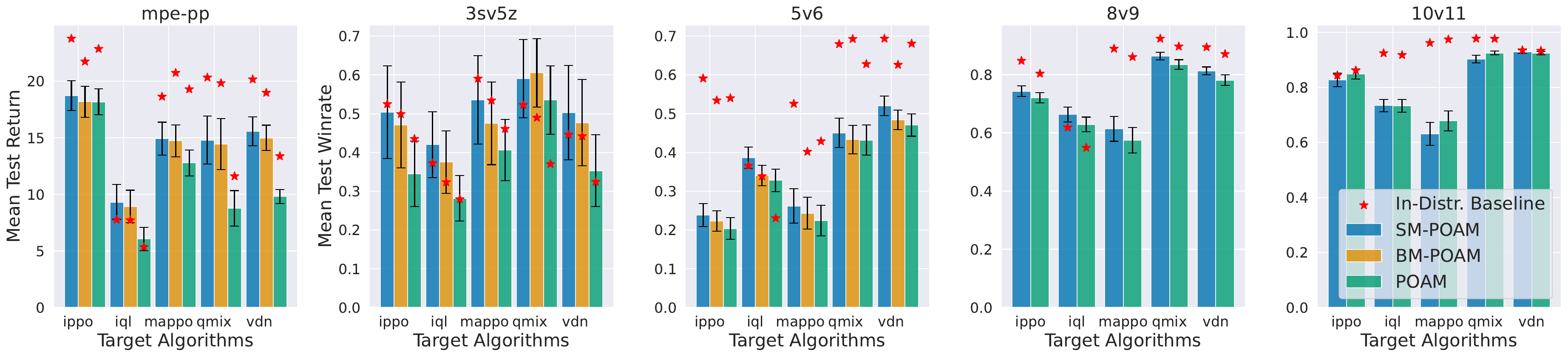}
                \caption{Evaluation on unseen conventions (identical algorithms with different random seeds). The star denotes the in-convention baseline, and the histograms show performance on unseen conventions.}
            \label{fig:unseen-convention-poam}
            \end{subfigure}
            \hfill
            \begin{subfigure}[b]{\textwidth}
                \centering
                \includegraphics[width=\linewidth]{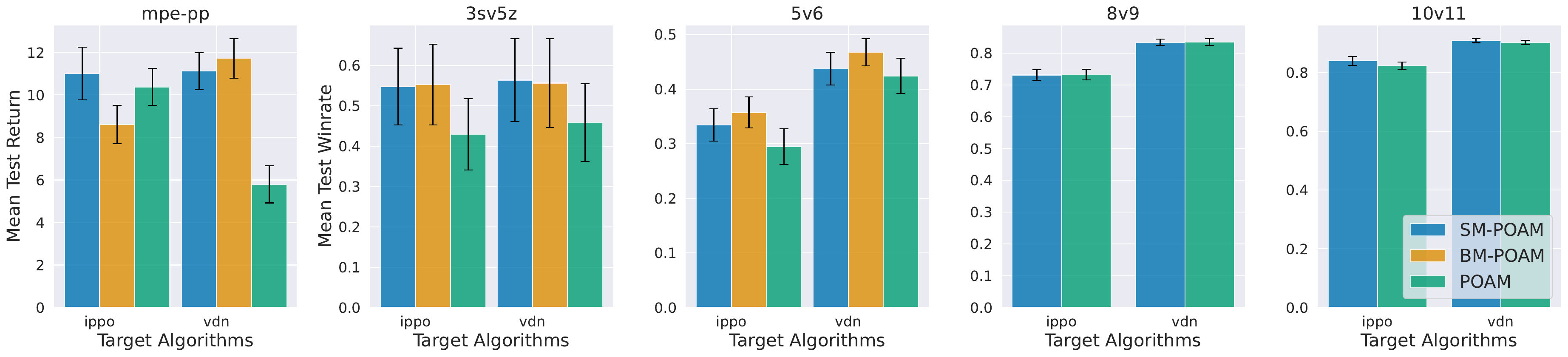}
                \caption{Evaluation on unseen agent types. Train set: MAPPO, QMIX, IQL; test set: IPPO, VDN.}
            \label{fig:unseen-agent-poam}
            \end{subfigure}
            
            \caption{Out-of-distribution (OOD) evaluation after training, assessing unseen conventions and unseen agent types during training, between learned agents (POAM, SM-POAM and BM-POAM) and uncontrolled agent types (IPPO, VDN, IQL, QMIX and MAPPO). The performance is evaluated by averaging all pairings of $N$ controlled and $M-N$ uncontrolled agents and their corresponding random seeds, called M--N score~\citep{wang2025n}.}
        \label{fig:type-ood}
        \end{figure*}
        \textbf{Answer to Question 3.} We evaluate all baselines alongside our proposed algorithms after training, to test their capability of generalization to unseen conventions and unseen agent types, respectively. The evaluation is conducted by the best model saved during training for each random seed of running experiments. Note that the wide range of confidence intervals is due to that the result averages performance across various number of controlled agents, and we further demonstrate the result per the number of controlled agents in Appendix~\ref{subsec:results-of-n-controlled-agents}. It can be seen from Figure~\ref{fig:unseen-convention-poam} that SM-POAM outperforms both BM-POAM and POAM on all scenarios, except for the 10v11. This implies that in general all Shapley axioms may be instrumental in tackling unseen conventions. As seen from Figure~\ref{fig:unseen-agent-poam}, the performance of BM-POAM is comparable to and even better than SM-POAM in most cases. This implies that the Efficiency axiom could weaken the capability of generalization to unseen agent types. This phenomenon recurs, and becomes even more pronounced, in the evaluation of IPPO-NAHT related algorithms, as shown in Figure~\ref{fig:unseen-agent-ippo} in Appendix~\ref{subsec:extra-results-ood-evaluation}.

    \subsection{Number and Weightings of Basis Games}
    \label{subsec:relationship-between-the-number-and-the-values-of-basis-games}
\begin{figure*}[!ht]
          \centering
          \begin{subfigure}[b]{0.26\textwidth}
            \centering
            \includegraphics[width=\linewidth]{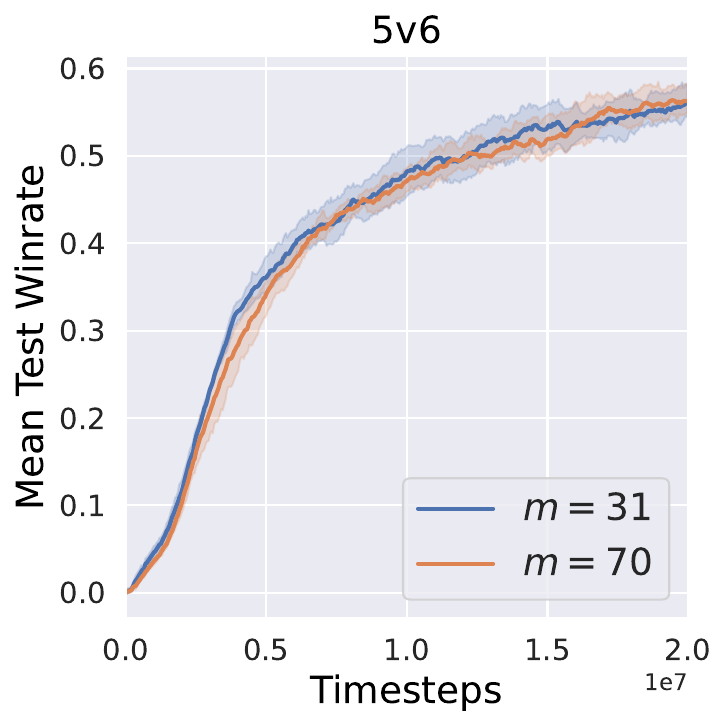}
            \caption{SM-POAM: $m=31$ vs. $m=70$ for $\lambda=0.85$.}
          \end{subfigure}
          ~
          \begin{subfigure}[b]{0.7\textwidth}
            \centering
            \includegraphics[width=\linewidth]{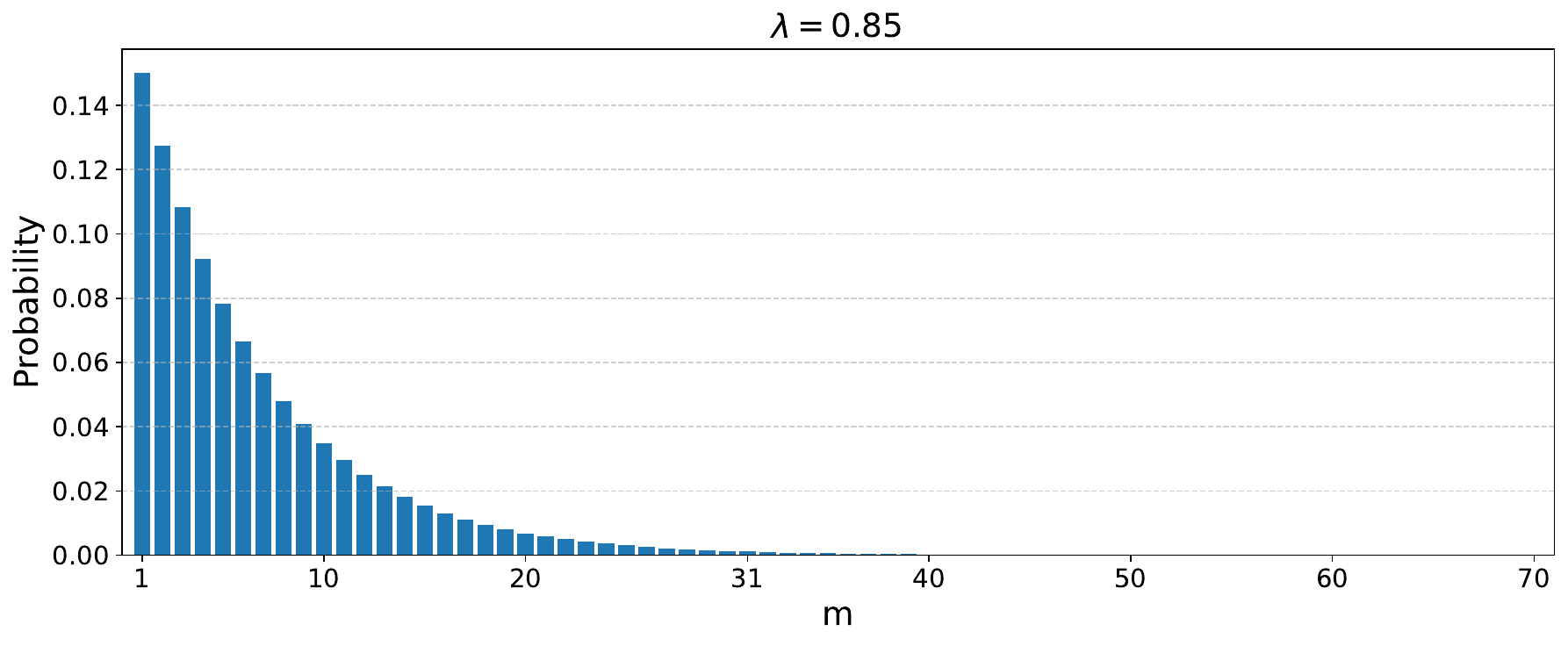}
            \caption{Demonstration of the shape of weightings of basis games with $\lambda=0.85$.}
          \end{subfigure}
        \caption{Controlling the number of effective basis games can be implemented by changing the parameters to control the shape of weightings of basis games. Note that the probability values for $m > 31$ are nearly zero.}
        \label{fig:control-number-of-basis-equal-to-change-values}
        \end{figure*}

        The weightings of basis games ($k_{C}'$ in Eq.~\eqref{eq:td-error-general}) are set by the probability values of a geometric distribution governed by $\lambda$, respecting the convention of RL. $\lambda$ will influence the shape of the geometric distribution. As $\lambda$ increases, the geometric distribution becomes a longer tail with the probability decaying slowly as the number of basis games $m$ increases. In other words, controlling $m$ can be achieved by simply changing the weightings of basis games, or equivalently, the $\lambda$, under the assumption of the geometric distribution. To verify this result in the NAHT, we conduct a case study, as shown in Figure~\ref{fig:control-number-of-basis-equal-to-change-values}. It can be observed that the number of effective basis games (the probability values over which are non-zero) for $m = 31$ and $m = 70$ is the same under $\lambda = 0.85$, resulting in nearly identical learning performance. This aligns with the conventional view of $\lambda$'s impact on TTD($\lambda$) in RL, primarily as a mechanism for controlling the tail length of the return~\citep{sutton2018reinforcement}[Chap. 12.3].
        
\begin{figure*}[!ht]
          \centering
          \begin{subfigure}[b]{0.26\textwidth}
            \centering
            \includegraphics[width=\linewidth]{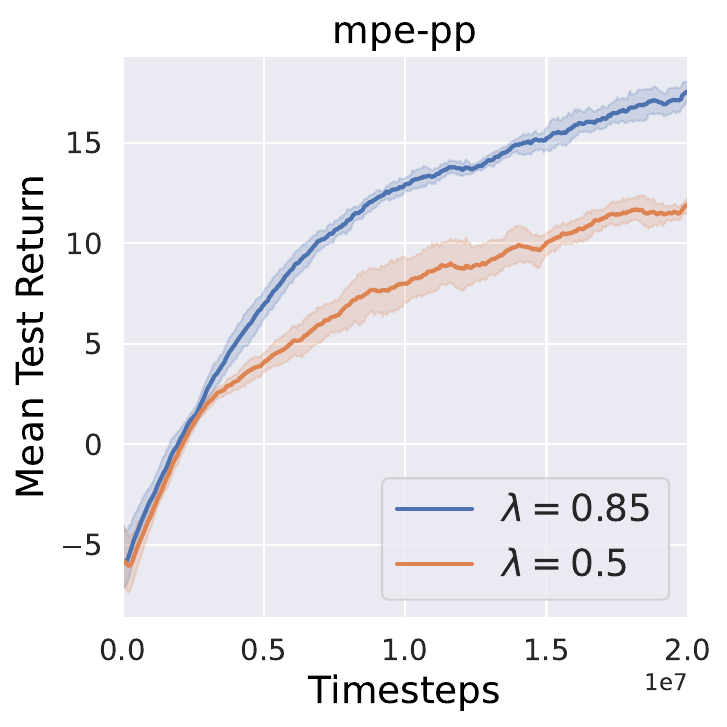}
            \caption{SM-POAM: $\lambda=0.5$ vs. $\lambda=0.85$ for $m=7$.}
            \label{fig:shapley-machine-m-mpe}
          \end{subfigure}
          ~
          \begin{subfigure}[b]{0.64\textwidth}
            \centering
            \includegraphics[width=\linewidth]{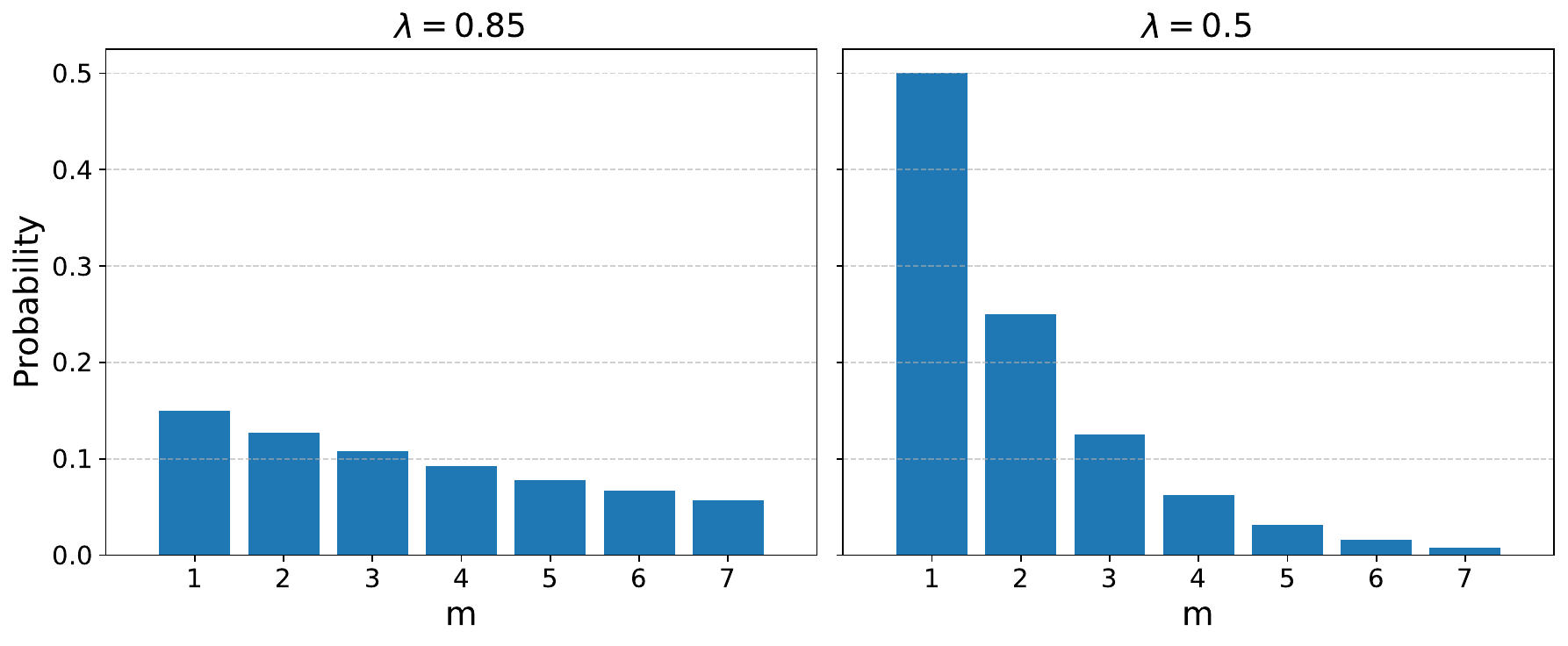}
            \caption{Demonstration of weighting shapes of the basis games with $m=7$, for $\lambda=0.85$ and $\lambda=0.5$, respectively.}
            \label{fig:shapley-machine-m-demo-mpe}
          \end{subfigure}
        \caption{Influence of different values of basis games given the same number of effective basis games $m$. Note that the shapes of geometric distributions vary significantly with different values of $\lambda$, under the same effective number of basis games.}
        \label{fig:different-values-same-number-basis-games}
        \end{figure*}

        In contrast, we demonstrate that even with the same number of basis games, the values of their corresponding weightings significantly influence the learning performance. As shown in Figure~\ref{fig:different-values-same-number-basis-games}, given the same number of active basis games as $m=7$, the learning processes with $\lambda=0.5$ and $\lambda=0.85$ result in very different performance. This provides an alternative insight into how weightings influence the TTD($\lambda$), different from the traditional RL perspective.
        
\section{Discussion, Limitations and Future Work}
\label{sec:conclusion}
    We studied reinforcement learning (RL) in open multi-agent systems through the lens of the n-agent ad hoc teamwork (NAHT), and introduced a cooperative game-theoretic framework that makes explicit how openness is distinguished from the cooperative multi-agent reinforcement learning (CMARL). Central to our approach is the NAHT decomposition, which separates internal cooperation among controlled agents from external teamwork with uncontrolled agents. Building on this structure, we proposed an axiomatic framework that derives value function learning from the Shapley axioms (Efficiency, Symmetry, and Linearity) by enforcing them as structural constraints. This perspective shifts the design of RL algorithms from heuristic constructions to principled designs in the NAHT. Since CMARL is an extreme case of the NAHT, our framework also works on it.

    Our axiomatic framework reveals a novel connection between cooperative game theory and reinforcement learning: The Linearity axiom provides a theoretical explanation for the $n$-step return methods and the TTD($\lambda$) in the NAHT. This result is coincidentally related to the latest finding in single-agent offline RL that the $n$-step return methods with selected horizons can improve scalability to the increasing size of datasets~\citep{ParkFMEKL25}. As the NAHT encompasses the case of single-agent RL with other unknown agents, our framework suggests that, through the TTD($\lambda$), the increasing size of datasets could be a promising way to address the NAHT, which is left for future work. 
    
    More broadly, the axiomatic perspective suggests new opportunities for exploring richer classes of weighting schemes beyond the standard geometric forms in temporal-difference learning for the NAHT (as demonstrated in Section~\ref{subsec:relationship-between-the-number-and-the-values-of-basis-games}). For example, weightings $k_{C}$ satisfying $\sum (k_{C})^2=1$ as shown in Figure~\ref{fig:state-space-cooperative-game-case} could define alternative weighting schedules. Because the NAHT problem can be reduced to the special case of single-agent RL with other uncontrolled agents, with an appropriate connection between the world model (transition function) and the concept of other uncontrolled agents, this result can be relaxed to the broader single-agent RL.

    A key limitation of our work is that it is still unclear about how to deal with the scenario where the predefined largest horizon (the episode length) is far less than the number of non-empty coalition basis games.  Our current horizon assignment is $n_i = i$, as clarified in Appendix~\ref{subsec:links-to-rl}. Our tentative strategy to mitigate this shortcoming is approximating the number of basis games as the episode length. In future work, this can be further investigated by changing our current horizon assignment to another one, e.g., $n_i = f(i, s)$, where $f(i, s) \in \mathbb{N}$ is a function of states $s$ and coalition indices $i$, based on Definition~\ref{def:powerset-enum}. 

\section*{Acknowledgements}
Jianhong Wang and Jonathan Lawry are supported by the Engineering and Physical Sciences Research Council (EPSRC)
[Grant Ref: EP/Y028732/1]. Samuel Kaski is supported by UKRI Turing AI World-Leading Researcher Fellowship, EP/W002973/1 and EU, ERC ODD-ML 101201120.
\bibliography{main}
\bibliographystyle{plainnat}
\newpage
\appendix

\section{Related Work}
\label{sec:related-work}
    \subsection{TD($\lambda$) in Reinforcement Learning and Multi-Agent Reinforcement Learning}
        Related work on temporal-difference learning with eligibility traces centers on the seminal TD($\lambda$) algorithm proposed by~\cite{sutton1988learning}, which unified one-step TD and Monte Carlo methods via a trace-decay parameter $\lambda$ and introduced the forward- and backward-view equivalence for efficient multi-step credit assignment. Subsequent theoretical analyses~\citep{dayan1992convergence,dayan1994td,tsitsiklis1996analysis} established convergence guarantees and characterized the bias-variance trade-off inherent in $\lambda$-returns, while extensions by~\cite{watkins1989learning,peng1994incremental} adapted eligibility traces for off-policy control, Q($\lambda$). Advances in function approximation, including true online TD($\lambda$)~\citep{seijen2014true} and gradient-TD methods, further broadened applicability to large-scale and nonlinear settings, inspiring $n$-step return techniques such as generalized advantage estimation (GAE)~\citep{schulman2015high} for policy optimization. In the multi-agent reinforcement learning (MARL), algorithms of PPO family such as MAPPO~\citep{yu2022surprising} and IPPO~\citep{de2020independent}, applied GAE to optimize policies and thus used the TD($\lambda$) as the target values to train critics. Our work highlights the theoretical link between the truncated TD($\lambda$), denoted as TTD($\lambda$)~\citep{cichosz1994truncating}, in the NAHT and the Linearity axiom of the Shapley value. It further shows that the number of $n$-step return components in the TTD($\lambda$) can be estimated by the number of non-empty coalition basis games. This insight provides a theoretical foundation that can inform the design of a broader class of MARL algorithms.

    \subsection{Theoretical Models for Ad Hoc Teamwork}
        We now discuss theoretical models describing the ad hoc teamwork (AHT). \cite{brafman1996partially} pioneered research into the AHT by investigating repeated matrix games involving a single teammate. Subsequent studies expanded this framework to scenarios with multiple teammates, notably by~\cite{agmon2012leading}. Later, \cite{agmon2014modeling} further relaxed earlier assumptions by allowing teammates' policies to be selected from a known set. \cite{stone2010teach} initially formalized the AHT through collaborative multi-armed bandits, albeit under notable assumptions such as prior knowledge of teammates' policies and environmental conditions. \cite{albrecht2013game} advanced this field significantly by introducing the stochastic Bayesian game (SBG), the first comprehensive theoretical framework accommodating dynamic environments and unknown teammate behaviours in the AHT. Building on the SBG, \cite{rahman2021towards} proposed the open stochastic Bayesian game (OSBG), addressing the open ad hoc teamwork (OAHT). \cite{ZintgrafDCWH21} modelled the AHT through interactive Bayesian reinforcement learning (IBRL) within Markov games, specifically targeting non-stationary teammate policies within single episodes. \cite{xie2021learning} introduced the hidden parameter Markov decision process (HiP-MDP) to handle situations where teammates' policies vary across episodes but remain stationary during individual episodes. Most recently, \cite{wang2024open} extended the OSBG by incorporating principles from cooperative game theory, introducing the open stochastic Bayesian coalitional affinity game (OSBG-CAG), which theoretically justifies a graph-based representation for joint Q-value functions and includes rigorous convergence proofs for Q-learning algorithms in open team settings. Our work adopts the perspective of cooperative game theory, introducing an NAHT decomposition based on state-specific cooperative games to address the NAHT. This well captures the structural difference between the CMARL and the NAHT. Based on the NAHT decomposition, we derive an axiomatic framework to address the NAHT, which introduces a novel perspective for solving this problem. The key difference in assumptions between the work by \cite{wang2024open} and our work is that the work by \cite{wang2024open} was built on the non-transferable utility games, while our work is built on the transferable utility games. For more details about the difference between these two cooperative game models, please refer to~\cite{chalkiadakis2011computational}.
    
    \subsection{Shapley Value in Multi-Agent Reinforcement Learning}
        Early studies~\citep{wang2020shapley} incorporated the Shapley value into credit assignment scheme in a principled way and proposed an algorithm named as SQDDPG, underpinned by the equivalence between cooperative-game theoretical models in dynamic scenarios and the team reward Markov game. \cite{wang2022shaq} further improved the theory by proving the existence of the Shapley value under the team reward Markov game through construction, and proposed an algorithm, relying on the construction, named as SHAQ --- an algorithm with promising convergence to find an optimal joint policy. It also shed light on the relationship between the Shapley value and the relevant value decomposition and credit assignment approaches for the CMARL. \cite{li2021shapley} improved the stability of SQDDPG and proposed an algorithm named Shapley Counterfactual Credits. \cite{han2021multiagent} incorporated the Shapley value into the model-based PPO as each agent's advantage value, through the estimated coalition values by trained transition and reward models. \cite{chai2024aligning} continued the idea from~\cite{han2021multiagent}, by replacing the transition and reward models with more powerful world models. On the other hand, \cite{xue2022efficient} incorporated the Shapley value into multi-agent communication, as a criterion for forming communication between agents. One of our work's aims is learning the Shapley value to address the NAHT, a generalized problem to the CMARL. More importantly, the Shapley value is learned by satisfying its axioms, rather than by constructing their explicit formulas as implemented in the previous work. This may help overcome the notorious learning instability issues caused by learning substantial ``coalitional values'' in the existing algorithms based on the Shapley value. Furthermore, the insight of axiomatic characterization can facilitate improving the flexibility of the Shapley value approaches in a wide variety of tasks. Specifically, one can safely drop any part of a learning algorithm, if it is justified that the targeting task does not align with the corresponding axiom. 
        
\section{Extra Background}
\label{sec:additional-background}
    \subsection{$\lambda$-Return and TD($\lambda$)}
    \label{subsec:lambda-return}
        We now introduce an extension of return named $\lambda$-return. Mathematically, a $\lambda$-return $G_{t}^{\lambda}$ for infinite-horizon cases can be expressed as follows:
        \begin{equation}
            \begin{split}
                G_{t:t+n} = R_{t} + \gamma R_{t+1} + \cdots + \gamma^{n-1} R_{t+n-1} + \gamma^{n} v_{\pi}(s_{t+n}),\\
                G_{t}^{\lambda} = (1 - \lambda) \sum_{n=1}^{\infty} \lambda^{n-1} G_{t:t+n}.
            \end{split} 
        \end{equation}
        Here, $v_{\pi}$ denotes the state-value function under policy $\pi$. Similarly, the $\lambda$-return for finite-horizon cases can be expressed as follows:
        \begin{equation}
            G_{t}^{\lambda} = (1 - \lambda) \sum_{n=1}^{T-t} \lambda^{n-1} G_{t:t+n} + \lambda^{T-t} G_{t:T}.
        \end{equation}
        Note that if $\lambda=1$, updating according to the $\lambda$-return is a Monte Carlo algorithm. In contrast, if $\lambda=0$, then the $\lambda$-return reduces to $G_{t:t+1}$, the one-step return. The Monte Carlo algorithm is known as its high variance but low bias, while one-step return is known as its low variance but high bias. For this reason, the change of $\lambda$ can be seen as a tradeoff between bias and variance. The TD prediction using $\lambda$-return as the target value is referred to as TD($\lambda$)~\citep{sutton1988learning}.

        A variant of TD($\lambda$) by shaping $G_{t}^{\lambda}$ with a designated horizon $H \in \{t+1, ..., T\}$ is referred to as truncated TD($\lambda$), shortened as TTD($\lambda$)~\citep{cichosz1994truncating}. Mathematically, $G_{t}^{\lambda}$ with a designated horizon $H$ can be formulated as follows:
        \begin{equation}
            G_{t}^{\lambda,H} = (1 - \lambda) \sum_{n=1}^{H-t} \lambda^{n-1} G_{t:t+n} + \lambda^{H-t} G_{t:H}.
        \end{equation}

        Note that in the convention of RL, the horizon $H$ is defined as the time difference from the starting point of an episode, denoted by $t=0$. 
        
        \begin{remark}
            In this paper, to simplify the presentation of our work, we redefine the horizon as the time difference from each timestep $t$, denoted as $m$, such that $m=H-t$. With the new definition of horizon, we can rewrite the above formula of $G_{t}^{\lambda,H}$ as $G_{t}^{\lambda,m}$, as follows:
            \begin{equation}
                \begin{split}
                    G_{t}^{\lambda,m} &= (1 - \lambda) \sum_{n=1}^{m} \lambda^{n-1} G_{t:t+n} + \lambda^{m} G_{t:t+m} \\
                    &=(1-\lambda)\sum_{n=1}^{m-1}\lambda^{n-1}G_{t:t+n} +\big[(1-\lambda)\lambda^{m-1}+\lambda^m\big]G_{t:t+m} \\
                    &=(1-\lambda)\sum_{n=1}^{m-1}\lambda^{n-1}G_{t:t+n} +\lambda^{m-1}G_{t:t+m}.
                \end{split}
            \end{equation}
        \end{remark}

    \subsection{Superadditive Game and Cooperative Multi-Agent Reinforcement Learning}
    \label{subsec:superadditive-game-cmarl}
        The superadditive game is a subclass of characteristic games that satisfies an additional condition: $v(C \cup D) \geq v(C) + v(D)$, for any two distinct coalitions $C,D \subseteq \mathcal{M}$ with $C \cap D = \emptyset$. Previous work in the CMARL~\citep{wang2020shapley} has shown the equivalence on the task objective function between the state-based superadditive games with action spaces and the team reward Markov games with coalition structures.\footnote{In \cite{wang2020shapley}, state-based convex games with action spaces were proved to be equivalent to team reward Markov games. However, setting $C \cap D = \emptyset$ can relax a convex game to a superadditive game.} Intuitively, the Shapley value is a solution to the stability of a team formation, resembling cooperation. Since the NAHT is a generalization of the CMARL, this conclusion is still valid in principle. It has been already proved that the Shapley value exists in superadditive games~\citep{shapley1953value} and each game instance in superadditive games can be uniquely represented as $\sum k_{C} v_{C}^{1}$, where $k_{C} \geq 0$~\citep{dubey1975uniqueness}. In cooperative game theory, any game can be transformed to a superadditive game if its value $v$ is transformed by a \textit{superadditive cover} $v^{*}$ such that $v^{*}(C) = \max_{CS(C)} \sum_{D \in CS(C)} v(D)$~\citep{greco2011complexity}. As a result, it will not lose generality if we consider the cooperative game space as superadditive games in our work.

    \subsection{Axioms of Shapley Values}
    \label{subsec:axioms-of=shapley-value}
        Efficiency and Symmetry have been well investigated in the literature of multi-agent reinforcement learning~\citep{wang2020shapley,li2021shapley,han2021multiagent,wang2022shaq,chai2024aligning}, but Additivity still lacks attention. In detail, Additivity means that $\phi(w_{1}+w_{2}) = \phi(w_{1}) + \phi(w_{2})$, for any cooperative games $w_{1},w_{2} \in \mathcal{G}$. If we consider $m$ possible games, then we have the following expression: $\phi(\sum_{i=1}^{m} w_{i}) = \sum_{i=1}^{m} \phi(w_{i})$, where $w_{1}, w_{2}, ..., w_{m} \in \mathcal{G}$ and the sum $\sum_{i=1}^{m} w_{i}$ can be seen as a game reproduced by the above games. Linearity is a stronger condition than Additivity~\citep{dubey1975uniqueness,young1985monotonic}. Specifically, Linearity requires $\phi(\sum_{i=1}^{m} \alpha_{i} w_{i}) = \sum_{i=1}^{m} \alpha_{i} \phi(w_{i})$, for all $\alpha_{i} \in \mathbb{R}$. In contrast, Additivity only requires to satisfy the above equation when $\alpha_{i} = 1$. In our work, we use the Linearity axiom to establish the Additivity axiom, to adapt to our derivation.
    
    \subsection{Baseline Algorithms: POAM and IPPO}
    \label{subsec:poam}
        \textbf{POAM.} In general, POAM is an algorithm built upon IPPO~\citep{de2020independent}. The primary difference between POAM and IPPO is as follows: For each agent's policy and value, in addition to its observation used in IPPO, POAM takes as input an embedding vector to predict other agents' potential behaviours. More specifically, this embedding vector is trained by an encoder-decoder structure, wherein the encoder is modeled as RNNs and decoder is modeled as MLPs. For brevity, we use $-i$ to denote the set of all agents excluding agent $i$. Let $h_{t,i} =\{o_{k,i}, a_{k-1,i}\}_{k=1}^{t}$ denote agent $i$'s history of observations and actions up to timestep $t$ and $e_{t,i} \in \mathbb{R}^{n}$ denote the resulting embedding vector of $n$ dimensions. The encoder parameterized by $\theta^{e}$ is defined as $f_{\theta^{e}}^{enc}: \mathcal{H}_{i} \rightarrow \mathbb{R}^{n}$. The embedding vector is decoded by two decoder networks: the observation decoder $f_{\theta^{o}}^{dec}: \mathbb{R}^{n} \rightarrow \mathcal{O}_{-i}$, and the action decoder $f_{\theta^{a}}^{dec}: \mathbb{R}^{n} \rightarrow \Delta(\mathcal{A}_{-i})$. The decoder networks are respectively trained to predict the observations and actions of all other agents in the team at timestep $t$, $o_{t,-i}$ and $a_{t,-i}$, to encourage $e_{t,i}$ to contain information about collective behaviors corresponding to $h_{t,i}$. While the observation decoder directly predicts the observed $-i$'s observations, the action decoder predicts the parameters of a probability distribution over the $-i$'s actions $\pi_{-i}(a_{t,-i}; f_{\theta^{a}}^{dec}(f_{\theta^{e}}^{enc}(h_{t,i})))$. As we consider continuous observations and discrete actions, the loss function with using categorical distribution to model $\pi_{-i}$ is as follows:
        \begin{equation}
            L_{\theta^{e}, \theta^{o}, \theta^{a}}(h_{t,i}, o_{t,-i}, a_{t,-i}) = || f_{\theta^{o}}^{dec}(f_{\theta^{e}}^{enc}(h_{t,i})) - o_{t,-i} ||^{2} - \log \pi_{-i}(a_{t,-i}; f_{\theta^{a}}^{dec}(f_{\theta^{e}}^{enc}(h_{t,i}))).
        \label{eq:loss-encoder-decoder}
        \end{equation}
        
        For brevity, we define $g(h) := (h, f_{\theta^{e}}^{enc}(h))$ in the following description.
        
        POAM employed generalized advantage estimation (GAE)~\citep{schulman2015high} to form the policy gradient, and the value function used to form GAE is trained by the following loss function:
        \begin{equation}
            L_{\theta^{c}}(h_{t,i}) = \frac{1}{2} \left( V_{i}^{\theta^{c}}(g(h_{t,i})) - \hat{V}_{t,i} \right)^{2},
        \label{eq:loss-critic}
        \end{equation}
        where $\hat{V}_{t,i}$ is the finite-horizon TD($\lambda$) return, for which the horizon is up to the episode length $T$. Each agent $i$'s GAE estimation is as follows:
        \begin{equation}
            A_{t,i} = \sum_{l=0}^{T} (\gamma \lambda)^{l} \delta_{t+l,i},
        \end{equation}
        where $\delta_{t,i} = R_{t} + \gamma V_{i}(g(h_{t,i})) - V_{i}^{\theta^{c}}(g(h_{t,i}))$ and $V_{i}(g(h_{t,i}))$ denotes the non-parameterized target individual value. Based on the GAE defined above, the policy optimization loss is defined as
        \begin{equation}
            L_{\theta}(h_{t,i}, a_{t,i}) = \min \left\{ \frac{\pi_{\theta}(a_{t,i} | g(h_{t,i}))}{\pi_{\theta_{old}}(a_{t,i} | g(h_{t,i}))} A_{t,i}, \text{clip}\left( \frac{\pi_{\theta}(a_{t,i} | g(h_{t,i}))}{\pi_{\theta_{old}}(a_{t,i} | g(h_{t,i}))}, 1 - \epsilon, 1 + \epsilon \right) A_{t,i}\right\}.
        \label{eq:loss-policy}
        \end{equation}

        \textbf{IPPO.} By removing the loss for the encoder-decoder structure as shown in Eq.~\eqref{eq:loss-encoder-decoder} and replacing the input $g(h_{t,i})$ by $h_{t,i}$ in the above set of equations, we get the loss functions of IPPO.

\section{Preliminary Results}
\label{subsec:a-preliminary-result}
    \begin{lemma}
    \label{lemm:subgame-transform}
        Given a fixed state $s \in {\cal S}$, each state-specific cooperative game $v_{s} \in \mathcal{G}(s)$ can be uniquely represented by $v_{s} = \sum_{\emptyset \neq C \subseteq \mathcal{M}} k_{C}' \cdot v_{C,s}^{z}$, where $k_{C}' = \frac{k_{C}}{z}$ and $z \neq 0$.
    \end{lemma}
    \begin{proof}
        We can represent each $k_{C}$ in an equivalent form as $\frac{k_{C} \cdot z}{z}$, where $z \neq 0$. Substituting this term into the formula $v_{s} = \sum_{\emptyset \neq C \subseteq \mathcal{M}} k_{C} v_{C,s}^{1}$ in Definition~\ref{def:generalised-basis-games}, we can get the following formula such that
        \begin{equation*}
            v_{s} = \sum_{\emptyset \neq C \subseteq \mathcal{M}} \frac{k_{C} \cdot z}{z} \cdot v_{C, s}^{1}.
        \end{equation*}
        Since $v_{C, s}^{z} = z \cdot v_{C, s}^{1}$ by Definition~\ref{def:generalised-basis-games}, we have the following formula such that
        \begin{equation*}
            v_{s} = \sum_{\emptyset \neq C \subseteq \mathcal{M}} \frac{k_{C}}{z} \cdot v_{C, s}^{z}.
        \end{equation*}
    \end{proof}
    \begin{proposition}
    \label{prop:positive-Kc}
        For the class of superadditive games formed by a set of state-specific basis games $\{v^z_{C,s} | \emptyset \neq C \subseteq \mathcal{M} \}$, such that $v_{s} = \sum_{\emptyset \neq C \subseteq \mathcal{M}} k_{C}' \cdot v_{C,s}^{z}$, it holds that $k_{C}' \geq 0$, for all $\emptyset \neq C \subseteq \mathcal{M}$.
    \end{proposition}
    \begin{proof}
        We first consider a cooperative game space $\mathcal{G}$ (a broader space of superadditive games). As per~\cite{dubey1975uniqueness}, a game $v: \mathcal{G} \rightarrow \mathbb{R}_{\geq 0}$ belonging to $\mathcal{G}$ can be uniquely represented as follows: $$v = \sum_{\emptyset \neq C \subseteq \tilde{N}} k_{C} v_{C}^{1}.$$ The analytic form of $k_{C}$ under $\mathcal{G}$ is represented as follows:
        
        \begin{equation*}
            k_{C} = \sum_{T \subseteq C} (-1)^{|C| - |T|} v(T).
        \end{equation*}
        
        The condition for $v$ to be a superadditive game is as follows:
        \begin{equation*}
            v(T_{1} \cup T_{2}) \geq v(T_{1}) + v(T_{2}), \qquad T_{1}, T_{2} \subseteq C.
        \end{equation*}
        Following the result from~\cite{dubey1975uniqueness}, if the above condition holds, $k_{C} \geq 0$ has to be satisfied.\footnote{For example, if $C=\{1,2\}$, then $T = \emptyset, \{1\}, \{2\}, \{1,2\}$. As a result, $k_{C} = v(\emptyset)- v(\{1\}) - v(\{2\}) + v(\{1,2\})$, where $v(\emptyset) = 0$ by default. If $v$ is assumed to be a superadditive game, $k_{C} \geq 0$ has to hold.}
        
        We now extend the $\mathcal{G}$ to a state-specific cooperative game space $\mathcal{G}(s)$, for a fixed state $s \in \mathcal{S}$. Since the $\mathcal{G}(s)$ is isomorphic to a game space $\mathcal{G}$, the characteristics satisfied in the $\mathcal{G}$ also holds in the $\mathcal{G}(s)$. Therefore, $v_{s}: \mathcal{G}(s) \rightarrow \mathbb{R}_{\geq 0}$ can be uniquely represented as $v_{s} = \sum_{\emptyset \neq C \subseteq \mathcal{M}} k_{C} v_{C,s}^{1}$. As we consider $\mathcal{G}(s)$ as a space of superadditive games, $k_{C} \geq 0$ holds.

        By Lemma~\ref{lemm:subgame-transform}, we have $v_{s} = \sum_{\emptyset \neq C \subseteq \mathcal{M}} k_{C}' \cdot v_{C,s}^{z}$, where $k_{C}' = \frac{k_{C}}{z}$ and $z \neq 0$. This implies that any set of basis games in the form $\{v^z_{C,s} | \emptyset \neq C \subseteq \mathcal{M} \}$ can form a state-specific cooperative game belonging to superadditive games. The proof is completed.
    \end{proof}

    \begin{remark}
    \label{rmk:reason-for-k-greater-than-zero}
        Following discussion in Section~\ref{subsec:shapley-value}, we aim to restrict the state-specific cooperative game class to the superadditive game. By Proposition~\ref{prop:positive-Kc}, we can conclude that $k_{C}' \geq 0$ is a sufficient condition for $v_{s}$ belonging to the superadditive game class. In other words, $k_{C}' \geq 0$ is a key condition for reaching cooperation within a team of agents. However, $k_{C}' = 0$ is a singularity condition, which leads to 0 for all state-specific cooperative games in values, with no actual meaning in practice. For this reason, we only choose $k_{C}' > 0$ as the condition in our work.
    \end{remark}

\section{Introducing Markovian Dynamics and Rewards to Game Structures}
\label{subsec:links-to-rl}
    Based on the game structure for the NAHT in Definition~\ref{def:ad-hoc-teamwork-in-cooperative-game}, we now introduce Markovian dynamics to establish the link between the state-specific cooperative game values of consecutive states, aligned with the environmental setting in Dec-POMDPs~\citep{oliehoek2016concise}. 

    The only postulate is that the state-specific cooperative game value for each state $s_{t}$ can be represented in the form of rewards: $v_{s_{t}}(\mathcal{M}) = \mathbb{E}_{\pi}[\sum_{\tau=0}^{T-t-1} \gamma^{\tau} R_{t+\tau} \ | \ s_{t}]$. By the convention of RL~\citep{sutton2018reinforcement}[Chap. 7], it is feasible to use the expected $n$-step return to express the $v_{s_{t}}(\mathcal{M})$ under a given horizon $n$ to define the transformation, such that
    \begin{equation}
        \begin{split}
            v_{s_{t}}(\mathcal{M}) = \mathbb{E}_{\pi} \left[ \sum_{\tau=0}^{n-1} \gamma^{\tau} R_{t+\tau} + \gamma^{n} v_{s_{t+n}}(\mathcal{M}) \ \middle\vert \ s_{t} \right] := \mathbb{E}_{\pi} \left[ G_{t:t+n} \ \middle\vert \ s_{t} \right].
        \end{split}
    \label{eq:game-value-expansion}
    \end{equation}
    
    For brevity, we denote the state-specific basis game value $v_{C, s}^{v_{s}(\mathcal{M})}(\mathcal{M})$ by $V(C, s)$ for all $C \subseteq \mathcal{M}$. Then, we get the following fact:
    \begin{fact}
    \label{fact: transform}
        $V(C_{1}, s) = V(C_{2}, s) = \cdots = V(C_{m}, s) = v_{s}(\mathcal{M})$ for all $C_{i} \in \mathcal{P}_{+}(\mathcal{M})$, as per Definition~\ref{def:game-set} and $\mathcal{M} \supseteq C_{i}$, where $\mathcal{P}_{+}(\mathcal{M})$ indicates the powerset of $\mathcal{M}$ excluding the empty set.
    \end{fact}

    \begin{definition}[\textbf{Coalition Ordering by Size and Horizon Assignment}]
    \label{def:powerset-enum}
        We first set $m:=2^M-1$. Given a coalition enumeration $(C_1,\dots,C_m)$ satisfying (1) $|C_i|\le |C_{i+1}|$ for all $i$ (non-decreasing in size); (2) within each size, ties are broken by a fixed total order (e.g., lexicographic ordering over agent indices), we can define a sequence of non-decreasing horizons $(n_1,\dots,n_m)$ such that $n_{i} \geq n_{i-1}$ for all $i \ge 2$ and $n_1:=1$. Accordingly, we represent the basis game values $V(C_i,s)$ as their corresponding expected $n$-step returns:
        \[
        V(C_i,s)\;:=\;\mathbb E_\pi\!\big[G_{t:t+n_{i}}\mid s_t=s\big],\qquad (i=1,\dots,m).
        \]
    \end{definition}
    
    We motivate the horizon assignment in Definition~\ref{def:powerset-enum} using the variance comparison in Appendix~\ref{sec:justification-of-size-lex}. Under the stated assumptions and for a fixed state $s$, the basis game value estimators $\widehat V(C_i,s)$ and the sampled $n_i$-step returns $G_{t:t+n_i}$ have variances that are both non-decreasing in $i$.
    This definition concerns the following situation of coalition ordering: When the number of agents increases (i.e., $C_{i}$ becomes larger in size), more diverse agent types and behaviours emerge. As a result, the estimator of the corresponding $V(C_{i}, s)$ has a higher variance. 
    \begin{proposition}
    \label{prop:powerset-identity}
        By Definition~\ref{def:powerset-enum}, with the horizon assignment $n_i=i$, for every state $s\in\mathcal S$,
        \begin{equation*}
            \begin{split}
                \big(V(C_1,s), \dots, V(C_m,s)\big) \;=\; \big(\mathbb E_\pi[G_{t:t+1}\mid s],\dots,\mathbb E_\pi[G_{t:t+m}\mid s]\big),
            \end{split}
        \end{equation*}
        and, each $V(C_i,s)$ is equal to $v_{s}(\mathcal{M})$ in value.
    \end{proposition}
    \begin{proof}
        The result can be immediately obtained from the definition of $V(C_i,s)$ in Definition~\ref{def:powerset-enum}, and the statement that each $V(C_i,s)$ is equal to $v_{s}(\mathcal{M})$, directly following Fact~\ref{fact: transform}.
    \end{proof}
        
\section{Justification of Coalition Ordering by Size and Horizon Assignment}
\label{sec:justification-of-size-lex}
    \subsection{Sketch of Justification Procedure}
        In this section, we will justify the insight behind the coalition ordering by size and horizon assignment, as shown in Definition~\ref{def:powerset-enum}. We fix a policy $\pi$ and use its state-value function $v_{\pi}:\mathcal{S}\to\mathbb{R}$ as the bootstrap value function, as in Section~\ref{subsec:lambda-return}. The function $v_{\pi}$ is held fixed during trajectory sampling. We then define a discrete-time stochastic process of TD error terms $(Z_{n+1})_{n \in \mathbb{N}_{0}}$:
        \[
        Z_{n+1}\;:=\;R_{t+n}+\gamma v_{\pi}(s_{t+n+1})-v_{\pi}(s_{t+n}).
        \]
        As a result, the $n$-step return $G_{t:t+n}$, as an estimator of the expected $n$-step return, has the following recursive relationship:
        \[
        G_{t:t+n+1}=G_{t:t+n}+\gamma^{n}Z_{n+1}.
        \]
        We first show in Lemma~\ref{lemm:variance-increment} that, given $s_t=s$, the variance of the $n$-step return $\text{Var}(G_{t:t+n}\mid s)$ is non-decreasing in a monotonically increasing horizon assignment when the TD-error sequence $(Z_n)$ has non-negative autocovariances. Specifically, extending the horizon from $n$ to $n+1$ adds the discounted TD-error term $\gamma^n Z_{n+1}$, whose covariance with the existing return is non-negative. The return variance therefore increases strictly when this additional term has positive conditional variance and remains unchanged when it has zero conditional variance. Then, we make the coalition-variance monotonicity assumption in Assumption~\ref{assm:coalition-variance-monotonicity}, which can be justified by Remark~\ref{rmk:satification-of-coalition-variance-monotonicity} with evidence from the definition of the game structure for the NAHT and Remark~\ref{rmk:verification-of-coalition-variance-monotonicity} with evidence from a verification experiment. Briefly, Assumption~\ref{assm:coalition-variance-monotonicity} concerns the following situation: When the number of agents increases (i.e., $C_{i}$ becomes larger in size), more diverse agent types and behaviours emerge. As a result, the estimator of the corresponding $V(C_{i}, s)$ has a higher variance.
        
        For the state-specific basis-game representation, we identify the bootstrap value with the team value, such that $v_{\pi}(s)=v_s(\mathcal{M})$. By Lemma~\ref{lemm:variance-increment} and Assumption~\ref{assm:coalition-variance-monotonicity}, we prove in Proposition~\ref{prop:variance-strat} that there exist non-decreasing sequences of variances for both the sequence of estimators of $V(C_{i}, s)$ under non-decreasing coalitions in size and the sequence of $n$-step returns $G_{t:t+n_i}$ under a non-decreasing horizon assignment. This motivates us to define the following term under this monotone-variance regularity:
         \[
        V(C_i,s)\;:=\;\mathbb E_\pi\!\big[G_{t:t+n_i}\mid s_t=s\big],\qquad (i=1,\dots,m).
        \]
        In other words, the $n$-step return $G_{t:t+n_i}$ is defined as an unbiased estimator for the basis game value $V(C_i,s)$ for each coalition $C_i$. The justification of Definition~\ref{def:powerset-enum} is completed.


    \subsection{Theoretical and Empirical Justification}
        \begin{lemma}
        \label{lemm:variance-increment}
            If $(Z_n)$ has non-negative autocovariances given $s_t=s$, then $n\mapsto \mathrm{Var}(G_{t:t+n}\mid s)$ is non-decreasing. Also, $\mathrm{Var}(G_{t:t+n+1}\mid s) = \mathrm{Var}(G_{t:t+n}\mid s)$ if and only if $\mathrm{Var}(Z_{n+1}\mid s)=0$.
        \end{lemma}
        \begin{proof}
            Use $G_{t:t+n+1}=G_{t:t+n}+\gamma^{n}Z_{n+1}$ and expand the variance conditioning on $s_{t}=s$, we can get the following formula:
            \[
            \mathrm{Var}(G_{t:t+n+1}\mid s)
            =\mathrm{Var}(G_{t:t+n}\mid s)
            +2\gamma^{n}\mathrm{Cov}\!\big(G_{t:t+n},Z_{n+1}\mid s\big)
            +\gamma^{2n}\mathrm{Var}(Z_{n+1}\mid s).
            \]
    
            \textbf{We first prove that if $\boldsymbol{(Z_n)}$ has non-negative autocovariances given $\boldsymbol{s_t=s}$, then $\boldsymbol{n\mapsto \mathrm{Var}(G_{t:t+n}\mid s)}$ is non-decreasing, as follows:}
            
            Using the telescoping sum of TD errors~\citep{sutton2018reinforcement}[Chap. 6], we have the following formula:
            \[
            G_{t:t+n} - v_{\pi}(s_{t}) = \sum_{k=0}^{n-1} \gamma^{k} Z_{k+1}.
            \]
    
            Since $v_{\pi}(s_t)=v_{\pi}(s)$ is constant given $s_t=s$, linearity of covariance yields the following equation:
            \[
            \mathrm{Cov}\!\big(G_{t:t+n},Z_{n+1}\mid s\big) = \sum_{k=0}^{n-1} \gamma^k \mathrm{Cov}(Z_{k+1}, Z_{n+1} \mid s).
            \]
    
            Therefore, if all autocovariances $\mathrm{Cov}(Z_i, Z_j\mid s)\ge 0$, then $\mathrm{Cov}\!\big(G_{t:t+n},Z_{n+1}\mid s\big)\geq 0$.
            
            Since $\mathrm{Var}(Z_{n+1}\mid s) \ge 0$ must hold, we obtain the following inequality:
            \[
            \mathrm{Var}(G_{t:t+n+1}\mid s) \ge \mathrm{Var}(G_{t:t+n}\mid s).
            \]
    
            That is, $n\mapsto \mathrm{Var}(G_{t:t+n}\mid s)$ is non-decreasing.
    
            \textbf{We next prove $\boldsymbol{\mathrm{Var}(G_{t:t+n+1}\mid s) = \mathrm{Var}(G_{t:t+n}\mid s)}$ if and only if $\boldsymbol{\mathrm{Var}(Z_{n+1}\mid s)=0}$, as follows:}
            
            Given a fact that $\mathrm{Var}(G_{t:t+n+1}\mid s)=\mathrm{Var}(G_{t:t+n}\mid s)$ if and only if the following condition holds: 
            \[
            \mathrm{Var}(Z_{n+1}\mid s)=0,\qquad \mathrm{Cov}\!\big(G_{t:t+n},Z_{n+1}\mid s\big)=0,
            \]
            
            we now aim to prove that $\mathrm{Var}(Z_{n+1}\mid s)=0$ is sufficient for $\mathrm{Cov}\!\big(G_{t:t+n},Z_{n+1}\mid s\big)=0$, and the proof is completed:
    
            If a non-negative random variable (r.v.) $W \ge 0$ satisfies $\mathbb{E}\left[ W \mid s\right]=0$, then $W=0$ almost surely (a.s.) given $s$~\citep{williams1991probability}[Chap. 9].
    
            Apply this with $W=(Z_{n+1} - \mathbb{E}\left[Z_{n+1} \mid s\right])^2$. If $\mathrm{Var}(Z_{n+1}\mid s)=0$, then
            \begin{equation*}
                \mathbb{E}\left[(Z_{n+1} - \mathbb{E}\left[Z_{n+1} \mid s\right])^2\mid s\right] = 0 \Rightarrow (Z_{n+1} - \mathbb{E}\left[Z_{n+1} \mid s\right])^2=0 \ \ \text{a.s.} \Rightarrow Z_{n+1} = \mathbb{E}\left[Z_{n+1} \mid s\right] \ \ \text{a.s.}
            \end{equation*}
            
            The covariance $\mathrm{Cov}\!\big(G_{t:t+n},Z_{n+1}\mid s\big)$ can be expressed as the following formula:
            \begin{equation*}
                \mathrm{Cov}\!\big(G_{t:t+n},Z_{n+1}\mid s\big) = \mathbb{E}\!\left[\bigl(G_{t:t+n}-\mathbb{E}[G_{t:t+n}\mid s]\bigr)\bigl(Z_{n+1}-\mathbb{E}[Z_{n+1}\mid s]\bigr)\,\middle|\, s\right]. 
            \end{equation*}
    
            Since $Z_{n+1} - \mathbb{E}\left[Z_{n+1} \mid s\right]=0$ a.s., we get the following result:
            \[
            \mathrm{Cov}\!\big(G_{t:t+n},Z_{n+1}\mid s\big) = 0.
            \]
        \end{proof}
        
        \begin{assumption}[\textbf{Coalition-Variance Monotonicity}]
        \label{assm:coalition-variance-monotonicity}
            Let $\widehat V(C,s)$ be an unbiased estimator of $V(C,s)$ from on-policy data restricted to agents in $C$, such that \(\mathbb{E}[\widehat V(C,s)]=V(C,s)\). $\mathrm{Var}(\widehat V(C,s))$ is non-decreasing in coalition size $|C|$.
        \end{assumption}
    
        \begin{remark}[\textbf{Explanation for Assumption~\ref{assm:coalition-variance-monotonicity}}]
        \label{rmk:satification-of-coalition-variance-monotonicity}
            Recall from Section~\ref{subsec:a-preliminary-result} that $v_{C, s}^{v_{s}(\mathcal{M})} = v_{s}(\mathcal{M}) v_{C, s}^{1}$. Following Definition~\ref{def:generalised-basis-games}, we can express $v_{C, s}^{v_{s}(\mathcal{M})}$ as follows:
            \begin{equation*}
                v_{C, s}^{v_{s}(\mathcal{M})}(D) = 
                \begin{cases}
                    v_{s}(\mathcal{M}) & \text{if } D \supseteq C, \\
                    0 & \text{otherwise}.
                \end{cases}
            \end{equation*}
            
            Recall that the state-specific basis game value $v_{C, s}^{v_{s}(\mathcal{M})}(\mathcal{M})$ is denoted by $V(C, s)$ for all $C \subseteq \mathcal{M}$. As described in Definition~\ref{def:ad-hoc-teamwork-in-cooperative-game}, a state-specific basis game value $v_{C, s}^{v_{s}(\mathcal{M})}(C)$ can effectively evaluate the performance of $v_{C, s}^{v_{s}(\mathcal{M})}(\mathcal{M}) = v_{s}(\mathcal{M})$, with a coalition of agents $C$\footnote{For any team of agents $D \supset C$, the state-specific basis game value would be the same as $v_{C, s}^{v_{s}(\mathcal{M})}(C)$, so it is ineffective to evaluate the performance of the $D$ with the larger size than $C$.} and a corresponding agent-type composition denoted by $\times_{i \in C} T_{i}$, where $T_{i}$ indicates the agent $i$'s type. Therefore, we can get the following result: The larger the size of coalition $C$, the more agent-type compositions need to be explored so that the estimator $\widehat V(C,s)$ of $V(C, s)$ can precisely estimate $v_{s}(\mathcal{M})$. In other words, as the size of coalition $C$ increases, higher variance is incurred to estimate $v_{s}(\mathcal{M})$, due to the growing agent-type composition space. Similarly, if the sizes of two coalitions are equal, their variances are equal. This leads to Assumption~\ref{assm:coalition-variance-monotonicity}.
        \end{remark}

        \begin{remark}[\textbf{Empirical Verification for Assumption~\ref{assm:coalition-variance-monotonicity}}]
        \label{rmk:verification-of-coalition-variance-monotonicity}
            To make Assumption~\ref{assm:coalition-variance-monotonicity} more convincing, we additionally provide a new proof-of-concept experiment to demonstrate its reasonableness. The key idea behind Assumption~\ref{assm:coalition-variance-monotonicity} is that a larger coalition $C$ is associated with a larger agent-type composition. Therefore, the variance of estimation during training would be larger. 
            
            Because we are not allowed to change the number of agents in the existing environment (which is fixed with pre-trained teammates), we justify Assumption~\ref{assm:coalition-variance-monotonicity} by showing the relation between the agent-type compositions and variances, because an agent-type composition is corresponding to a coalition of agents, as discussed in Remark~\ref{rmk:satification-of-coalition-variance-monotonicity}. In more detail, we would like to show that when the number of agent-type compositions increases, the variance will not decrease. We design three settings denoted as Exp A, Exp B and Exp C to capture different numbers of agent-type compositions. The number of agent-type compositions in descent would be $\text{Exp A} \geq \text{Exp B} \geq \text{Exp C}$, and the variance should be in the same order in expectation.\footnote{The agent type composition can be controlled by changing how the teammate types are sampled. We take the environment of 5v6 in SMAC as a testbed, where at most 5 agents in a team. Specifically, the original setting is sampling at most 4 teammates with various types, denoted as Exp A. In our new setting, we also fixed the cap of the number of teammates to sample as 3 and 2 teammates, denoted as Exp B and Exp C. In this way, the number of agent type compositions in descent would be $\text{Exp A} \geq \text{Exp B} \geq \text{Exp C}$.}
            \begin{figure}[h]
                \centering
                \includegraphics[width=0.4\linewidth]{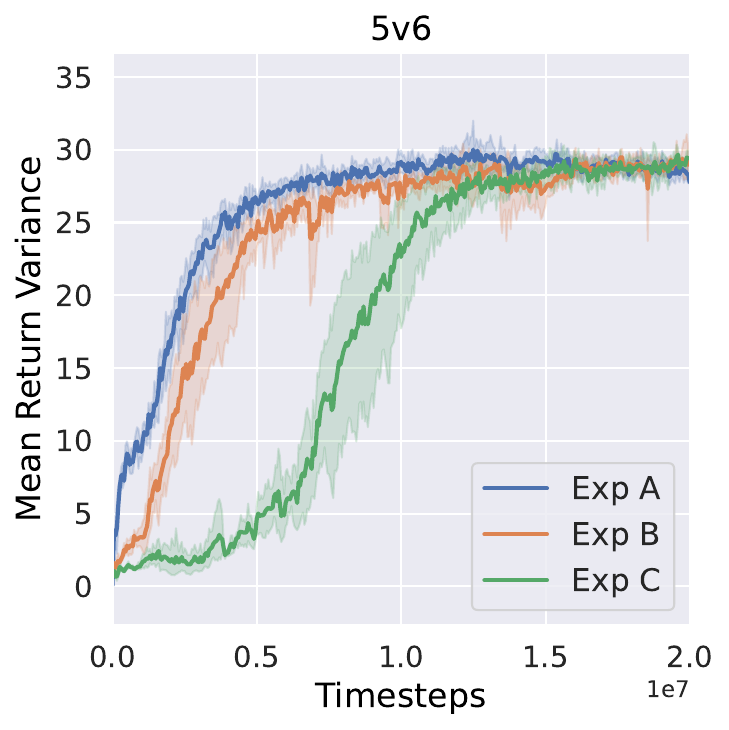}
                \caption{Empirical verification for Assumption~\ref{assm:coalition-variance-monotonicity}.}
            \label{fig:veri-exp}
            \end{figure}
            
            In Figure~\ref{fig:veri-exp}, it can be observed that the relation between variances of three setups above is almost always as the following order: $\text{Exp A} \geq \text{Exp B} \geq \text{Exp C}$, over the whole training procedure (i.e., Exp A bounded above by Exp B, and Exp B bounded above by Exp C).\footnote{The reason why the variances of three settings finally converge to the same value is that models under all scenarios have reached at least an approximate solution to the optimal one. As we mentioned in the NAHT decomposition in Section~\ref{sec:a-model-for-n-agent-ad-hoc-teamwork}, there could be some unseen teammate compositions (with unseen behaviours) appearing during the training process. As the training procedure proceeds, there could be more diverse teammate compositions (including both seen and unseen teammates) that lead to various evaluation results, which directly leads to increasingly higher variances. The variances due to the diverse teammate compositions dominate the variance decrease led by coalition values. The above result implies that the current approaches have not yet well learned to generalize to new teammates. However, it does not influence the contribution of this paper that improves on the existing methods, nor the conclusion we would like to show here. Instead, it demonstrates that this field is worth being investigated in the future.} This result conforms to our expectation above and thus justifies the reasonableness of Assumption~\ref{assm:coalition-variance-monotonicity}.
        \end{remark}
        
        \begin{proposition}[\textbf{Variance Monotonicity along the Coalition Enumeration}]\label{prop:variance-strat}
        Fix $s\in\mathcal S$ and the sequences $(C_i)$ and $(n_i)$ in Definition~\ref{def:powerset-enum}.
        Under the non-negative autocovariance condition of Lemma~\ref{lemm:variance-increment} and Assumption~\ref{assm:coalition-variance-monotonicity},
        the sequences
        \[
        \left(\mathrm{Var}(\widehat V(C_i,s))\right)_{i=1}^{m} \quad\text{and}\quad \left( \mathrm{Var}(G_{t:t+n_i}\mid s) \right)_{i=1}^{m}
        \]
        are both non-decreasing.
        \end{proposition}
        \begin{proof}
        Since $(n_i)$ is non-decreasing, Lemma~\ref{lemm:variance-increment} yields a non-decreasing variance sequence $\left( \mathrm{Var}(G_{t:t+n_i}\mid s) \right)_{i=1}^{m}$.
        Assumption~\ref{assm:coalition-variance-monotonicity} gives non-decreasing $\mathrm{Var}(\widehat V(C_i,s))$ as $|C_i|$ increases. Since $(C_i)$ is a non-decreasing sequence in size, as defined in Definition~\ref{def:powerset-enum}, the result follows.
        \end{proof}

\section{Proof for Results of Incorporating Linearity Axiom}
\label{sec:proof-linearity-axiom}
    \begin{proposition}
        Given the expression of state-specific basis game values in Eq.~\eqref{eq:game-value-expansion-coalition}, applying the Linearity axiom to state-specific cooperative games of Eq.~\eqref{eq:naht-decomposition} can lead to the following equation:
        \begin{equation*}
            \phi(v_{s_{t}}) = \mathbb{E}_{\pi} \left[ \sum_{n=1}^{m} k_{C_{n}}' \cdot \sum_{\tau=0}^{n-1} \gamma^{\tau} \phi(R_{t+\tau}) + \gamma^{n} \phi(v_{s_{t+n}}) \ \middle\vert \ s_{t} \right].
        \end{equation*}
    \end{proposition}
    \begin{proof}
        By Lemma~\ref{lemm:subgame-transform}, a state-specific cooperative game of Eq.~\eqref{eq:naht-decomposition} can be equivalently represented by the form: $v_{s} = \sum_{\emptyset \neq C \subseteq \mathcal{M}} k_{C}' \cdot v_{C,s}^{v_{s}(\mathcal{M})}$ with setting $z = v_{s}(\mathcal{M})$ and $k_{C}' = \frac{k_{C}}{v_{s}(\mathcal{M})}$, and $v_{s}(\mathcal{M}) \neq 0$.\footnote{If $v_{s}(\mathcal{M}) = 0$, it implies that a team of agents unable to cooperate, contradicting the purpose of the NAHT.} Herein, the set of $v_{C,s}^{v_{s}(\mathcal{M})}$ become a set of new state-specific basis games and $k_{C}'$ are their corresponding weightings.
        
        Introducing a linear multidimensional operator $\phi: \mathcal{G}(s) \rightarrow \mathbb{R}^{M}$ (i.e., supposing that $\phi$ satisfies the Linearity axiom) on a state-specific cooperative game $v_{s}$ defined by Lemma~\ref{lemm:subgame-transform}, we derive the following formula:
        \begin{equation}
            \phi(v_{s}) = \sum_{\emptyset \neq C \subseteq \mathcal{M}} k_{C}' \cdot \phi \left( v_{C, s}^{v_{s}(\mathcal{M})} \right).
        \label{eq:shapley-value-linear}
        \end{equation}
    
        Recall that $v_{C_{n}, s_{t}}^{v_{s_{t}}(\mathcal{M})}(\mathcal{M})$ is denoted by $V(C_{n}, s_{t})$ in Eq.~\eqref{eq:game-value-expansion-coalition}. Substituting the game expansion of $v_{C_{n}, s_{t}}^{v_{s}(\mathcal{M})}$, such that
        \begin{equation*}
            v_{C_{n}, s_{t}}^{v_{s}(\mathcal{M})} = \mathbb{E}_{\pi} \left[ \sum_{\tau=0}^{n-1} \gamma^{\tau} R_{t+\tau} + \gamma^{n} v_{s_{t+n}} \ \middle\vert \ s_{t} \right],
        \end{equation*}
        into the Linearity axiom in Eq.~\eqref{eq:shapley-value-linear} and applying linearity of the expectation operator, we obtain the following formula:
        \begin{equation*}
            \phi(v_{s_{t}}) = \mathbb{E}_{\pi} \left[ \sum_{n=1}^{m} k_{C_{n}}' \cdot \sum_{\tau=0}^{n-1} \gamma^{\tau} \phi(R_{t+\tau}) + \gamma^{n} \phi(v_{s_{t+n}}) \ \middle\vert \ s_{t} \right].
        \end{equation*}
    \end{proof}

\section{Proof for Results of Incorporating Efficiency and Symmetry Axioms}
\label{sec:proof-linearity-symmetry-axiom}
    \begin{proposition}[\textbf{Representation of Transformed Rewards}]
        Given the condition that $\sum_{i=1}^{M} \phi_{i}(R_{t}) = R_{t}$, each agent's payoff allocation defined on rewards $R_{t}$, i.e., $\phi_{i}(R_{t})$ can be expressed as
        \begin{equation*}
            \phi_{i}(R_{t}) := R_{t} - \sum_{j \neq i} \left( \phi_{j}(v_{s_{t}}) - \gamma \phi_{j}(v_{s_{t+1}}) \right).
        \end{equation*}
    \end{proposition}
    \begin{proof}
        Recall that $\phi: \mathcal{G}(s) \rightarrow \mathbb{R}^{M}$ is a multidimensional linear transformation,\footnote{In contrast to the classical formulation of \cite{dubey1975uniqueness}, we additionally assume that the multidimensional transformation is linear across both the time and agent dimensions.} where $M$ is the number of agents. We now express $\phi(R_{t})$ according to the Efficiency axiom. 
        
        To satisfy the Efficiency axiom such that $\sum_{i=1}^{M} \phi_{i}(v_{s_{t}}) = v_{s_{t}}(\mathcal{M})$, it is reasonable to assume that $\sum_{i=1}^{M} \phi_{i}(R_{t}) = R_{t}$. In other words, each agent's value expansion can be expressed independently in its own $\phi_{i}(R_{t})$, justified by Theorem~\ref{thm:factored-dec-pomdp}.
        
        Next, we aim to show how each $\phi_{i}(R_{t})$ is expressed in terms of $\phi(v_{s_{t}})$.
        
        It is not difficult to observe that for each agent $i \in \mathcal{M}$, we have 
        \begin{equation*}
            \phi_{i}(v_{s_{t}}) = \phi_{i}(R_{t}) + \gamma \phi_{i}(v_{s_{t+1}}).
        \end{equation*}
        
        By the condition $\sum_{i=1}^{M} \phi_{i}(R_{t}) = R_{t}$, we can derive a formula such that 
        \begin{equation*}
            \phi_{i}(R_{t}) = R_{t} - \sum_{j \neq i} \phi_{j}(R_{t}).
        \end{equation*}
        
        Since $\phi_{j}(R_{t}) = \phi_{j}(v_{s_{t}}) - \gamma \phi_{j}(v_{s_{t+1}})$, we can express $\phi_{i}(R_{t})$ as follows:
        \begin{equation*}
            \phi_{i}(R_{t}) := R_{t} - \sum_{j \neq i} \left( \ \phi_{j}(v_{s_{t}}) - \gamma \phi_{j}(v_{s_{t+1}}) \ \right).
        \end{equation*}
    \end{proof}

    \begin{theorem}
        The payoff allocation operator satisfying permutation-equivariance is a sufficient condition for the Symmetry axiom.
    \end{theorem}
    \begin{proof}
        Let $\mathcal{M}=\{1,\dots,M\}$. Let $\mathcal{G}$ be the set of cooperative games $v:2^\mathcal{M}\to\mathbb{R}$, and $S_{C}$ be a permutation group acting on $C \subseteq \mathcal{M}$.
        
        Let $S_{C}$ act on an arbitrary $C \subseteq \mathcal{M}$ by exchanging agents:
        \begin{equation*}
            (\sigma\!\cdot\! v)(C)\;:=\; v(\sigma^{-1}(C)),\qquad \sigma\in S_{\mathcal{M}},\; C\subseteq \mathcal{M}.
        \end{equation*}

        Let a payoff allocation operator $F:\mathcal{G} \to \mathbb{R}^M$ output payoffs $F(v)=(F_1(v),\dots,F_n(v))$.
        
        Let $S_\mathcal{M}$ act on $\mathbb{R}^M$ by permuting coordinates: $(\sigma\!\cdot\! x)_i := x_{\sigma^{-1}(i)}$.

        $F$ is permutation-equivariant if
        \begin{equation}
            F(\sigma\!\cdot\! v)\;=\;\sigma\!\cdot\! F(v),\qquad \forall\,\sigma \in S_\mathcal{M},\; v \in \mathcal{G}.
        \label{eq:PE}
        \end{equation}

        Recall that agents $i$ and $j$ are symmetric in $v$ if
        \begin{equation*}
            v(C \cup \{i\})=v(C \cup \{j\}),\quad \forall\, C \subseteq \mathcal{M} \setminus \{i,j\}.
        \end{equation*}
        
        Equivalently, the transposition $\tau=(i\ j)$ leaves the game $v$ invariant:
        \begin{equation}
            \tau\!\cdot\! v \;=\; v. 
        \label{eq:*}
        \end{equation}

        Next, we aim to show $F_i(v)=F_j(v)$ to prove the statement in the theorem.

        By Eqs.~\eqref{eq:PE} and \eqref{eq:*}, we have
        \begin{equation*}
            F(v)\;=\;F(\tau\!\cdot\! v)\;=\;\tau\!\cdot\! F(v).
        \end{equation*}
        
        Note that $\tau\!\cdot\! F(v)$ is just $F(v)$ with coordinates $i$ and $j$ swapped. Therefore, $F(v)=\tau\!\cdot\! F(v)$ implies the $i$-th and $j$-th entries are equal:
        $$
        F_i(v)=F_j(v).
        $$
    \end{proof}
    
    \begin{theorem}
        Shapley Machine is an algorithm enforcing $V_{i}$ to fulfil Efficiency, Symmetry and Linearity, so the $V_{i}$ is the Shapley value for Dec-POMDPs.
    \end{theorem}
    \begin{proof}
        By construction, the Shapley Machine enforces the individual value functions $V_i$ (payoff allocation functions) to satisfy the axioms of Efficiency, Symmetry and Linearity. 
        Each state-specific cooperative game space $\mathcal{G}(s)$ that aims to meet the requirement of Dec-POMDPs is isomorphic to the canonical cooperative game space $\mathcal{G}$. 
        Hence, the properties of payoff allocation functions that hold on $\mathcal{G}$ also hold on $\mathcal{G}(s)$. 
        
        Now, recall that the Shapley value is uniquely characterized as the value function that satisfies the axioms of Efficiency, Symmetry and Additivity (Theorem~\ref{thm:shapley-axioms}). 
        Since Linearity is a stronger condition than Additivity (indeed, setting all $\alpha_{i}=1$ in the definition of Linearity yields Additivity), any payoff allocation function that is Efficient, Symmetric, and Linear must coincide with the Shapley value. 
        Therefore, the individual value functions $V_i$ produced by the Shapley Machine are exactly the Shapley values for Dec-POMDPs. 
    \end{proof}
    
\section{Experimental Details}
\label{sec:experimental-details}
    \subsection{Implementation Details}
    \label{subsec:implementation-details}
        Our algorithm is built upon POAM and IPPO. All loss functions for training the encoder-decoder model, policy networks and value networks have been remained. Please refer to Appendix~\ref{subsec:poam} for details. For conciseness, we only list the novel loss functions proposed in this paper.
        
        \subsubsection{Shapley Machine}
            We now describe the details about implementing our proposed algorithm, referred to as Shapley Machine. In general, our algorithm is established based on the base algorithms POAM and IPPO, with modification to fulfil all the three axioms of the Shapley value: Efficiency, Linearity and Symmetry. Since the Symmetry axiom has been realised in POAM and IPPO, we only need to fulfil the Efficiency and Linearity axioms. 
            
            \textbf{Implementing the Linearity Axiom.} In general, both POAM and IPPO have implemented TD($\lambda$), which does not strictly conform to the principle of the Linearity axiom. To address this problem, we change the TD($\lambda$) to TTD($\lambda$), where the number of $n$-step return components is equal to the number of the non-empty coalition basis games, as informed by our theory. Note that in some scenarios the episode length is smaller than the number of non-empty coalition basis games. In these cases, the TTD($\lambda$) can be reduced to the TD($\lambda$) for the finite-horizon tasks with the episode length as $T$, equivalently, the TTD($\lambda$) with $m=T$. Alternatively, we can select a value of $m$ for each task based on extra conditions, which is left for the future work. In this paper, we set $m=T$ for the scenarios 8v9 and 10v11. For the 8v9 scenario, the episode length as $120$ is not too far from the number of non-empty coalition basis games as $127$. For the 10v11 scenario, the episode length is $150$, while the number of non-empty coalition basis games is $511$.
            
            \textbf{Implementing the Efficiency Axiom.} By introducing the partial observation in Dec-POMDPs, the Eq.~\eqref{eq:shape-rewards} we have derived the condition for realizing the Efficiency axiom is transformed as follows:
            \begin{equation*}
                \phi_{i}(R_{t}) = R_{t} - \sum_{j \neq i} \left( V_{j}(h_{t,j}) - \gamma V_{j}(h_{t+1,j}) \right),
            \end{equation*}
            where $V_{j}(h_{t,j})$ indicates an agent $j$'s non-parameterized individual value. During the practical training procedure, the $V_{j}(h_{t,j})$ generated by the individual value network could be severely inaccurate in the beginning, which may result in the instability of learning. To mitigate this issue, we add an extra coefficient $\alpha \in (0, 1)$ to the term $\sum_{j \neq i} \left( V_{j}(h_{t,j}) - \gamma V_{j}(h_{t+1,j}) \right)$, such that
            \begin{equation}
                \phi_{i}(R_{t}) = R_{t} - \alpha \sum_{j \neq i} \left( V_{j}(h_{t,j}) - \gamma V_{j}(h_{t+1,j}) \right).
            \label{eq:shaping-reward-implementation}
            \end{equation}
            
            This coefficient $\alpha$ can be either manually set up as a fixed value, or implemented by a scheduler starting from $0$ to some preset upper limit. The $R_{t}$ in $L_{\theta}(h_{t,i}, a_{t,i})$ and $L_{\theta^{c}}(h_{t,i})$ is replaced by the above $\phi_{i}(R_{t})$. To clarify this change, the two new losses are expressed as $\hat{L}_{\theta}(h_{t,i}, a_{t,i})$ and $\hat{L}_{\theta^{c}}(h_{t,i})$.
        
            Furthermore, it is needed to search the underlying factorization scheme of the transition function in Dec-POMDPs, according to Theorem~\ref{thm:factored-dec-pomdp}. To implement this requirement, we need to fulfil the following condition:
            \begin{equation}
            \label{eq:efficiency-eq}
                \sum_{i=1}^{M} V_{i}(h_{t,i}) = v_{h_{t}}(\mathcal{M}).
            \end{equation}
            Here, $h_{t} = (h_{t,1}, \ldots, h_{t,M})$ denotes the joint history, and $v_{h_{t}}(\mathcal{M})$ denotes the team value conditioned on this joint history. The above equality is implemented as a regularization term during training.
            
            To maintain consistency with the critic losses, we consider to use the $\lambda$-return denoted by $\hat{G}_{t}^{\lambda}$ to represent $v_{h_{t}}(\mathcal{M})$. In turn, the above equation can be expressed as
            \begin{equation*}
                v_{h_{t}}(\mathcal{M}) = \mathbb{E}_{\pi} \left[ G_{t}^{\lambda} \middle| h_{t} \right],
            \end{equation*}
            
            where each $\hat{G}_{t:t+n}$ contributing to the $\hat{G}_{t}^{\lambda}$ represented in TTD($\lambda$) is expressed as follows:
            \begin{equation*}
                \begin{split}
                    \hat{G}_{t:t+n} &= R_{t} + \gamma R_{t+1} + \cdots + \gamma^{n-1} R_{t+n-1} + \gamma^{n} v_{h_{t+n}}(\mathcal{M}) \\
                    &= R_{t} + \gamma R_{t+1} + \cdots + \gamma^{n-1} R_{t+n-1} + \gamma^{n} \sum_{i=1}^{M} V_{i}(h_{t+n,i}) \quad \text{(By Eq.~\eqref{eq:efficiency-eq})}.
                \end{split}
            \end{equation*}
            
            Substituting the above formula into Eq.~\eqref{eq:efficiency-eq}, we can obtain the regularization term, referred to as the \textbf{efficiency loss} for one timestep $t$, as follows:
            \begin{equation*}
                L_{\theta^{c}}^{e}(h_{t}) = \frac{1}{2} \left( \hat{G}_{t}^{\lambda} - \sum_{i=1}^{M} V_{i}^{\theta^{c}}(h_{t,i}) \right)^{2}.
            \end{equation*}
        
            In summary, the total loss function of the Shapley Machine for POAM (SM-POAM) is as follows:
            \begin{equation*}
                L_{\texttt{SM-POAM}} = \frac{1}{T} \sum_{t=1}^{T} \left( \sum_{i \in \mathcal{C}_{t}} \hat{L}_{\theta}(h_{t,i}, a_{t,i}) + \beta_{1} \sum_{i \in \mathcal{M}} \hat{L}_{\theta^{c}}(h_{t,i}) + \beta_{2} L_{\theta^{c}}^{e}(h_{t}) + L_{\theta^{e}, \theta^{o}, \theta^{a}}(h_{t,i}, o_{t,-i}, a_{t,-i}) \right),
            \end{equation*}
            where $T$ is the episode length; $\beta_{1}, \beta_{2} \in (0, 1)$ are two coefficients to control the importance of the two losses; $\mathcal{C}_{t}$ indicates the controlled agent set at timestep $t$ and $\mathcal{M}$ indicates the ad hoc team following the convention in \cite{wang2025n}.

            Similarly, the total loss function of the Shapley Machine for IPPO (SM-IPPO) is as follows:
            \begin{equation*}
                L_{\texttt{SM-IPPO}} = \frac{1}{T} \sum_{t=1}^{T} \left( \sum_{i \in \mathcal{C}_{t}} \hat{L}_{\theta}(h_{t,i}, a_{t,i}) + \beta_{1} \sum_{i \in \mathcal{M}} \hat{L}_{\theta^{c}}(h_{t,i}) + \beta_{2} L_{\theta^{c}}^{e}(h_{t}) \right).
            \end{equation*}

            \textbf{Implementation of $\boldsymbol{k_{C}' > 0}$.} As mentioned in Section~\ref{subsec:a-preliminary-result}, it is necessary to fulfil $k_{C}' > 0$ for reaching cooperation, which can be implemented following the convention of TD($\lambda$) in the RL~\citep{sutton1988learning}. Specifically, the weightings for $m$ basis games $(k_{C_{1}}', k_{C_{2}}', \cdots, k_{C_{m}}')$ are generated using a geometric distribution $P_{\lambda}$ with the parameter $0 < \lambda < 1$, such that $k_{C_{n}}' = P_{\lambda}(n)$, resulting a tuple $((1-\lambda), (1-\lambda)\lambda, \cdots, (1-\lambda)\lambda^{m-2}, \lambda^{m-1})$. With this condition, Eq.~\eqref{eq:td-error-general} becomes the well-known \textbf{truncated TD($\boldsymbol{\lambda}$)}, shortened as \textbf{TTD($\boldsymbol{\lambda}$)}~\citep{cichosz1994truncating}.
            
            \textbf{Partial Observations.} In practice, a controlled agent is only able to receive an observation, following the settings of Dec-POMDPs. Therefore, an agent is required to infer the state of the environment as an individual hidden state, through the history of observations. To this end, the policy and individual networks are realized by recurrent neural networks (RNNs) (e.g., GRUs~\citep{chung2014empirical}), where observations or representations transformed from observations are used as inputs.

            \textbf{Implementation of the Symmetry Axiom.} It can be observed that both POAM and IPPO are implemented by the sharing parameters, with agent ID to differentiate agent identities. Given an agent ID, its input is either individual observations or individual observations + teammate embeddings, where the teammate embeddings are transformed from individual observations through an encoder network. 
        
            \textbf{Estimation of Uncontrolled Agent Individual Values.} We maintain all possible controlled agent policy and individual value networks as implemented by~\cite{wang2025n}, given that the maximum number of controlled agents is known in the experimental settings of the NAHT, but only part of controlled agents in an episode can make decision. Thanks to the parameter sharing technique, this can be simply implemented by maintaining only a policy and/or an individual value network, with a set of agent IDs to represent and differentiate all possible controlled agents. During the training phase, the uncontrolled agents' individual values are also required to implement the Efficiency axiom, as informed in Eq.~\eqref{eq:efficiency}, but they are unknown to the controlled agents; all training paradigms that we design should be based on the view of the controlled agents. To address this problem, the uncontrolled agents' individual values are approximated by the maintained individual value function with the agent IDs of those inactive controlled agents, given that the observations of the uncontrolled agents are available for controlled agents during training.

            \begin{example}[Estimating an Uncontrolled Agent's Individual Value]
            \label{ex:uncontrolled-value-estimation}
            Consider a team $\mathcal{M}=\{c_1,c_2,u\}$ consisting of two controlled agents, $\mathcal{C}=\{c_1,c_2\}$, and one uncontrolled agent,
            $\mathcal{U}=\{u\}$. Suppose the implementation maintains a shared policy network and a shared individual value network, with three available controlled-agent IDs: $\{1,2,3\}$. In an episode, $c_1$ and $c_2$ use IDs $1$ and $2$, respectively, leaving ID $3$ unused.

            Let $V_{\psi}(h,i)$ denote the shared individual value network evaluated using observation history $h$ and agent ID $i$. Since the uncontrolled agent's observations are available during training, we approximate its
            individual value by supplying its observation history to the shared network with the unused ID:
            \[
                \widehat{V}_{u}(h_{t,u})
                := V_{\psi}(h_{t,u},3).
            \]
            As a result, the controlled agent associated with ID $3$ does not participate in the episode, but its ID is used to obtain a proxy value estimate for $u$.
            \end{example}
            
        \subsubsection{POAM and IPPO}
            The implementations of POAM and IPPO have been detailed in Appendix~\ref{subsec:poam}. 

            In summary, the total loss function of POAM is as follows:
            \begin{equation*}
                L_{\texttt{POAM}} = \frac{1}{T} \sum_{t=1}^{T} \left( \sum_{i \in \mathcal{C}_{t}} L_{\theta}(h_{t,i}, a_{t,i}) + \beta_{1} \sum_{i \in \mathcal{M}} L_{\theta^{c}}(h_{t,i}) + L_{\theta^{e}, \theta^{o}, \theta^{a}}(h_{t,i}, o_{t,-i}, a_{t,-i}) \right).
            \end{equation*}
            
            The total loss function of IPPO is as follows:
            \begin{equation*}
                L_{\texttt{IPPO}} = \frac{1}{T} \sum_{t=1}^{T} \left( \sum_{i \in \mathcal{C}_{t}} L_{\theta}(h_{t,i}, a_{t,i}) + \beta_{1} \sum_{i \in \mathcal{M}} L_{\theta^{c}}(h_{t,i}) \right).
            \end{equation*}

        \subsubsection{Banzhaf Machine}
            The overall loss function and implementation of the Banzhaf Machine closely follow those of POAM and IPPO. The only distinction is that Banzhaf Machine employs the TTD($\lambda$) in place of the TD($\lambda$) that is applied in POAM and IPPO. This implies that Banzhaf Machine uses $\phi_i(R_t)=R_t$ as its implementation, but this does not mean that it is the only implementation.

    \subsection{Experimental Domains}
    \label{subsec:experimental-domains}
        We now briefly introduce the experimental domains for running experiments. If one would like to know more about details, please refer to \cite{wang2025n}.
        \subsubsection{MPE Predator Prey (MPE-PP)} 
            The MPE Predator Prey (MPE-PP) environment is a predator-prey task implemented within the Multi-Agent Particle Environment (MPE) framework. It simulates interactions within a two-dimensional space populated by two static obstacles, where three pursuer agents must cooperate to capture a single adversarial evader. A successful capture is defined as at least two pursuers simultaneously colliding with the evader, upon which the pursuers receive a positive reward of +1. If the capture is unsuccessful, no reward is granted. This environment is designed to test the ability of agents to coordinate under spatial and dynamic constraints. The length of an episode is $T=100$, 

        \subsubsection{The StarCraft Multi-Agent Challenge (SMAC)}
            The StarCraft Multi-Agent Challenge (SMAC) serves as a benchmark suite for evaluating CMARL algorithms in partially observable settings. Built on top of the StarCraft II game engine, SMAC presents a variety of micromanagement tasks where each agent (e.g., a Marine or Stalker) operates based on limited local observations and must coordinate actions with teammates to overcome enemy units. In this work, we focus on four specific SMAC scenarios:
            \textbf{5v6}: five allied Marines versus six enemy Marines,
            \textbf{8v9}: eight allied Marines versus nine enemy Marines,
            \textbf{10v11}: ten allied Marines versus eleven enemy Marines,
            \textbf{3s5z}: three allied Stalkers versus five enemy Zealots.
            At each timestep, agents receive a shaped reward proportional to the damage they inflict on opponents, along with bonus rewards of 10 points for each enemy defeated and 200 points for achieving overall victory by eliminating all adversaries. The total return is normalized so that the maximum achievable return in each scenario is 20. The action space in SMAC is discrete, enabling each agent to choose actions such as attacking a particular enemy, moving in a specific direction, or remaining idle. The length of an episode varies across different scenarios: 3sv5z with $T=250$, 5v6 with $T=70$, 8v9 with $T=120$ and 10v11 with $T=150$.

    \subsection{Evaluation Metrics}
    \label{subsec:evaluation-metrics}
        For in-distribution evaluation, the mean return (or winrate) is computed over $E$ randomly sampled episodes. In each episode, we form a joint policy $\pi^{(\mathcal{M})}$ by sampling $N$ agents uniformly from $\mathcal{C}$ and the remaining $M-N$ agents from $\mathcal{U}$. For the out-of-distribution (OOD) evaluation, the M--N score~\citep{wang2025n} is applied. Given a set of controllable agents $\mathcal{C}$ and uncontrolled agents $\mathcal{U}$, the M--N score measures how well mixed teams of these agents cooperate in the NAHT setting. The score is computed exhaustively by changing the number of controllable agents included in the team. Specifically, for each $N \in \{1, \dots, M-1\}$, we form a joint policy $\pi^{(\mathcal{M})}$ by sampling $N$ agents uniformly from $\mathcal{C}$ and the remaining $M-N$ agents from $\mathcal{U}$. The mixed teams are then evaluated on the task for $E$ episodes. This procedure yields $(M-1)\cdot E$ returns (or winrates) in total, whose average defines the M--N score. 
        
        In this experiment, each convention of an uncontrolled agent is a random seed that pretrains it. For the OOD evaluation on the unseen conventions, the M--N score is computed by averaging the per-convention M--N scores across 4 other conventions for each agent type. For the OOD evaluation on the unseen agent types, the per-convention M--N score is computed by averaging the per-convention M--N score across 5 conventions for each unseen agent type. All results are obtained by first computing each metric with 128 episodes and then averaging these per-seed metrics across 5 random seeds, with 95\% confidence intervals.
        
    \subsection{Experimental Settings}
    \label{subsec:experimental-settings}
        The training procedure of the NAHT process is briefly introduced here. For more details, please refer to \cite{wang2025n}. For each scenario (e.g. MPE, 3sv5z, 5v6, 8v9 and 10v11), there are five groups of pretrained agents (e.g. IQL, MAPPO, VDN, QMIX and IPPO) acting as the uncontrolled agents. Given that each task specifies a fixed total number of agents $M$, the number of controlled agents is $N$. For each episode evaluation, the number of uncontrolled agents $M-N$ is sampled, and then a group of pretrained agents is sampled.\footnote{Note that this is still a special case of openness. For a varying number of controlled agents, the number of uncontrolled agents is also varied in correspondence, constrained by the total number of agents related to the task specifications.} The distribution for all sampling procedures are modelled as the multinational distribution with no replacement. Each uncontrolled agent executes the greedy policy in both training and testing procedures. In contrast, each controlled agent executes the on-policy sampling via the parameterized policy during the training procedure, while the greedy policy during the testing procedure. 

    \subsection{Hyperparameter Settings}
    \label{subsec:hyperparameter-settings}
        Since our algorithm is established based on POAM and IPPO, most hyperparameter settings follow that in \cite{wang2025n}. First, the actors and critics are implemented in recurrent neural networks, with full parameter sharing. Specifically, they are implemented by two fully connected layers followed by a GRU layer and an output layer. Each layer has 64 dimensions with a ReLU activation function, and layer normalization is applied. The encoder-decoder networks for inferring agent characteristics are also implemented in parameter sharing. The encoder is implemented by a GRU layer, followed by a fully connected layer with a ReLU activation function and an output layer. The decoder is implemented by two fully connected layers with ReLU activation functions, followed by an output layer. Adam Optimizer is used to train all models. The detailed hyperparameters for experiments are shown in Tables~\ref{tab:shapley-machine-hyperparams} and \ref{tab:common-hyperparams}. For the scenarios such as MPE, 3sv5z and 5v6, the values of $m$ are set as the number of non-empty coalition basis games. For the scenarios such as 8v9 and 10v11, the values of $m$ are simply set as the length of an episode (see Appendix~\ref{subsec:implementation-details} for more details). Note that the term $\sum_{j \neq i} \left( V_{j}(s_{t}) - \gamma V_{j}(s_{t+1}) \right)$ in shaped rewards is standardised to match the scales of standardised rewards. To stabilize learning in large-scale scenarios, we add a value loss clip to Shapley Machine and Banzhaf Machine for POAM on both 8v9 and 10v11, and for IPPO on 10v11 only.
        
        \begin{table}[ht]
        \caption{Key hyperparameters of Shapley and Banzhaf Machine: $\lambda$ indicates the parameter for the weighting functions of TTD($\lambda$), $m$ indicates the number of basis games, $\alpha$ controls the importance of the term $\sum_{j \neq i} \left( V_{j}(s_{t}) - \gamma V_{j}(s_{t+1}) \right)$ in Eq.~\eqref{eq:shaping-reward-implementation}, $\beta_{1}$ controls the importance of the critic loss $\sum_{i \in \mathcal{M}} \hat{L}_{\theta^{c}}(h_{t,i})$, and $\beta_{2}$ controls the importance of the efficiency loss $L_{\theta^{c}}^{e}(h_{t})$.}
        \label{tab:shapley-machine-hyperparams}
        \centering
        \begin{tabular}{ccccccc}
            \hline\hline
            Algorithm & Task & $\lambda$ & $m$ & $\alpha$ & $\beta_{1}$ & $\beta_{2}$ \\
            \hline\hline
            Shapley Machine 
              & MPE-PP & $0.85$ & $7$   & $0.01$  & $0.5$ & $0.01$  \\
              & 3sv5z  & $0.85$ & $7$   & $0.01$  & $0.5$ & $0.01$  \\
              & 5v6    & $0.85$ & $31$  & $0.01$ & $0.5$ & $0.001$ \\
              & 8v9    & $0.95$ & $120$ & $0.01$  & $0.5$ & $0.001$ \\
              & 10v11  & $0.95$ & $150$ & $0.01$  & $0.5$ & $0.001$ \\
            \hline
            Banzhaf Machine 
              & MPE-PP & $0.85$ & $7$   & N/A  & $0.5$ & N/A  \\
              & 3sv5z  & $0.85$ & $7$   & N/A  & $0.5$ & N/A  \\
              & 5v6    & $0.85$ & $31$  & N/A & $0.5$ & N/A \\
              & 8v9    & $0.95$ & $120$ & N/A  & $0.5$ & N/A \\
              & 10v11  & $0.95$ & $150$ & N/A  & $0.5$ & N/A \\
            \hline\hline
        \end{tabular}
        \end{table}

        \begin{table}[ht]
        \caption{Common hyperparameters of RL settings.}
        \label{tab:common-hyperparams}
        \centering
        \begin{tabular}{cc}
            \hline\hline
            Hyperparameter & Value \\
            \hline\hline
            LR & $0.0005$ \\
            Epochs & $5$ \\
            Minibatches & $1$ \\
            Buffer size & $256$ \\
            Entropy coefficient& $0.05$ \\
            Clip & $0.2$ \\
            ED LR & $0.0005$ \\
            ED epochs & $1$ \\
            ED Minibatches & $1$ \\
            Optim\_alpha (Adam) & $0.99$ \\
            Optim\_eps (Adam) & $0.00001$ \\
            Use\_obs\_norm & True \\
            Use\_orthogonal\_init & True \\
            Use\_adv\_std & True \\
            Standardise\_rewards & True \\
            num\_parallel\_envs & 8 \\
            \hline\hline
        \end{tabular}
        \end{table}
        
    \subsection{Computational Resources}
    \label{subsec:computational-resources}
        All experiments are conducted on Intel Xeon Gold 6230 CPUs and Nvidia V100-SXM2 GPUs. Each experiment run on MPE takes approximately 7 hours, utilizing 20 CPU cores and 1 GPU. Each experiment run on SMAC takes between 8 and 19 hours, utilizing 30 CPU cores and 1 GPU. All experiments are trained with 20M timesteps. The post-training evaluation for each scenario takes between 6 and 16 hours, utilizing 8 CPU cores and 1 GPU.
        
\section{Additional Experiments}
\label{sec:additional-experiments}
    \subsection{Empirical Evidence for the Strength of the Axiomatic Framework}
    \label{subsec:further-evidence-on-the-benefit-of-structured-algorithm}
        We now discuss the phenomenon observed from the performance under stochastic policy exploration during training. Discussing this phenomenon is meaningful: When a machine learning algorithm is deployed and has to adapt to real-world environments through online learning due to the mismatch between simulators and real-world environments, it is expected that random exploration diminishes rapidly over time to avoid the damage on real-world environments (e.g. physical systems).
        
\begin{figure}[ht!]
            \centering
            \includegraphics[width=\linewidth]{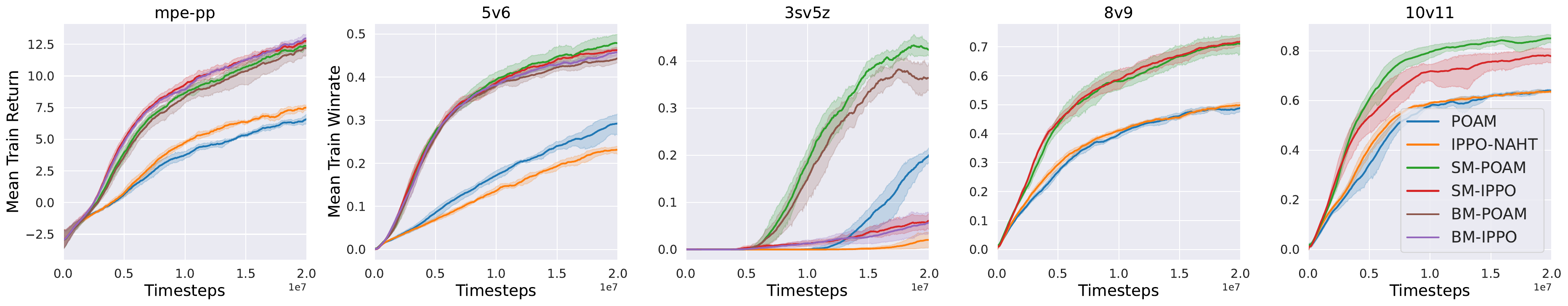}
            \caption{Performance under the stochastic exploration policy during training in NAHT.}
        \label{fig:train-return}
        \end{figure}

\begin{figure}[ht!]
            \centering
            \includegraphics[width=\linewidth]{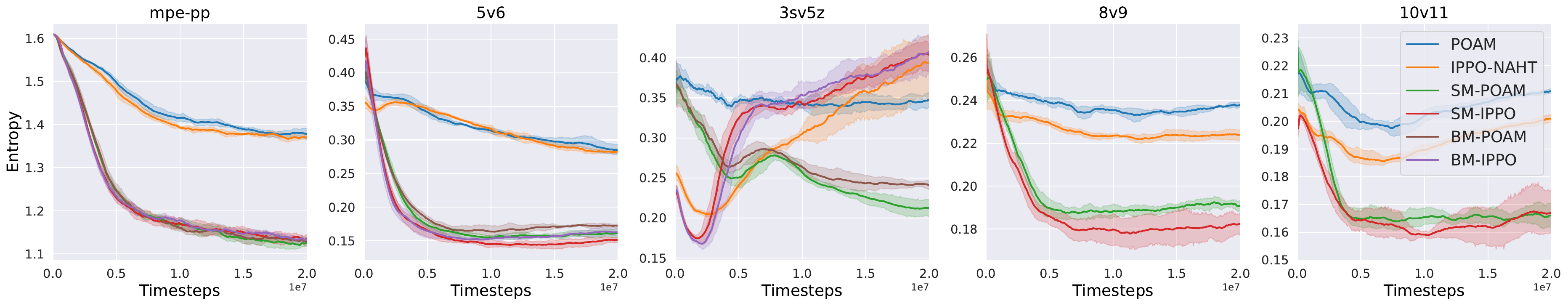}
            \caption{Policy entropy across scenarios during training in NAHT.}
        \label{fig:entropy}
        \end{figure}

\begin{figure}[ht!]
            \centering
            \includegraphics[width=\linewidth]{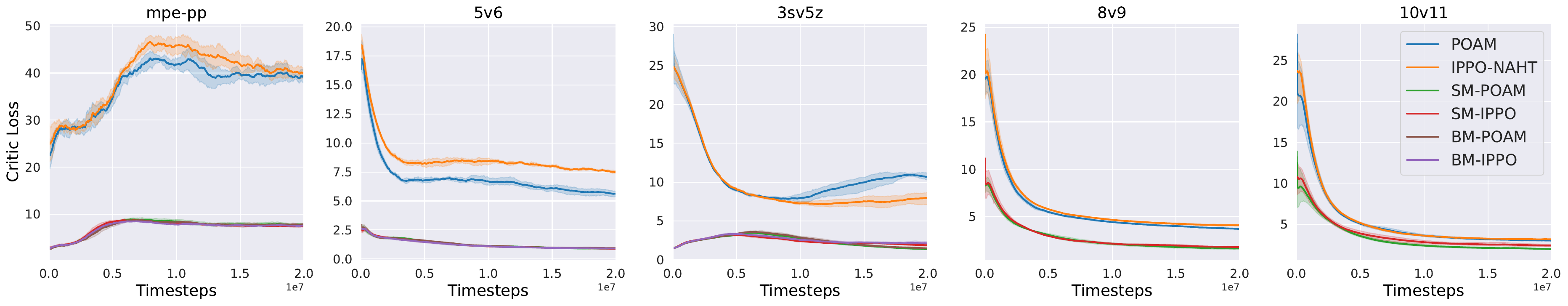}
            \caption{Critic loss across scenarios during training in NAHT.}
        \label{fig:train-critic-loss}
        \end{figure}

        Recall that the PPO family algorithms implement exploration by the natural stochastic policy sampling from the learned policy distribution and maximizing policy entropy. As seen from Figure~\ref{fig:train-return}, either Shapley Machine or Banzhaf Machine can outperform their base algorithm counterparts with a large margin. The reasons are as follows: As shown in Figure~\ref{fig:entropy}, policy entropy decreases rapidly over the course of training (except for the IPPO implementation on 3sv5z, which underperforms due to incompatibility between IPPO and the task). This indicates that the learned policies become more deterministic than those of the base algorithms, despite all methods being optimized with an entropy-maximization objective. A possible explanation for this phenomenon is illustrated in Figure~\ref{fig:train-critic-loss}: The critic losses of Shapley and Banzhaf Machines decrease faster than that of their base algorithm counterparts. This provides evidence for the advantage of structured individual value functions under our proposed axiomatic framework.

        \begin{figure}[ht!]
            \begin{subfigure}[b]{\textwidth}
                \centering
                \includegraphics[width=\linewidth]{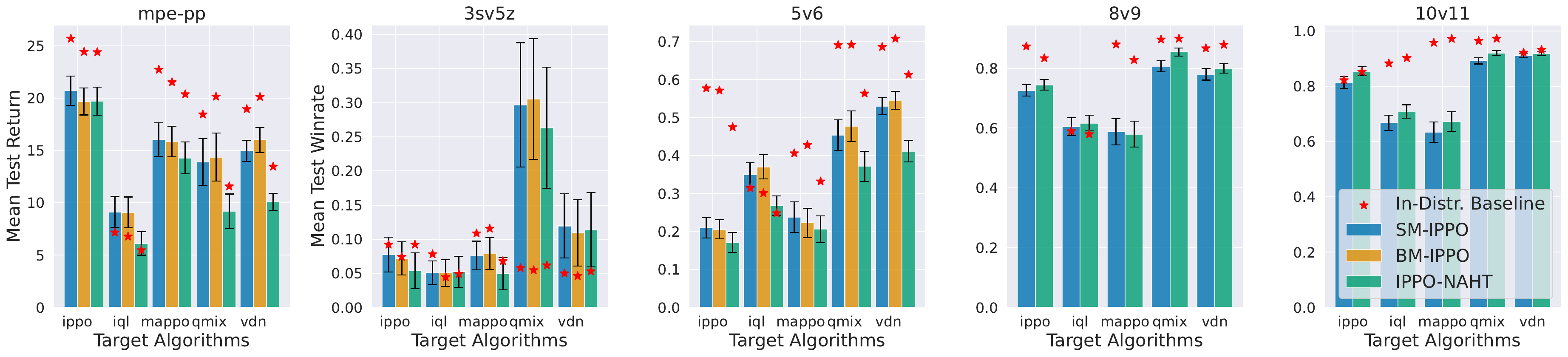}
                \caption{Evaluation on unseen conventions (identical algorithms with different random seeds). The star denotes the in-convention baseline, and the histograms show performance on unseen conventions.}
            \label{fig:unseen-convention-ippo}
            \end{subfigure}
            \hfill
            \begin{subfigure}[b]{\textwidth}
                \centering
                \includegraphics[width=\linewidth]{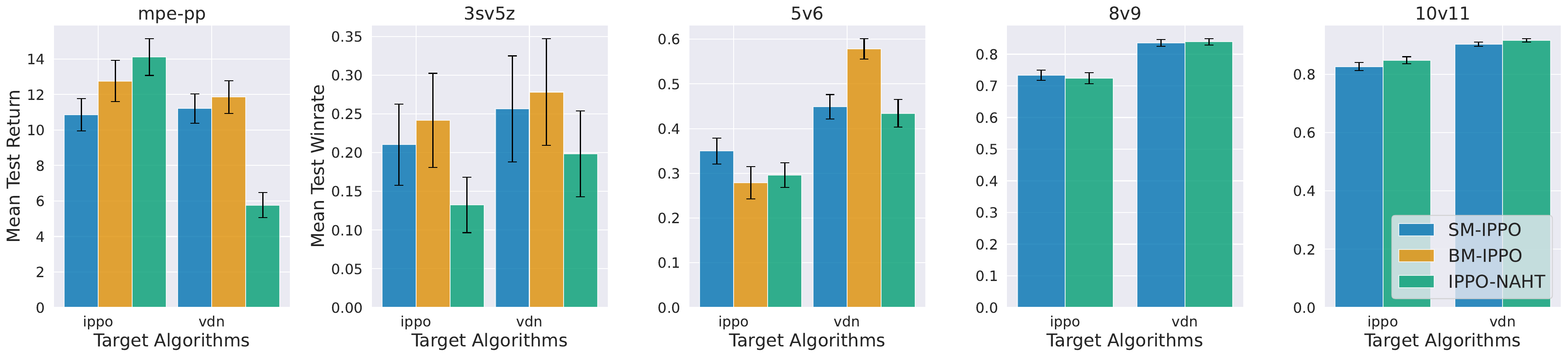}
                \caption{Evaluation on unseen agent types. Train set: MAPPO, QMIX, IQL; test set: IPPO, VDN.}
            \label{fig:unseen-agent-ippo}
            \end{subfigure}
        \caption{Out-of-distribution (OOD) evaluation after training, assessing unseen conventions and unseen agent types during training, between learned agents (IPPO-NAHT, SM-IPPO and BM-IPPO) and uncontrolled agent types (IPPO, VDN, IQL, QMIX and MAPPO). The performance is evaluated by averaging all pairings of N controlled and $M-N$ uncontrolled agents and their corresponding random seeds, called M--N score~\citep{wang2025n}.}
        \label{fig:type-ood-ippo}
        \end{figure}

        \begin{figure}[ht!]
        \centering
            \begin{subfigure}[b]{\textwidth}
                \centering
                \includegraphics[width=\linewidth]{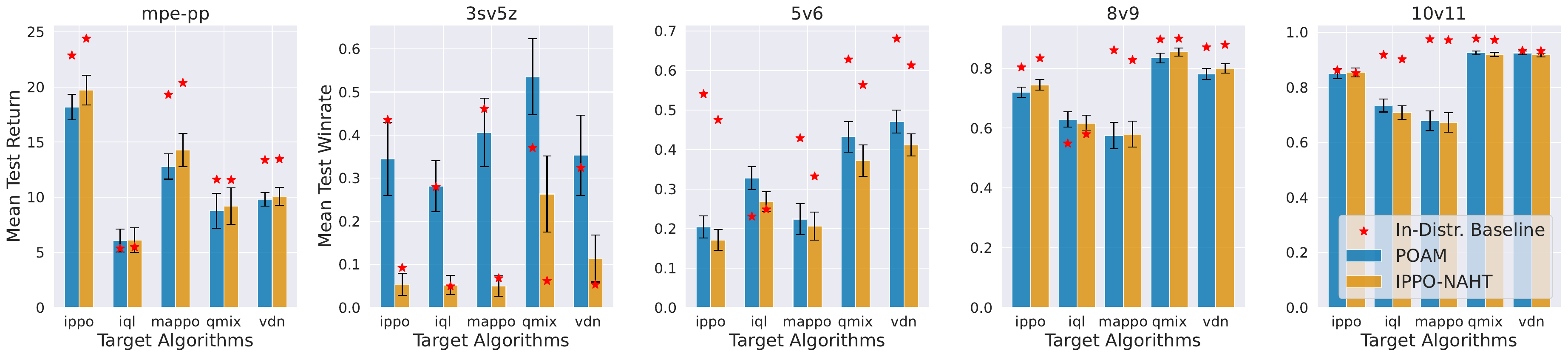}
                \caption{Evaluation on unseen agent types between POAM and IPPO (IPPO-NAHT).}
            \label{fig:unseen-convention-baseline}
            \end{subfigure}
            \hfill
            \begin{subfigure}[b]{\textwidth}
                \centering
                \includegraphics[width=\linewidth]{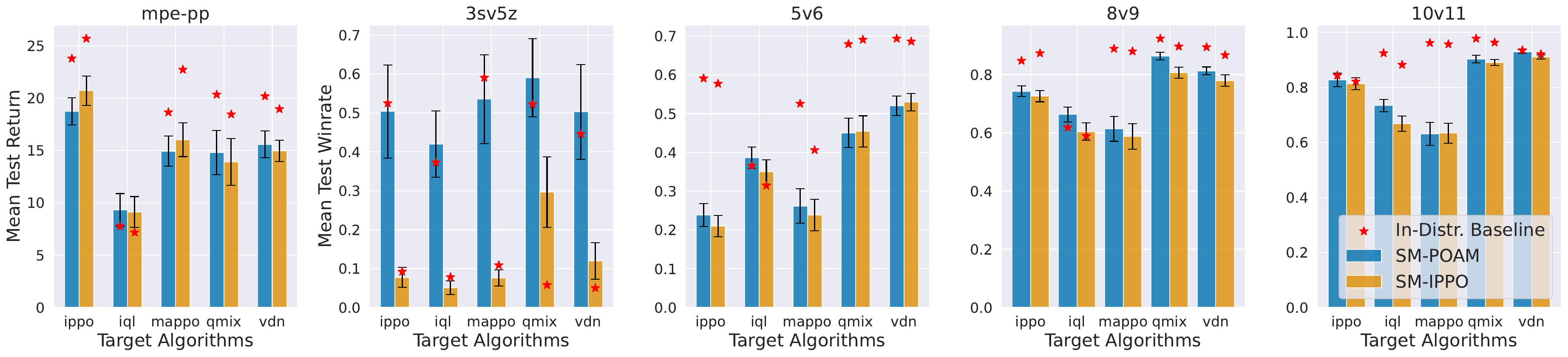}
                \caption{Evaluation on unseen agent types between SM-POAM and SM-IPPO.}
            \label{fig:unseen-convention-sm}
            \end{subfigure}
            \hfill
            \begin{subfigure}[b]{\textwidth}
                \centering
                \includegraphics[width=0.6\linewidth]{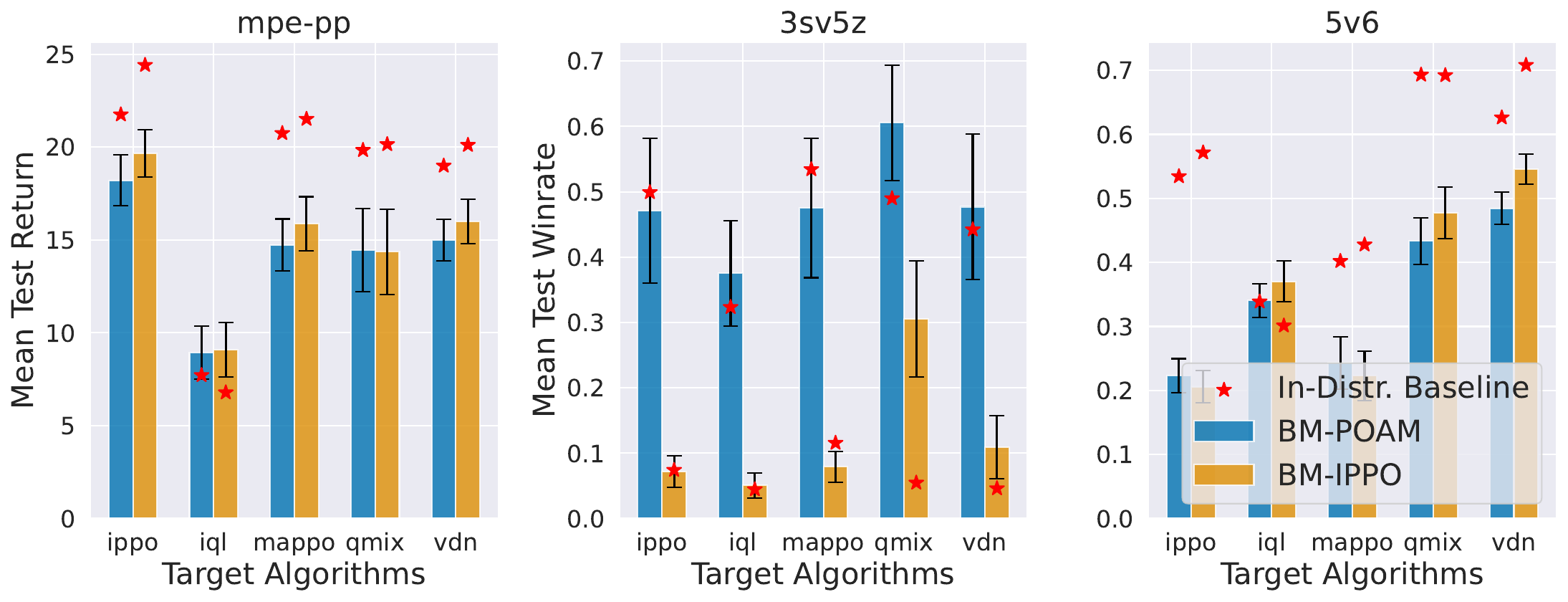}
                \caption{Evaluation on unseen agent types between BM-POAM and BM-IPPO. Since the implementation of Banzhaf Machine is equivalent to its counterpart baseline algorithms for 8v9 and 10v11 due to approximating the number of non-empty coalition basis games  as the episode length, we ignore their evaluation here.}
            \label{fig:unseen-convention-bm}
            \end{subfigure}
        \caption{Out-of-distribution (OOD) evaluation after training, assessing unseen conventions during training, between learned agents (IPPO-NAHT, SM-IPPO, and BM-IPPO) and uncontrolled agent types (IPPO, VDN, IQL, QMIX and MAPPO). The performance is evaluated by averaging all pairings of N controlled and $M-N$ uncontrolled agents and their corresponding random seeds, called M--N score~\citep{wang2025n}.}
        \label{fig:convention-ood-other}
        \end{figure}
        
    \subsection{Analyzing Effects of Efficiency Axioms}
    \label{subsec:analyzing-effects-of-efficiency-axioms}
        \begin{figure}[ht!]
            \centering
            \includegraphics[width=0.95\linewidth]{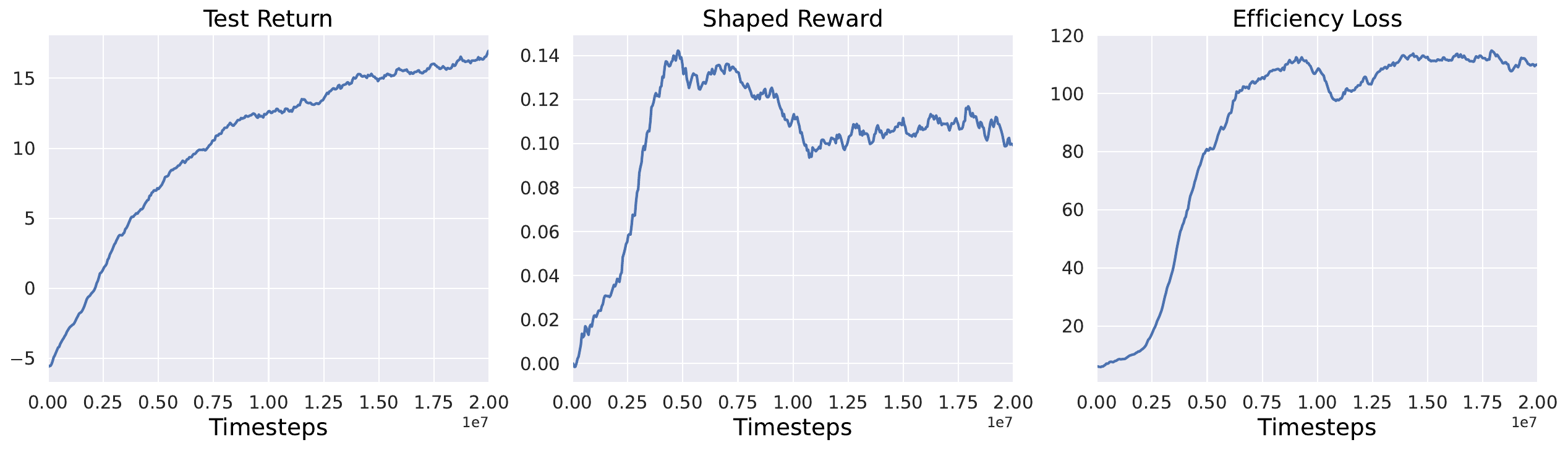}
            \caption{One run of MPE to show the variation of test return, shaped reward and efficiency loss.}
        \label{fig:mpe-efficiency-demo}
        \end{figure}
    
        \begin{figure}[ht!]
            \centering
            \includegraphics[width=0.95\linewidth]{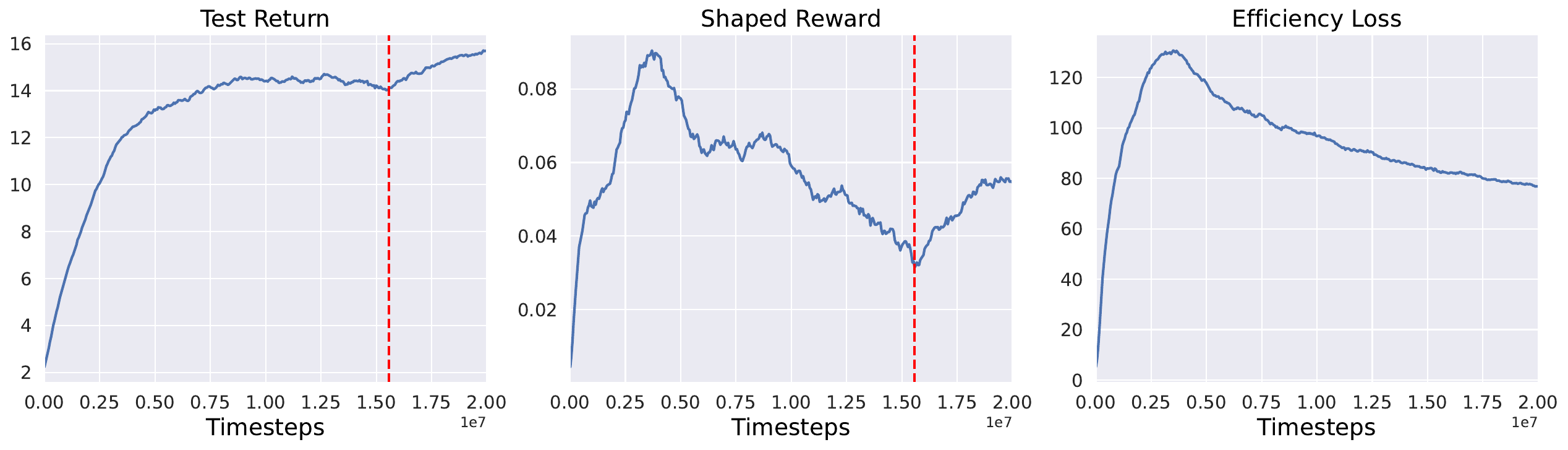}
            \caption{One run of 5v6 to show the variation of test return, shaped reward and efficiency loss.}
        \label{fig:5v6-efficiency-demo}
        \end{figure}
        We now demonstrate the association between the changing of the shaping rewards and the response of corresponding testing returns. As shown in Figure~\ref{fig:mpe-efficiency-demo}, the shaped reward increases in the beginning and then gradually reaches a plateau. The efficiency loss also exhibits the similar phenomenon. As shown in Figure~\ref{fig:5v6-efficiency-demo}, it can be observed that when instability happens during training as evidenced by the sudden changes of test returns, the shaped reward can consistently manifest this situation (as highlighted in red vertical dashed lines). This feature is crucial in the NAHT, since it could be a frequent and typical case when an unseen agent appears in the environment, which may perturb the training stability. This justifies the effectiveness of the shaping rewards $\phi_{i}(R_{t})$ we propose to fulfil the Efficiency axiom. More importantly, this highlights the importance and necessity of the Efficiency axiom to perceive and measure the changing of environments.
        
    \subsection{Extra Results of OOD Evaluation}
    \label{subsec:extra-results-ood-evaluation}
\begin{figure}[h]
          \centering
          \includegraphics[width=0.4\textwidth]{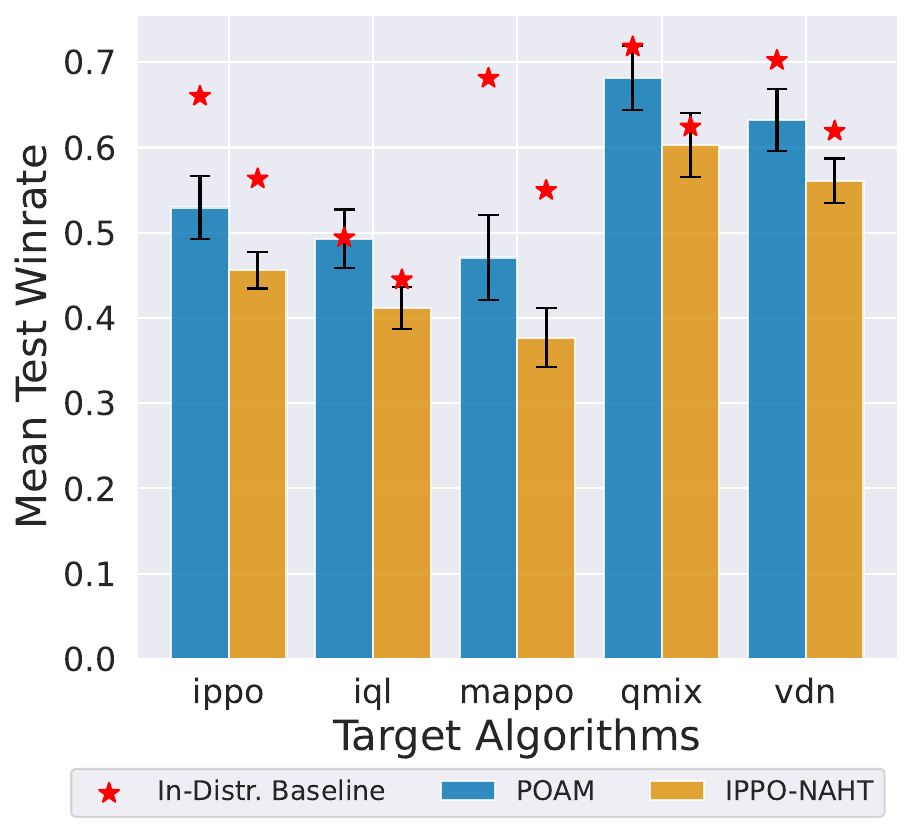}
          \caption{Average evaluation performance on unseen conventions across 4 SMAC tasks. The star denotes the in-convention baseline, and the histograms denotes the unseen conventions.}
        \label{fig:unseen-convention-smac-avg}
        \end{figure}

        We also conducted experiments on evaluating the OOD performance of Shapley Machine and Banzhaf Machine on the base algorithm IPPO (IPPO-NAHT). As seen from Figure~\ref{fig:unseen-convention-ippo}, Banzhaf Machine is competitive to Shapley Machine on dealing with unseen conventions. Although it seems that the result is opposite to the conclusion drawn from the POAM related results, the underlying reason could be the general weaker performance of IPPO than POAM due to the lack of agent modelling, as shown in Figure~\ref{fig:unseen-convention-smac-avg}. This could inhibit the full capability of Shapley Machine. As seen from Figure~\ref{fig:unseen-agent-ippo}, the general performance of Banzhaf Machine is better than Shapley Machine. This is consistent with the conclusion drawn from the POAM related results. In Figure~\ref{fig:convention-ood-other}, we show additional comparisons between algorithms: POAM vs. IPPO, SM-POAM vs. SM-IPPO and BM-POAM vs. BM-IPPO. We find no strong evidence that the axioms are sensitive to the choice of base algorithm.
        
    \subsection{Potential Instability for Large-Scale Scenarios}
    \label{subsec:potential-reasons-for-instability-for-large-scale-scenarios}
\begin{figure}[ht!]
            \centering
            \includegraphics[width=0.85\linewidth]{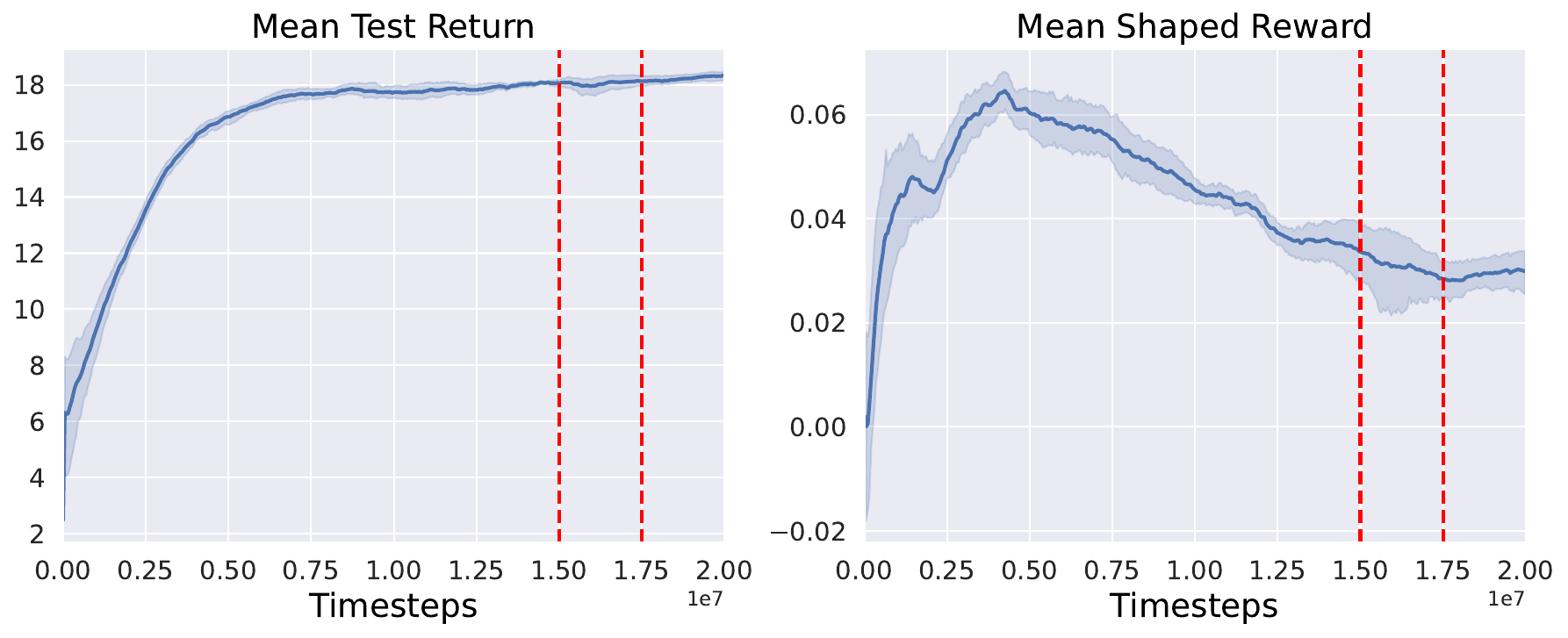}
            \caption{A case of 8v9 with SM-POAM to show the instability of learning progress via the mean shaped rewards. The range bounded by two red dashed lines shows the fluctuation of the learning process.}
        \label{fig:8v9-evidence-of-value-instability}
        \end{figure}

\begin{figure}[ht!]
            \centering
            \includegraphics[width=0.85\linewidth]{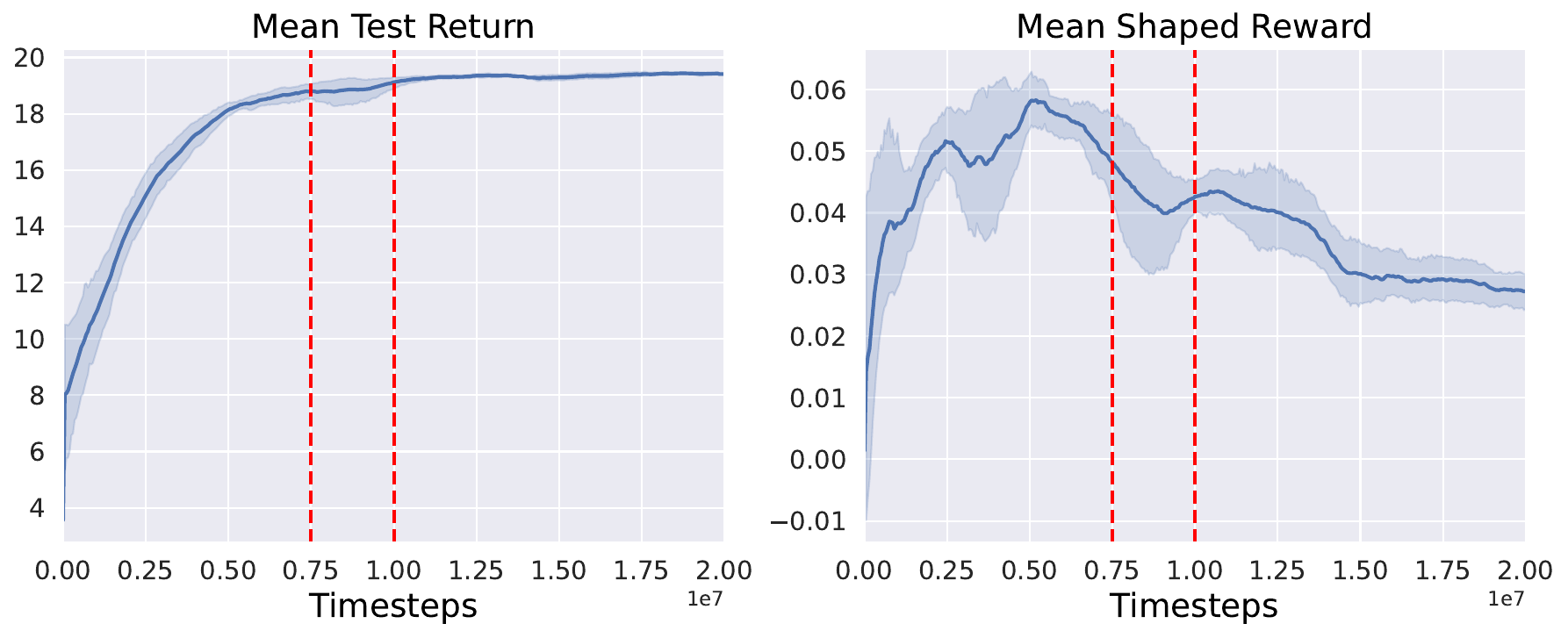}
            \caption{A case of 10v11 with SM-POAM to show the instability of learning progress via the mean shaped rewards. The range bounded by two red dashed lines shows the fluctuation of the learning procedure.}
        \label{fig:10v11-evidence-of-value-instability}
        \end{figure}

        During our repeated trials, we observe that the performance of Shapley Machine is not stable in large-scale scenarios (e.g. 8v9 and 10v11), compared with in small-scale scenarios (e.g. MPE-PP, 3sv5z and 5v6). We now analyze the potential reason behind this phenomenon. As observed from Figures~\ref{fig:8v9-evidence-of-value-instability} and \ref{fig:10v11-evidence-of-value-instability}, it can be confirmed that the instability of shaped rewards is the key reason to cause the instability of learning procedure for large-scale scenarios. According to the functionality of shaping rewards discussed in Appendix~\ref{subsec:analyzing-effects-of-efficiency-axioms}, we hypothesize that the instability could be caused by the unstable changing of value difference in the shaping reward: $\sum_{j \neq i} \left( V_{j}(s_{t}) - \gamma V_{j}(s_{t+1}) \right)$. Specifically, when the number of agents increases, the accumulating prediction error of individual value functions will be amplified. 
        
        To mitigate the above issue, we have added the value loss clip to stabilize the learning procedure of individual value functions for the large-scale scenarios such as 8v9 and 10v11. As seen from Figure~\ref{fig:test-return}, even with the value loss clip the instability of SM-IPPO is still out of control. This could be the cascading effect caused by the lack of agent modelling in contrast to SM-POAM, the instability of which is far beyond the capability of value loss clip. This reflects the value of agent modelling in POAM for tackling the large-scale multi-agent scenarios.
        
\begin{figure}[h!]
            \centering
            \includegraphics[width=0.85\linewidth]{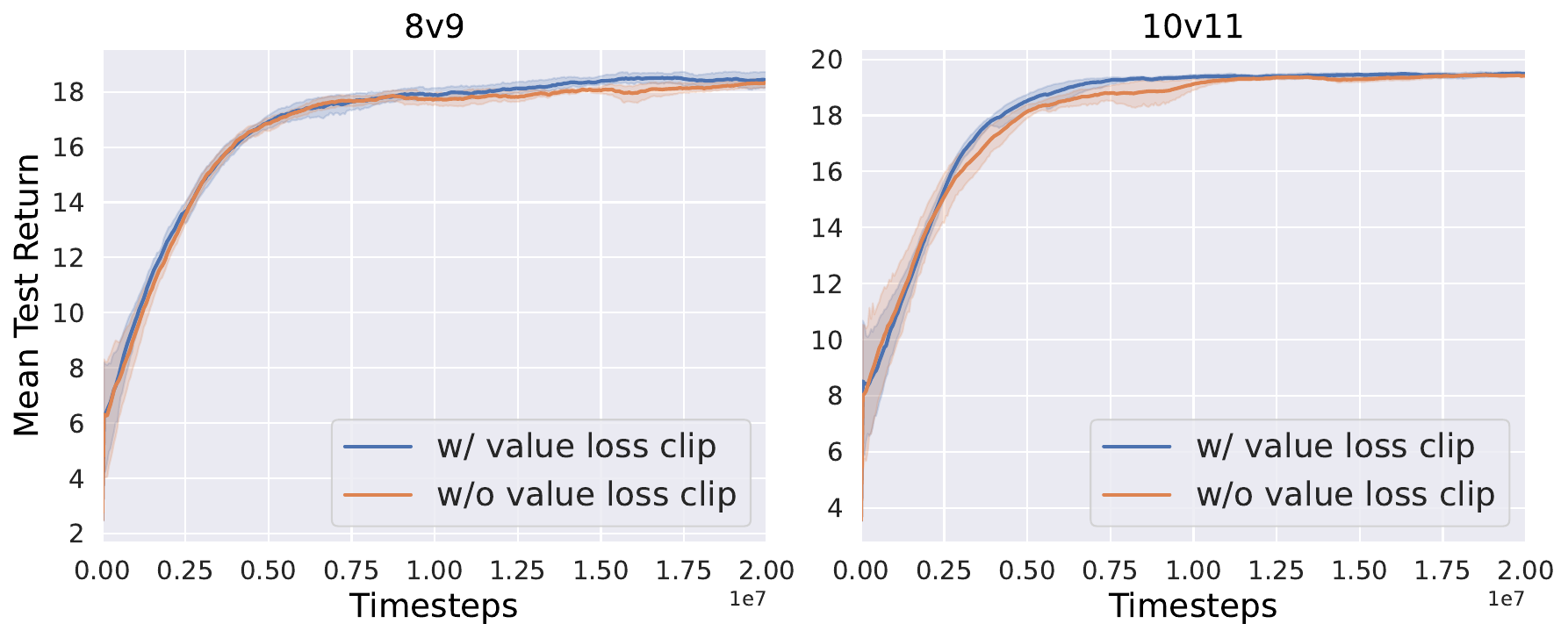}
            \caption{Comparison between the learning procedure of SM-POAM with and without the value loss clip in the 8v9 and 10v11 scenarios.}
        \label{fig:result-value-loss-clip}
        \end{figure}

        As seen from Figure~\ref{fig:result-value-loss-clip}, the learning procedures with the value loss clip for the SM-POAM apparently perform more stably than those without. This verifies our initial hypothesis. Because this strategy will slow down and even hinder the performance for the other three scenarios, we have not posted this as a common strategy. We believe this deserves further investigation in the future before any claims can be made about its general performance.
        
    \subsection{OOD Evaluation Across Different Numbers of Controlled Agents}
    \label{subsec:results-of-n-controlled-agents}
\begin{figure}[!ht]
        \centering
            \begin{subfigure}[b]{\textwidth}
                \centering
                \includegraphics[width=\linewidth]{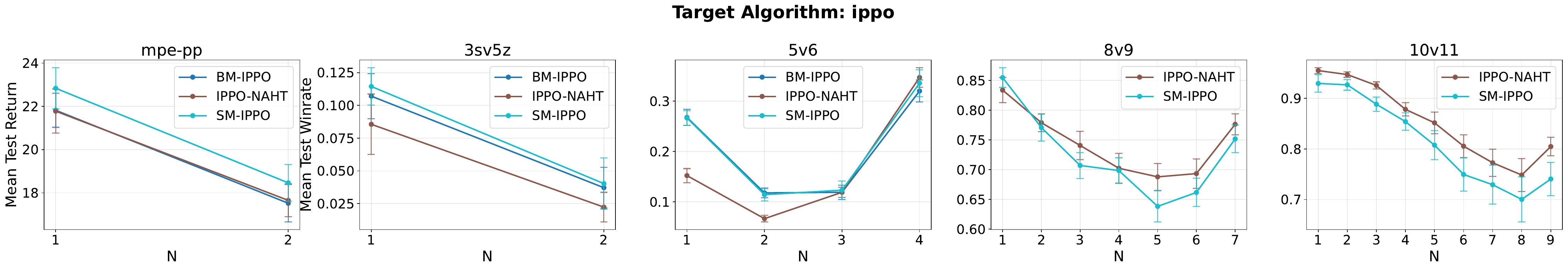}
            \end{subfigure}
            \hfill
            \begin{subfigure}[b]{\textwidth}
                \centering
                \includegraphics[width=\linewidth]{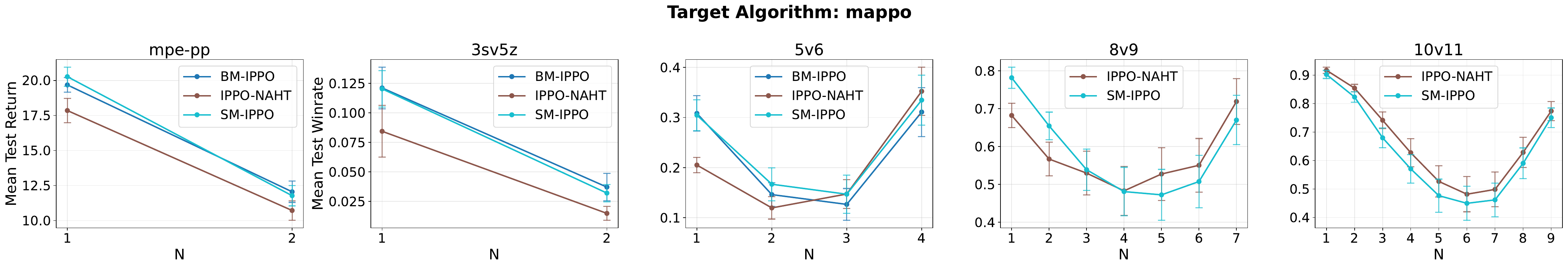}
            \end{subfigure}
            \hfill
            \begin{subfigure}[b]{\textwidth}
                \centering
                \includegraphics[width=\linewidth]{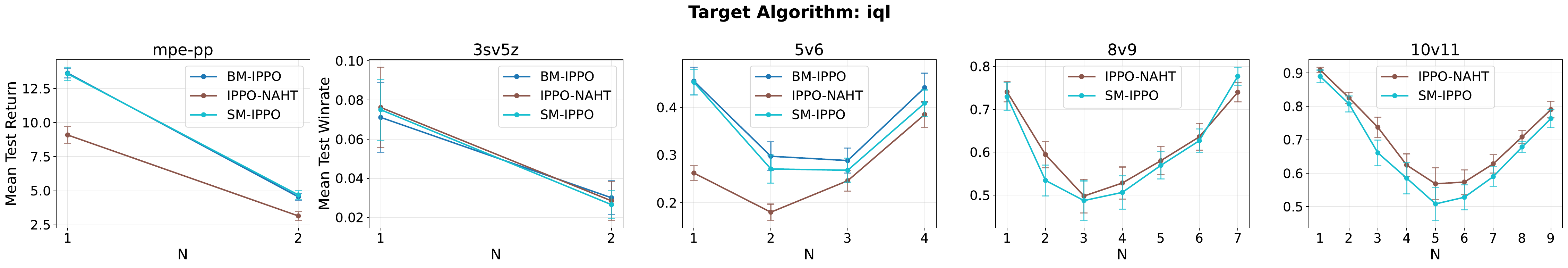}
            \end{subfigure}
            \hfill
            \begin{subfigure}[b]{\textwidth}
                \centering
                \includegraphics[width=\linewidth]{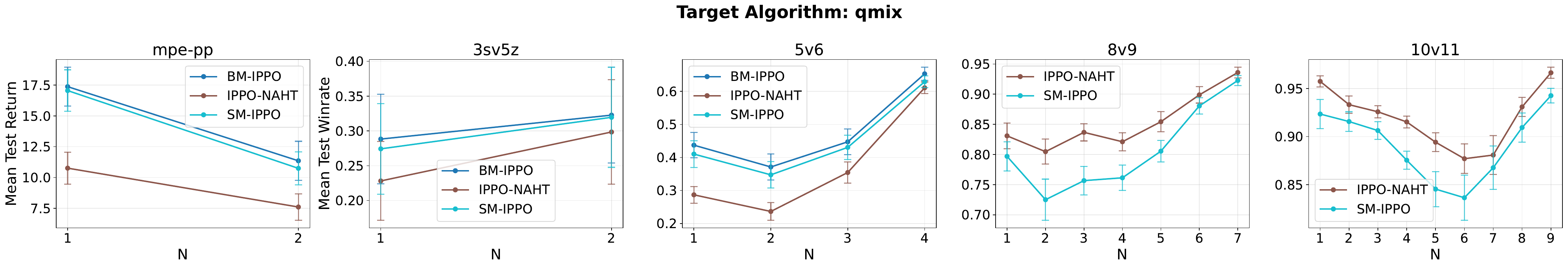}
            \end{subfigure}
            \hfill
            \begin{subfigure}[b]{\textwidth}
                \centering
                \includegraphics[width=\linewidth]{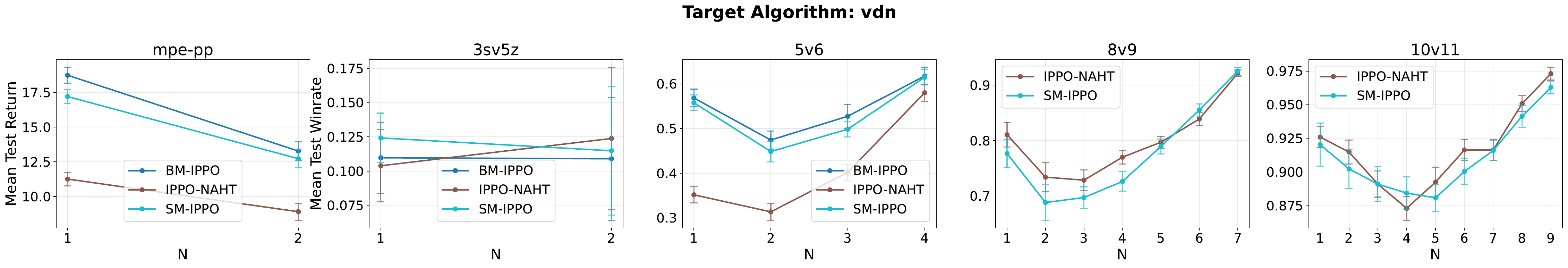}
            \end{subfigure}
        \caption{Out-of-distribution (OOD) evaluation after training, assessing unseen conventions during training, between learned agents (IPPO-NAHT, SM-IPPO and BM-IPPO) and uncontrolled agent types (IPPO, VDN, IQL, QMIX and MAPPO).}
        \label{fig:convention-ood-n-ippo}
        \end{figure}

\begin{figure}[ht!]
        \centering
            \begin{subfigure}[b]{\textwidth}
                \centering
                \includegraphics[width=\linewidth]{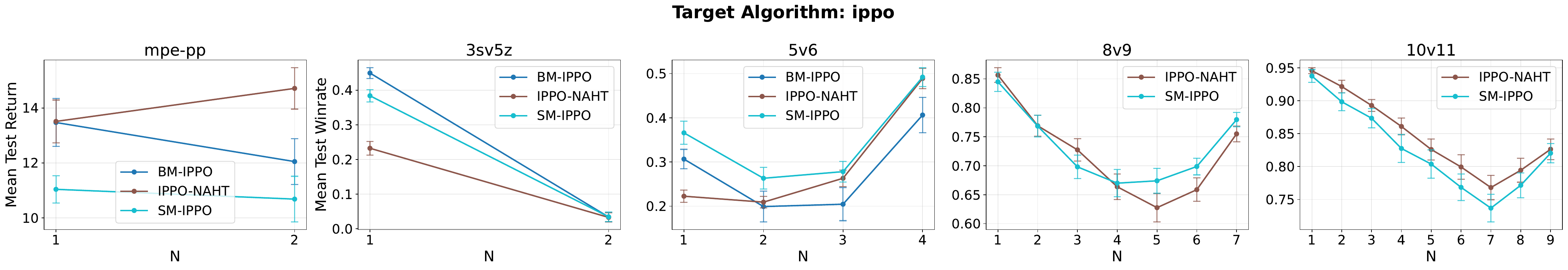}
            \end{subfigure}
            \hfill
            \begin{subfigure}[b]{\textwidth}
                \centering
                \includegraphics[width=\linewidth]{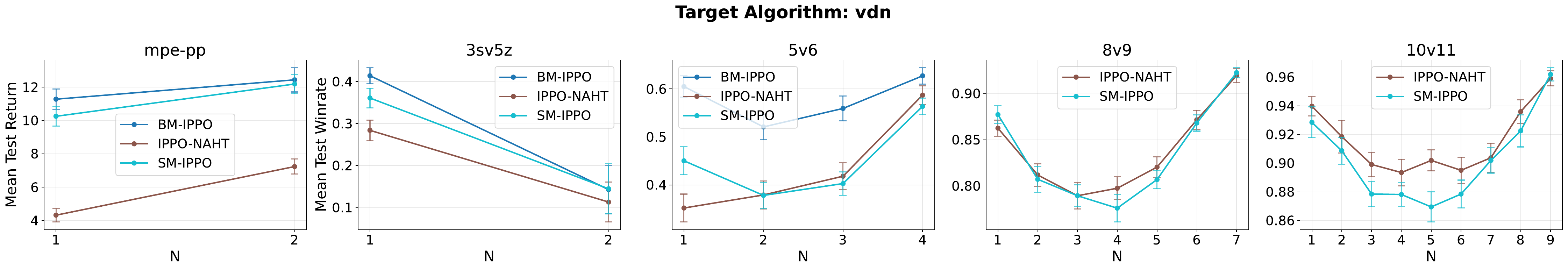}
            \end{subfigure}
        \caption{Out-of-distribution (OOD) evaluation after training, assessing unseen agent types during training, between learned agents (IPPO-NAHT, SM-IPPO and BM-IPPO) and unseen uncontrolled agent types (IPPO and VDN).}
        \label{fig:type-ood-n-ippo}
        \end{figure}

\begin{figure}[ht!]
        \centering
            \begin{subfigure}[b]{\textwidth}
                \centering
                \includegraphics[width=\linewidth]{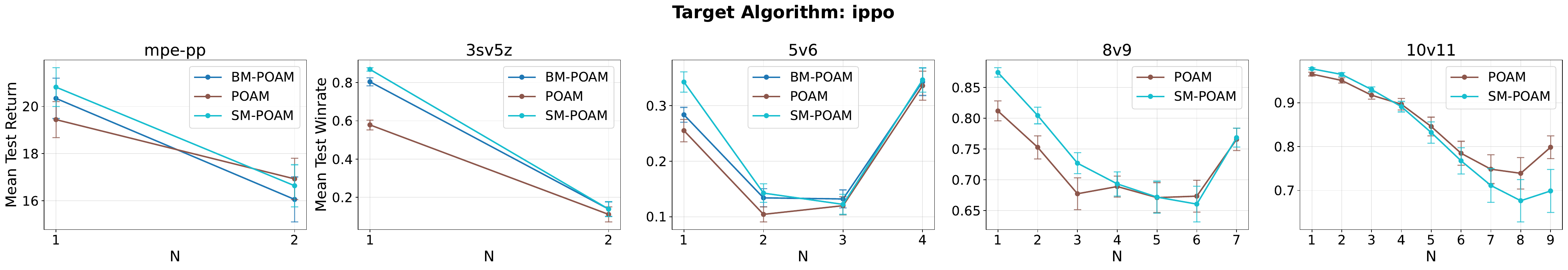}
            \end{subfigure}
            \hfill
            \begin{subfigure}[b]{\textwidth}
                \centering
                \includegraphics[width=\linewidth]{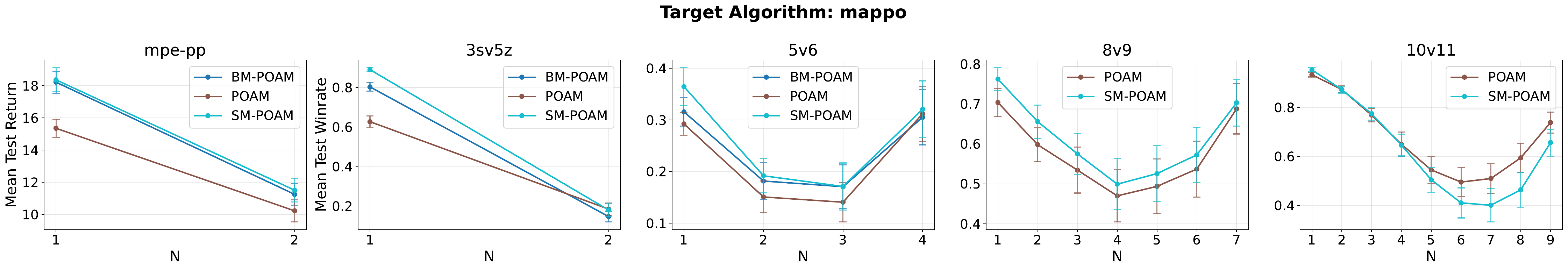}
            \end{subfigure}
            \hfill
            \begin{subfigure}[b]{\textwidth}
                \centering
                \includegraphics[width=\linewidth]{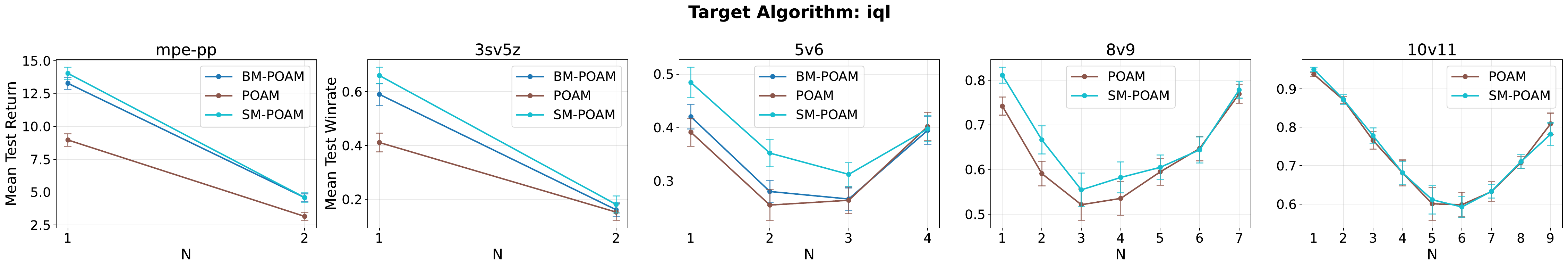}
            \end{subfigure}
            \hfill
            \begin{subfigure}[b]{\textwidth}
                \centering
                \includegraphics[width=\linewidth]{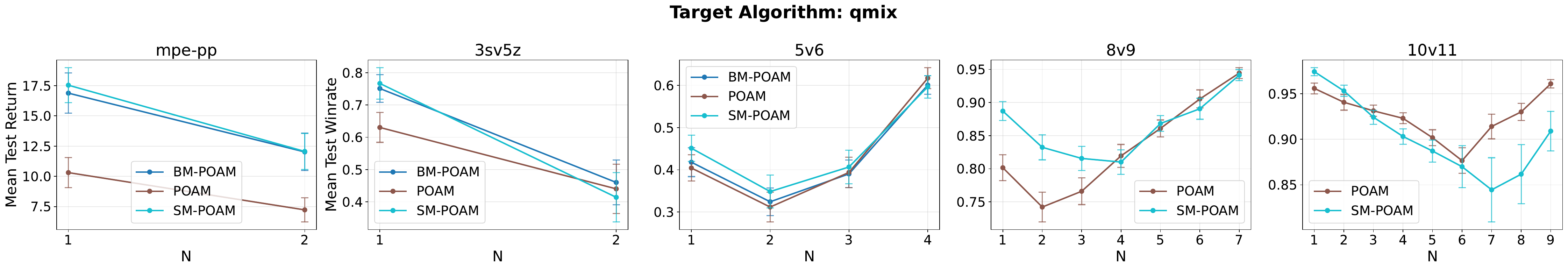}
            \end{subfigure}
            \hfill
            \begin{subfigure}[b]{\textwidth}
                \centering
                \includegraphics[width=\linewidth]{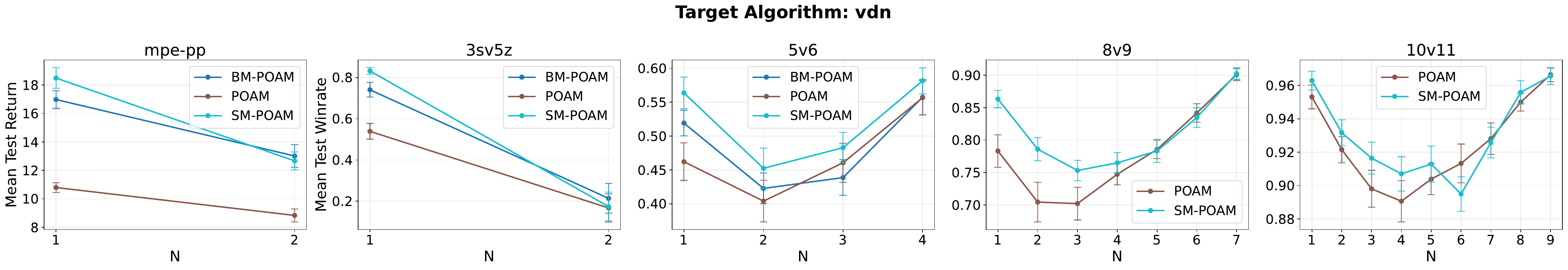}
            \end{subfigure}
        \caption{Out-of-distribution (OOD) evaluation after training, assessing unseen conventions during training, between learned agents (POAM, SM-POAM, and BM-POAM) and uncontrolled agent types (IPPO, VDN, IQL, QMIX and MAPPO).}
        \label{fig:convention-ood-n-poam}
        \end{figure}

\begin{figure}[ht!]
        \centering
            \begin{subfigure}[b]{\textwidth}
                \centering
                \includegraphics[width=\linewidth]{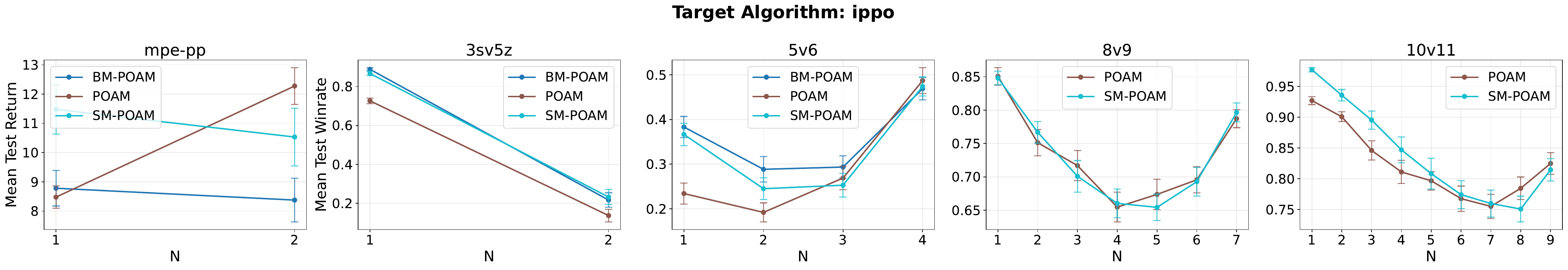}
            \end{subfigure}
            \hfill
            \begin{subfigure}[b]{\textwidth}
                \centering
                \includegraphics[width=\linewidth]{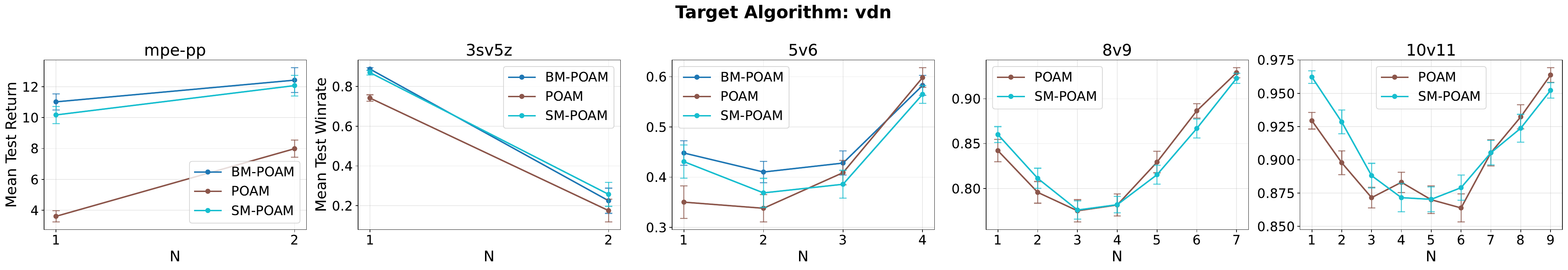}
            \end{subfigure}
        \caption{Out-of-distribution (OOD) evaluation after training, assessing unseen agent types during training, between learned agents (POAM, SM-POAM and BM-POAM) and unseen uncontrolled agent types (IPPO and VDN).}
        \label{fig:type-ood-n-poam}
        \end{figure}

        We report the results on unseen conventions and agent types for each specific number of controlled agents $N$ in Figures~\ref{fig:convention-ood-n-ippo}-\ref{fig:type-ood-n-poam}, by fixing the N value in the M--N score. We would like to highlight that the NAHT problem is a far more difficult task to deal with than the CMARL problem. As a result, our work is only a starting point in this direction, and we believe that the potential stability issue is due to the lack of more exact axioms tailored to the concrete task to face. However, following the framework and principle we develop in this work, this can be improved by considering more axioms that we do not cover in this work, from either the game-theoretical perspective or the physical world.

\clearpage
\section{Broader Impact}
\label{sec:broader-impact}
    This paper introduces an axiomatic framework for learning in open multi-agent systems, with potential applications in domains such as smart grids, transportation networks, and multi-robot systems. By providing a principled foundation for coordinating with partially controllable and potentially unknown agents, the proposed approach may improve the reliability, adaptability, and generalisation of multi-agent systems deployed in real-world environments.

    \subsection{Positive Impacts} 
    A key potential benefit lies in infrastructure systems, such as power distribution networks, where coordination among heterogeneous and partially controllable agents (e.g., distributed energy resources, consumer devices) is essential. More robust and generalisable learning algorithms could improve efficiency, reduce energy waste, and enhance grid stability. Similarly, in autonomous transportation and logistics, improved coordination with unknown or non-cooperative agents may lead to safer and more efficient traffic management and resource allocation.

    The proposed axiomatic perspective may also contribute to trustworthy AI, as it provides interpretable structural constraints on learning algorithms. By grounding value function design in well-established principles from cooperative game theory, the framework may improve transparency and facilitate reasoning about agent behaviour in safety-critical settings.

    \subsection{Potential Risks and Negative Impacts}
    Despite these benefits, the deployment of open multi-agent systems introduces several risks. First, improved coordination capabilities could be applied in strategic or adversarial settings, such as automated market manipulation or competitive multi-agent environments, where agents learn to exploit or collude with others. The ability to generalise to unseen agent behaviours may amplify such risks.

    Second, systems trained under the proposed framework may exhibit unintended coordination behaviours when interacting with human users or legacy systems, especially if the underlying assumptions about cooperation do not hold. In safety-critical domains, such behaviours could lead to undesirable or unpredictable outcomes.

    Third, while the framework introduces structural constraints, it does not inherently guarantee fairness or alignment with human values. The allocation of value among agents, even when principled from a game-theoretic perspective, may not correspond to socially desirable outcomes in all contexts.
\clearpage
\newpage
\section*{NeurIPS Paper Checklist}

\begin{enumerate}

\item {\bf Claims}
    \item[] Question: Do the main claims made in the abstract and introduction accurately reflect the paper's contributions and scope?
    \item[] Answer: \answerYes{} 
    \item[] Justification: All key contributions covered in the abstract and introduction can be well matched to the theoretical results described in Sections \ref{sec:a-model-for-n-agent-ad-hoc-teamwork} and \ref{sec:shapley-value-for-n-agent-ad-hoc-teamwork}, as well as experimental results as shown in Section \ref{sec:experiments}. The scope of this paper is restricted to n-agent ad hoc teamwork, an open multi-agent system problem.
    \item[] Guidelines:
    \begin{itemize}
        \item The answer \answerNA{} means that the abstract and introduction do not include the claims made in the paper.
        \item The abstract and/or introduction should clearly state the claims made, including the contributions made in the paper and important assumptions and limitations. A \answerNo{} or \answerNA{} answer to this question will not be perceived well by the reviewers. 
        \item The claims made should match theoretical and experimental results, and reflect how much the results can be expected to generalize to other settings. 
        \item It is fine to include aspirational goals as motivation as long as it is clear that these goals are not attained by the paper. 
    \end{itemize}

\item {\bf Limitations}
    \item[] Question: Does the paper discuss the limitations of the work performed by the authors?
    \item[] Answer: \answerYes{} 
    \item[] Justification: We have discussed the key limitation of this work in the Section~\ref{sec:conclusion}. The key limitation lies in a parameter of our algorithm that could not well adapt to large-scale agent teams, but it can be improved by relaxing the assumption of our theoretical framework.
    \item[] Guidelines:
    \begin{itemize}
        \item The answer \answerNA{} means that the paper has no limitation while the answer \answerNo{} means that the paper has limitations, but those are not discussed in the paper. 
        \item The authors are encouraged to create a separate ``Limitations'' section in their paper.
        \item The paper should point out any strong assumptions and how robust the results are to violations of these assumptions (e.g., independence assumptions, noiseless settings, model well-specification, asymptotic approximations only holding locally). The authors should reflect on how these assumptions might be violated in practice and what the implications would be.
        \item The authors should reflect on the scope of the claims made, e.g., if the approach was only tested on a few datasets or with a few runs. In general, empirical results often depend on implicit assumptions, which should be articulated.
        \item The authors should reflect on the factors that influence the performance of the approach. For example, a facial recognition algorithm may perform poorly when image resolution is low or images are taken in low lighting. Or a speech-to-text system might not be used reliably to provide closed captions for online lectures because it fails to handle technical jargon.
        \item The authors should discuss the computational efficiency of the proposed algorithms and how they scale with dataset size.
        \item If applicable, the authors should discuss possible limitations of their approach to address problems of privacy and fairness.
        \item While the authors might fear that complete honesty about limitations might be used by reviewers as grounds for rejection, a worse outcome might be that reviewers discover limitations that aren't acknowledged in the paper. The authors should use their best judgment and recognize that individual actions in favor of transparency play an important role in developing norms that preserve the integrity of the community. Reviewers will be specifically instructed to not penalize honesty concerning limitations.
    \end{itemize}

\item {\bf Theory assumptions and proofs}
    \item[] Question: For each theoretical result, does the paper provide the full set of assumptions and a complete (and correct) proof?
    \item[] Answer: \answerYes{} 
    \item[] Justification: All theoretical claims are numbered and we have clarified clearly the corresponding sections in Appendix that include their proofs. In short words, all proofs and justifications for assumptions are given in Appendices \ref{subsec:a-preliminary-result} to \ref{sec:proof-linearity-symmetry-axiom}.
    \item[] Guidelines:
    \begin{itemize}
        \item The answer \answerNA{} means that the paper does not include theoretical results. 
        \item All the theorems, formulas, and proofs in the paper should be numbered and cross-referenced.
        \item All assumptions should be clearly stated or referenced in the statement of any theorems.
        \item The proofs can either appear in the main paper or the supplemental material, but if they appear in the supplemental material, the authors are encouraged to provide a short proof sketch to provide intuition. 
        \item Inversely, any informal proof provided in the core of the paper should be complemented by formal proofs provided in appendix or supplemental material.
        \item Theorems and Lemmas that the proof relies upon should be properly referenced. 
    \end{itemize}

    \item {\bf Experimental result reproducibility}
    \item[] Question: Does the paper fully disclose all the information needed to reproduce the main experimental results of the paper to the extent that it affects the main claims and/or conclusions of the paper (regardless of whether the code and data are provided or not)?
    \item[] Answer: \answerYes{} 
    \item[] Justification: All implementation details, evaluation metrics and hyperparameters for reproducing results are disclosed in Appendices \ref{subsec:implementation-details}, \ref{subsec:evaluation-metrics} and \ref{subsec:hyperparameter-settings}.
    \item[] Guidelines:
    \begin{itemize}
        \item The answer \answerNA{} means that the paper does not include experiments.
        \item If the paper includes experiments, a \answerNo{} answer to this question will not be perceived well by the reviewers: Making the paper reproducible is important, regardless of whether the code and data are provided or not.
        \item If the contribution is a dataset and\slash or model, the authors should describe the steps taken to make their results reproducible or verifiable. 
        \item Depending on the contribution, reproducibility can be accomplished in various ways. For example, if the contribution is a novel architecture, describing the architecture fully might suffice, or if the contribution is a specific model and empirical evaluation, it may be necessary to either make it possible for others to replicate the model with the same dataset, or provide access to the model. In general. releasing code and data is often one good way to accomplish this, but reproducibility can also be provided via detailed instructions for how to replicate the results, access to a hosted model (e.g., in the case of a large language model), releasing of a model checkpoint, or other means that are appropriate to the research performed.
        \item While NeurIPS does not require releasing code, the conference does require all submissions to provide some reasonable avenue for reproducibility, which may depend on the nature of the contribution. For example
        \begin{enumerate}
            \item If the contribution is primarily a new algorithm, the paper should make it clear how to reproduce that algorithm.
            \item If the contribution is primarily a new model architecture, the paper should describe the architecture clearly and fully.
            \item If the contribution is a new model (e.g., a large language model), then there should either be a way to access this model for reproducing the results or a way to reproduce the model (e.g., with an open-source dataset or instructions for how to construct the dataset).
            \item We recognize that reproducibility may be tricky in some cases, in which case authors are welcome to describe the particular way they provide for reproducibility. In the case of closed-source models, it may be that access to the model is limited in some way (e.g., to registered users), but it should be possible for other researchers to have some path to reproducing or verifying the results.
        \end{enumerate}
    \end{itemize}

\item {\bf Open access to data and code}
    \item[] Question: Does the paper provide open access to the data and code, with sufficient instructions to faithfully reproduce the main experimental results, as described in supplemental material?
    \item[] Answer: \answerYes{} 
    \item[] Justification: The opensource code is uploaded as supplemental materials, along with instructions to reproduce key results.
    \item[] Guidelines:
    \begin{itemize}
        \item The answer \answerNA{} means that paper does not include experiments requiring code.
        \item Please see the NeurIPS code and data submission guidelines (\url{https://neurips.cc/public/guides/CodeSubmissionPolicy}) for more details.
        \item While we encourage the release of code and data, we understand that this might not be possible, so \answerNo{} is an acceptable answer. Papers cannot be rejected simply for not including code, unless this is central to the contribution (e.g., for a new open-source benchmark).
        \item The instructions should contain the exact command and environment needed to run to reproduce the results. See the NeurIPS code and data submission guidelines (\url{https://neurips.cc/public/guides/CodeSubmissionPolicy}) for more details.
        \item The authors should provide instructions on data access and preparation, including how to access the raw data, preprocessed data, intermediate data, and generated data, etc.
        \item The authors should provide scripts to reproduce all experimental results for the new proposed method and baselines. If only a subset of experiments are reproducible, they should state which ones are omitted from the script and why.
        \item At submission time, to preserve anonymity, the authors should release anonymized versions (if applicable).
        \item Providing as much information as possible in supplemental material (appended to the paper) is recommended, but including URLs to data and code is permitted.
    \end{itemize}

\item {\bf Experimental setting/details}
    \item[] Question: Does the paper specify all the training and test details (e.g., data splits, hyperparameters, how they were chosen, type of optimizer) necessary to understand the results?
    \item[] Answer: \answerYes{} 
    \item[] Justification: All training and testing details are shown in Appendices \ref{subsec:evaluation-metrics} and \ref{subsec:hyperparameter-settings}.
    \item[] Guidelines:
    \begin{itemize}
        \item The answer \answerNA{} means that the paper does not include experiments.
        \item The experimental setting should be presented in the core of the paper to a level of detail that is necessary to appreciate the results and make sense of them.
        \item The full details can be provided either with the code, in appendix, or as supplemental material.
    \end{itemize}

\item {\bf Experiment statistical significance}
    \item[] Question: Does the paper report error bars suitably and correctly defined or other appropriate information about the statistical significance of the experiments?
    \item[] Answer: \answerYes{} 
    \item[] Justification: All experimental results are demonstrated following mean with 95\% CIs over 5 random seeds.
    \item[] Guidelines:
    \begin{itemize}
        \item The answer \answerNA{} means that the paper does not include experiments.
        \item The authors should answer \answerYes{} if the results are accompanied by error bars, confidence intervals, or statistical significance tests, at least for the experiments that support the main claims of the paper.
        \item The factors of variability that the error bars are capturing should be clearly stated (for example, train/test split, initialization, random drawing of some parameter, or overall run with given experimental conditions).
        \item The method for calculating the error bars should be explained (closed form formula, call to a library function, bootstrap, etc.)
        \item The assumptions made should be given (e.g., Normally distributed errors).
        \item It should be clear whether the error bar is the standard deviation or the standard error of the mean.
        \item It is OK to report 1-sigma error bars, but one should state it. The authors should preferably report a 2-sigma error bar than state that they have a 96\% CI, if the hypothesis of Normality of errors is not verified.
        \item For asymmetric distributions, the authors should be careful not to show in tables or figures symmetric error bars that would yield results that are out of range (e.g., negative error rates).
        \item If error bars are reported in tables or plots, the authors should explain in the text how they were calculated and reference the corresponding figures or tables in the text.
    \end{itemize}

\item {\bf Experiments compute resources}
    \item[] Question: For each experiment, does the paper provide sufficient information on the computer resources (type of compute workers, memory, time of execution) needed to reproduce the experiments?
    \item[] Answer: \answerYes{} 
    \item[] Justification: All computer resources for experiments are disclosed in Appendix \ref{subsec:computational-resources}.
    \item[] Guidelines:
    \begin{itemize}
        \item The answer \answerNA{} means that the paper does not include experiments.
        \item The paper should indicate the type of compute workers CPU or GPU, internal cluster, or cloud provider, including relevant memory and storage.
        \item The paper should provide the amount of compute required for each of the individual experimental runs as well as estimate the total compute. 
        \item The paper should disclose whether the full research project required more compute than the experiments reported in the paper (e.g., preliminary or failed experiments that didn't make it into the paper). 
    \end{itemize}
    
\item {\bf Code of ethics}
    \item[] Question: Does the research conducted in the paper conform, in every respect, with the NeurIPS Code of Ethics \url{https://neurips.cc/public/EthicsGuidelines}?
    \item[] Answer: \answerYes{} 
    \item[] Justification: This paper does not violate any term in the code of ethics.
    \item[] Guidelines:
    \begin{itemize}
        \item The answer \answerNA{} means that the authors have not reviewed the NeurIPS Code of Ethics.
        \item If the authors answer \answerNo, they should explain the special circumstances that require a deviation from the Code of Ethics.
        \item The authors should make sure to preserve anonymity (e.g., if there is a special consideration due to laws or regulations in their jurisdiction).
    \end{itemize}

\item {\bf Broader impacts}
    \item[] Question: Does the paper discuss both potential positive societal impacts and negative societal impacts of the work performed?
    \item[] Answer: \answerYes{} 
    \item[] Justification: We have clearly discussed both positive and potentially negative impacts in Appendix \ref{sec:broader-impact}.
    \item[] Guidelines:
    \begin{itemize}
        \item The answer \answerNA{} means that there is no societal impact of the work performed.
        \item If the authors answer \answerNA{} or \answerNo, they should explain why their work has no societal impact or why the paper does not address societal impact.
        \item Examples of negative societal impacts include potential malicious or unintended uses (e.g., disinformation, generating fake profiles, surveillance), fairness considerations (e.g., deployment of technologies that could make decisions that unfairly impact specific groups), privacy considerations, and security considerations.
        \item The conference expects that many papers will be foundational research and not tied to particular applications, let alone deployments. However, if there is a direct path to any negative applications, the authors should point it out. For example, it is legitimate to point out that an improvement in the quality of generative models could be used to generate Deepfakes for disinformation. On the other hand, it is not needed to point out that a generic algorithm for optimizing neural networks could enable people to train models that generate Deepfakes faster.
        \item The authors should consider possible harms that could arise when the technology is being used as intended and functioning correctly, harms that could arise when the technology is being used as intended but gives incorrect results, and harms following from (intentional or unintentional) misuse of the technology.
        \item If there are negative societal impacts, the authors could also discuss possible mitigation strategies (e.g., gated release of models, providing defenses in addition to attacks, mechanisms for monitoring misuse, mechanisms to monitor how a system learns from feedback over time, improving the efficiency and accessibility of ML).
    \end{itemize}
    
\item {\bf Safeguards}
    \item[] Question: Does the paper describe safeguards that have been put in place for responsible release of data or models that have a high risk for misuse (e.g., pre-trained language models, image generators, or scraped datasets)?
    \item[] Answer: \answerNA{} 
    \item[] Justification: Our work is mainly about theoretical aspects, with no relation to such issues.
    \item[] Guidelines:
    \begin{itemize}
        \item The answer \answerNA{} means that the paper poses no such risks.
        \item Released models that have a high risk for misuse or dual-use should be released with necessary safeguards to allow for controlled use of the model, for example by requiring that users adhere to usage guidelines or restrictions to access the model or implementing safety filters. 
        \item Datasets that have been scraped from the Internet could pose safety risks. The authors should describe how they avoided releasing unsafe images.
        \item We recognize that providing effective safeguards is challenging, and many papers do not require this, but we encourage authors to take this into account and make a best faith effort.
    \end{itemize}

\item {\bf Licenses for existing assets}
    \item[] Question: Are the creators or original owners of assets (e.g., code, data, models), used in the paper, properly credited and are the license and terms of use explicitly mentioned and properly respected?
    \item[] Answer: \answerYes{} 
    \item[] Justification: All baseline algorithm (codes) we used for experiments have been credited by citing the corresponding work.
    \item[] Guidelines:
    \begin{itemize}
        \item The answer \answerNA{} means that the paper does not use existing assets.
        \item The authors should cite the original paper that produced the code package or dataset.
        \item The authors should state which version of the asset is used and, if possible, include a URL.
        \item The name of the license (e.g., CC-BY 4.0) should be included for each asset.
        \item For scraped data from a particular source (e.g., website), the copyright and terms of service of that source should be provided.
        \item If assets are released, the license, copyright information, and terms of use in the package should be provided. For popular datasets, \url{paperswithcode.com/datasets} has curated licenses for some datasets. Their licensing guide can help determine the license of a dataset.
        \item For existing datasets that are re-packaged, both the original license and the license of the derived asset (if it has changed) should be provided.
        \item If this information is not available online, the authors are encouraged to reach out to the asset's creators.
    \end{itemize}

\item {\bf New assets}
    \item[] Question: Are new assets introduced in the paper well documented and is the documentation provided alongside the assets?
    \item[] Answer: \answerYes{} 
    \item[] Justification: The code for training has been with a clear documentation.
    \item[] Guidelines:
    \begin{itemize}
        \item The answer \answerNA{} means that the paper does not release new assets.
        \item Researchers should communicate the details of the dataset\slash code\slash model as part of their submissions via structured templates. This includes details about training, license, limitations, etc. 
        \item The paper should discuss whether and how consent was obtained from people whose asset is used.
        \item At submission time, remember to anonymize your assets (if applicable). You can either create an anonymized URL or include an anonymized zip file.
    \end{itemize}

\item {\bf Crowdsourcing and research with human subjects}
    \item[] Question: For crowdsourcing experiments and research with human subjects, does the paper include the full text of instructions given to participants and screenshots, if applicable, as well as details about compensation (if any)? 
    \item[] Answer: \answerNA{} 
    \item[] Justification: Our work has no relation to this term.
    \item[] Guidelines:
    \begin{itemize}
        \item The answer \answerNA{} means that the paper does not involve crowdsourcing nor research with human subjects.
        \item Including this information in the supplemental material is fine, but if the main contribution of the paper involves human subjects, then as much detail as possible should be included in the main paper. 
        \item According to the NeurIPS Code of Ethics, workers involved in data collection, curation, or other labor should be paid at least the minimum wage in the country of the data collector. 
    \end{itemize}

\item {\bf Institutional review board (IRB) approvals or equivalent for research with human subjects}
    \item[] Question: Does the paper describe potential risks incurred by study participants, whether such risks were disclosed to the subjects, and whether Institutional Review Board (IRB) approvals (or an equivalent approval/review based on the requirements of your country or institution) were obtained?
    \item[] Answer: \answerNA{} 
    \item[] Justification: This paper is with no relation to this term.
    \item[] Guidelines:
    \begin{itemize}
        \item The answer \answerNA{} means that the paper does not involve crowdsourcing nor research with human subjects.
        \item Depending on the country in which research is conducted, IRB approval (or equivalent) may be required for any human subjects research. If you obtained IRB approval, you should clearly state this in the paper. 
        \item We recognize that the procedures for this may vary significantly between institutions and locations, and we expect authors to adhere to the NeurIPS Code of Ethics and the guidelines for their institution. 
        \item For initial submissions, do not include any information that would break anonymity (if applicable), such as the institution conducting the review.
    \end{itemize}

\item {\bf Declaration of LLM usage}
    \item[] Question: Does the paper describe the usage of LLMs if it is an important, original, or non-standard component of the core methods in this research? Note that if the LLM is used only for writing, editing, or formatting purposes and does \emph{not} impact the core methodology, scientific rigor, or originality of the research, declaration is not required.
    \item[] Answer: \answerNA{} 
    \item[] Justification: The core part of research does not involve LLMs.
    \item[] Guidelines:
    \begin{itemize}
        \item The answer \answerNA{} means that the core method development in this research does not involve LLMs as any important, original, or non-standard components.
        \item Please refer to our LLM policy in the NeurIPS handbook for what should or should not be described.
    \end{itemize}

\end{enumerate}

\end{document}